\pdfoutput=1

\documentclass[11pt]{article}

\usepackage[preprint]{acl}

\usepackage{times}
\usepackage{latexsym}
\usepackage{pgf-pie}
\usepackage{colortbl}
\usepackage{CJKutf8}
\usepackage{tabularx,booktabs}
\usepackage{pgfplots, pgfplotstable}
    \usepgfplotslibrary{groupplots}
\pgfplotsset{compat=1.9}
\usetikzlibrary{
    pgfplots.dateplot,
}
\usepackage{hyperref}

\usepackage{supertabular}
\usepackage{longtable}
\usepackage{amsmath}
\usepackage{mathabx}
\usepackage{booktabs} 
\usepackage{tabto}
\usepackage{multicol}
\usepackage{pifont}
\newcommand{\cmark}{\ding{51}}%
\newcommand{\xmark}{\ding{55}}%

\newcommand\cjkhl{\bgroup\markoverwith
  {\textcolor{yellow}{\rule[-.5ex]{2pt}{2.5ex}}}\ULon}

\usepackage[textsize=scriptsize]{todonotes}

\usepackage[T1]{fontenc}

\usepackage[utf8]{inputenc}

\usepackage{microtype}

\usepackage{inconsolata}

\usepackage{soul}
\usepackage{tikz}
\usepackage{longtable}
\usepackage[export]{adjustbox}
\usepackage{subcaption}
\usepackage{makecell}
\usepackage{cleveref}
\usepackage{graphicx}
\usepackage{xcolor, color}
\usepackage{graphicx}
\usepackage{multirow}
\usepackage{booktabs}
\usepackage{comment}
\usepackage{float}
\usepackage{enumitem}
\usepackage{pbox}
\usepackage{dcolumn}
\usepackage{anyfontsize} 
\definecolor{lightblue}{RGB}{126, 178, 221}
\definecolor{darkred}{RGB}{105,17,10}
\definecolor{lightyellow}{RGB}{251,242,214}
\definecolor{darkyellow}{RGB}{82,58,34}
\definecolor{lightgrey}{RGB}{230,230,230}
\definecolor{darkgrey}{RGB}{57,57,57}
\definecolor{lightgreen}{RGB}{56, 176, 0}
\definecolor{darkgreen}{RGB}{34,139,34}
\definecolor{amethyst}{rgb}{153, 102, 204}
\definecolor{lightpurple}{RGB}{200, 140, 200}
\definecolor{darkpurple}{RGB}{160, 0, 190}
\definecolor{color1bg}{RGB}{69,140,214} 
\definecolor{color7bg}{RGB}{225,157,242} 
\definecolor{color2bg}{RGB}{74,195,200} 
\definecolor{color3bg}{RGB}{134,217,90} 
\definecolor{color4bg}{RGB}{245,207,70} 
\definecolor{color5bg}{RGB}{255,158,80} 
\definecolor{color6bg}{RGB}{245,73,72} 
\definecolor{pink}{RGB}{247, 134, 170}
\definecolor{clozecolor1}{RGB}{225,157,242}
\definecolor{clozecolor2}{RGB}{142,136,247} 
\definecolor{clozecolor3}{RGB}{255, 159, 26} 
\definecolor{allcolors}{RGB}{102, 102, 102}
\definecolor{all3models}{RGB}{153, 153, 153}
\definecolor{bing}{RGB}{17, 138, 178}
\definecolor{google}{RGB}{6, 214, 160}
\definecolor{deepl}{RGB}{255, 209, 102}
\definecolor{m2m}{RGB}{239, 71, 111}
\definecolor{line1color}{RGB}{7, 42, 200}
\definecolor{line2color}{RGB}{165, 76, 20}
\definecolor{line3color}{RGB}{251, 111, 146}
\definecolor{piecolor1}{RGB}{132, 255, 201}
\definecolor{piecolor2}{RGB}{170, 178, 255}
\definecolor{piecolor3}{RGB}{236, 160, 255}
\definecolor{rankingcolor1}{RGB}{0, 119, 182}
\definecolor{rankingcolor2}{RGB}{247, 127, 0}
\definecolor{pie1}{RGB}{132, 255, 201}
\definecolor{pie2}{RGB}{170, 178, 255}
\definecolor{pie3}{RGB}{254, 215, 102}
\definecolor{green1}{HTML}{6A994E}
\definecolor{brown}{HTML}{c28a60}
\definecolor{gray1}{HTML}{9ba28f}
\definecolor{gray2}{HTML}{616154}
\definecolor{purple2}{HTML}{9d4edd}
\definecolor{purple1}{HTML}{560bad}
\definecolor{red1}{HTML}{c1121f}
\definecolor{green1}{HTML}{007200}
\definecolor{green2}{HTML}{38b000}
\definecolor{green3}{HTML}{004b23}
\definecolor{orange1}{HTML}{c08552}
\definecolor{brown}{HTML}{936639}
\definecolor{gray1}{HTML}{495057}

\newcommand{\hlc}[2][yellow]{{%
    \colorlet{foo}{#1}%
    \sethlcolor{foo}\hl{#2}}%
}

\usepackage{tabularx,booktabs}
\newcolumntype{Y}{>{\centering\arraybackslash}X}
\newcolumntype{d}[1]{D{.}{.}{#1}}
\usepackage{array}

\usepackage{devanagari}
\usepackage{scalerel,xparse}

\newcolumntype{M}[1]{>{\raggedright\arraybackslash}m{#1}}
\newcolumntype{P}[1]{>{\raggedright\arraybackslash}m{#1}}
\newcolumntype{C}[1]{>{\centering\arraybackslash}m{#1}}

\NewDocumentCommand\emojineo{}{
    \includegraphics[scale=0.05]{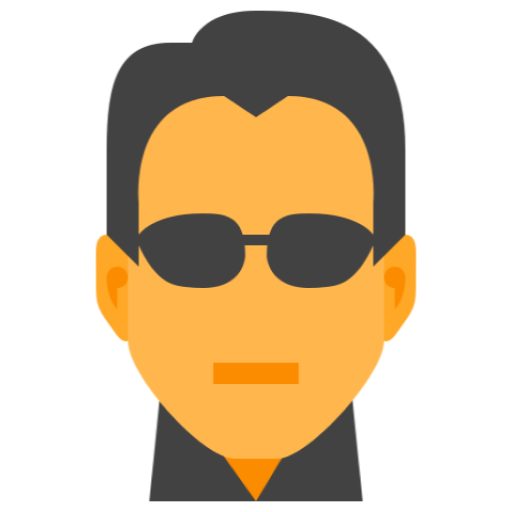}
}

\title{\emojineo{\sc Neo-Bench}: \\ Evaluating Robustness of Large Language Models with Neologisms}

\author{Jonathan Zheng, Alan Ritter, Wei Xu \\
  College of Computing \\
  Georgia Institute of Technology \\
  \fontsize{10}{10}{\texttt{jzheng324@gatech.edu};
  \texttt{\{wei.xu, alan.ritter\}@cc.gatech.edu}}\\
  }

\begin{document}
\maketitle
\begin{abstract}
The performance of Large Language Models (LLMs) degrades from the temporal drift between data used for model training and newer text seen during inference. One understudied avenue of language change causing data drift is the emergence of neologisms -- new word forms -- over time. We create a diverse resource of recent English neologisms by using several popular collection methods. We analyze temporal drift using neologisms by comparing sentences containing new words with near-identical sentences that replace neologisms with existing substitute words. Model performance is nearly halved in machine translation when a single neologism is introduced in a sentence. Motivated by these results, we construct a benchmark to evaluate LLMs' ability to generalize to neologisms with various natural language understanding tasks and model perplexity. Models with later knowledge cutoff dates yield lower perplexities and perform better in downstream tasks. LLMs are also affected differently based on the linguistic origins of words, indicating that neologisms are complex for static LLMs to address.  We release our benchmark at: \href{https://github.com/JonathanQZheng/NEO-BENCH}{https://github.com/JonathanQZheng/NEO-BENCH}. 
  
  \end{abstract}

\section{Introduction}
  \begin{figure}[!tbp]
    \centering
    \includegraphics[width=0.47\textwidth]{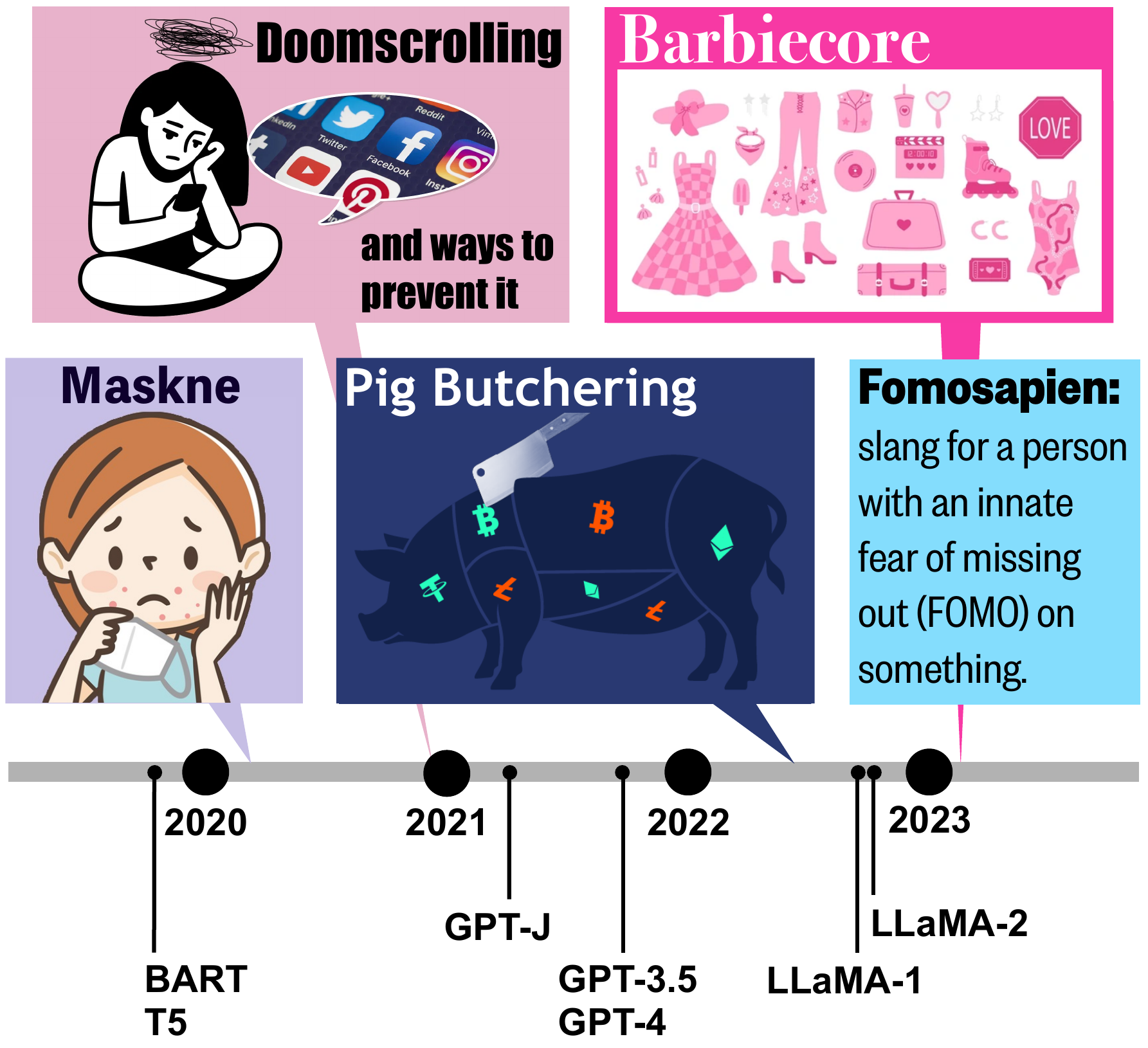}
\vspace*{-3mm}
  \caption{{\sc Neo-Bench} collects neologisms from 2020-2023 for LLM evaluation. ``Pig Butchering'' originated as a Mandarin expression \begin{CJK*}{UTF8}{gbsn}(杀猪盘)\end{CJK*}.}
    \label{fig:timeline}
\end{figure}

\begin{figure*}[!ht]
\pgfplotsset{
    testbar11/.style={
        xbar stacked,
        bar width=5.5pt,
        xtick pos=lower, ytick pos=left,
        xmin=0,xmax=101,
        ytick = data,
        ytick distance=1,
        yticklabel style={text width=1.75cm, align=right, yshift=0cm, font=\fontsize{6.5}{6}\selectfont},
        ytick style={yshift=0cm},
        legend style={at={(0.3,0.14)},anchor=west, font=\tiny},
        tick align = outside,
        height=0.687\textwidth,
        xlabel style = {font=\small},
        y=2.7mm,
        enlarge y limits={abs=0.95},
    }}
\begin{tikzpicture}
\tikzstyle{every node}=[font=\fontsize{5}{5}\selectfont]
\begin{axis}[testbar11, name=plot1, anchor=north, xtick=\empty, yticklabels = {Oracle Ensemble, Microsoft Bing, Google Translate, DeepL Translator, GPT-4, GPT-3.5, ALMA 7B, M2M100 1.2B}, legend style={at={(0.5,0.12)},anchor=north}, title style={at={(0.5,1.10)}, font=\small,anchor=north,yshift=-0.1}, title = \textbf{a) Sentences Containing Neologisms}, legend columns=-1,] 
 \addplot[fill=color1bg,nodes near coords] coordinates{
                                    (78, 7)
                                    (47, 6)
                                    (45, 4)
                                    (38, 3)
                                    (46, 5)
                                    (34, 2)
                                    (27, 1)
                                    (15, 0)                                    
                                    };
 \addplot[fill=color2bg,nodes near coords] coordinates{
                                   (9, 7) 
                                   (2, 6)
                                   (6, 4)
                                   (6, 3)
                                   (7, 5)
                                   (13, 2)
                                   (10, 1)
                                   (10, 0)
                                   };

 \addplot[fill=color3bg,nodes near coords] coordinates{
                                   (9, 7) 
                                   (21, 6)
                                   (11, 4)
                                   (11, 3)
                                   (13, 5)
                                   (10, 2)
                                   (8, 1)
                                   (22, 0)                                   
                                   };
                                   
 \addplot[fill=color4bg,nodes near coords] coordinates{
                                   (0, 7)
                                   (3, 6)
                                   (3, 4)
                                   (7, 3)
                                   (8, 5)
                                   (5, 2)
                                   (9, 1)
                                   (5, 0)
                                   };

 \addplot[fill=color4bg] coordinates{
                                   (1, 7)
                                   (0, 6)
                                   (0, 4)
                                   (0, 3)
                                   (0, 5)
                                   (0, 2)
                                   (0, 1)
                                   (0, 0)
                                   };
                                   
 \addplot[fill=color5bg,nodes near coords] coordinates{
                                   (3, 7)
                                   (15, 6)
                                   (27, 4)
                                   (32, 3)
                                   (21, 5)
                                   (27, 2)
                                   (38, 1)
                                   (31, 0)
                                   };

 \addplot[fill=color6bg,nodes near coords] coordinates{
                                   (0, 7) 
                                   (10, 6)
                                   (7, 4)
                                   (6, 3)
                                   (4, 5)
                                   (10, 2)
                                   (5, 1)
                                   (11, 0)
                                   };
 \addplot[fill=color7bg] coordinates{
                                   (0, 7)
                                   (0, 6)
                                   (1, 4)
                                   (0, 3)
                                   (1, 5)
                                   (1, 2)
                                   (0, 1)
                                   (0, 0)
                                   };
 \addplot[fill=color7bg,nodes near coords] coordinates{
                                   (0, 7)
                                   (2, 6)
                                   (0, 4)
                                   (0, 3)
                                   (0, 5)
                                   (0, 2)
                                   (3, 1)
                                   (6, 0)
                                   };
\end{axis}

\begin{axis}[height=0.226\textwidth, name=text1, xtick=\empty, at = {($(plot1.east)$)}, anchor=west,
ytick=\empty, yticklabels = {}, xmin=0, xmax=1, ymin=0, ymax=1, y=24mm,
legend style={at={(0.64,0.12)},anchor=west},  xlabel=\empty,
title style={at={(0.5,1.10)},font=\small, anchor=north,yshift=-0.1}, axis y line*=right,  dash pattern=on 1pt off 1pt, dash phase=1pt] 
\node[font=\small, text width=2.7cm, align=left] at (axis cs: 0.55,0.5) {{\fontsize{9}{9}\selectfont {\textbf{Example:}}} \vspace{-1pt}\newline {\fontsize{8}{1}\selectfont {Starting to think \vspace{-1pt}\newline \textbf{doomscrolling} \vspace{-1pt}\newline through the fall of \vspace{-1pt}\newline civilization is having \vspace{-1pt}\newline a negative effect on \vspace{-1pt}\newline my mental health.}}};
\end{axis}
\end{tikzpicture}

\vspace{-6pt}
\begin{tikzpicture}
\tikzstyle{every node}=[font=\fontsize{5}{5}\selectfont]
\begin{axis}[testbar11, name=plot2, at= {($(plot1.south)$)}, anchor=north, yticklabels = {Oracle Ensemble, Microsoft Bing,GPT-4, Google Translate, DeepL Translator, GPT-3.5, ALMA 7B, M2M100 1.2B}, legend style={at={(0.5,-0.20), font=\footnotesize,nodes={scale=0.5}},fill=none,anchor=north},
title style={at={(0.5,1.10)}, font=\small,anchor=north,yshift=-0.1},
title = \textbf{b) Same Sentences with Neologisms Replaced by Common Words}, legend columns=-1, enlarge y limits={abs=0.95},
] 
 \addplot[fill=color1bg,nodes near coords] coordinates{
                                    (96, 7)
                                    (73, 6)
                                    (70, 5)
                                    (72, 4)
                                    (72, 3)
                                    (67, 2)
                                    (44, 1)
                                    (37, 0)                                    
                                    };
 \addplot[fill=color2bg,nodes near coords] coordinates{
                                   (4, 7)
                                   (11, 6)
                                   (16, 5)
                                   (13, 4)
                                   (0, 3)
                                   (14, 2)
                                   (17, 1)
                                   (26, 0)
                                   };
 \addplot[fill=color2bg] coordinates{
                                   (0, 7)
                                   (0, 6)
                                   (0, 5)
                                   (0, 4)
                                   (1, 3)
                                   (0, 2)
                                   (0, 1)
                                   (0, 0)
                                   };

 \addplot[fill=color3bg,nodes near coords] coordinates{
                                   (0, 7)
                                   (3, 6)
                                   (0, 5)
                                   (4, 4)
                                   (4, 3)
                                   (4, 2)
                                   (2, 1)
                                   (11, 0)                                   
                                   };

 \addplot[fill=color3bg] coordinates{
                                   (0, 7)
                                   (0, 6)
                                   (1, 5)
                                   (0, 4)
                                   (0, 3)
                                   (0, 2)
                                   (0, 1)
                                   (0, 0)                                   
                                   };
                                   
 \addplot[fill=color4bg,nodes near coords] coordinates{
                                   (0, 7)
                                   (3, 6)
                                   (2, 5)
                                   (2, 4)
                                   (8, 3)
                                   (4, 2)
                                   (4, 1)
                                   (4, 0)
                                   };
                                   
 \addplot[fill=color4bg] coordinates{
                                   (0, 7)
                                   (0, 6)
                                   (0, 5)
                                   (0, 3)
                                   (0, 2)
                                   (0, 4)
                                   (0, 1)
                                   (0, 0)
                                   };
                                   
 \addplot[fill=color5bg,nodes near coords] coordinates{
                                   (0, 7)
                                   (6, 6)
                                   (9, 5)
                                   (6, 4)
                                   (4, 3)
                                   (11, 2)
                                   (19, 1)
                                   (15, 0)
                                   };

 \addplot[fill=color6bg,nodes near coords] coordinates{
                                   (0, 7)
                                   (2, 6)
                                   (2, 5)
                                   (2, 4)
                                   (11, 3)
                                   (0, 2)
                                   (11, 1)
                                   (3, 0)
                                   };

 \addplot[fill=color6bg] coordinates{
                                   (0, 7)
                                   (0, 6)
                                   (0, 5)
                                   (0, 4)
                                   (0, 3)
                                   (0, 2)
                                   (0, 1)
                                   (0, 0)
                                   };

 \addplot[fill=color7bg,nodes near coords] coordinates{
                                    (0, 7)
                                    (2, 6)
                                    (0, 5)
                                    (0, 4)
                                    (0, 3)
                                    (0, 2)
                                    (3, 1)
                                    (4, 0)
                                   };
        
 \addplot[fill=color7bg] coordinates{
                                   (0, 7)
                                   (0, 6)
                                   (0, 5)
                                   (1, 4)
                                   (0, 3)
                                   (0, 2)
                                   (0, 1)
                                   (0, 0)
                                   };

\legend{Good, {}, Unnatural, Literal, {}, Partial, {}, Mistranslation, Copy, Incomprehensible}
\end{axis}

\begin{axis}[height=0.226\textwidth, name=text1, xtick=\empty, at = {($(plot2.east)$)}, anchor=west,
ytick=\empty, yticklabels = {}, xmin=0, xmax=1, ymin=0, ymax=1, y=24mm,
legend style={at={(0.64,0.12)},anchor=west},  xlabel=\empty,
title style={at={(0.5,1.08)},font=\small, anchor=north,yshift=-0.1}, axis y line*=right, dashed, dash pattern=on 1pt off 1pt, dash phase=1pt] 
\node[font=\small, text width=2.7cm, align=left] at (axis cs: 0.55,0.5) {{\fontsize{9}{9}\selectfont {\textbf{Example:}}} \vspace{-1pt}\newline {\fontsize{8}{1}\selectfont {Starting to think \vspace{-1pt}\newline \textbf{smoking} \vspace{-1pt}\newline through the fall of \vspace{-1pt}\newline civilization is having \vspace{-1pt}\newline a negative effect on \vspace{-1pt}\newline my mental health.}}};
\end{axis}
\end{tikzpicture}
\vspace*{-7mm}
\caption{A single neologism can dramatically affect model output, as shown by human evaluation of Machine Translation models on sentences containing neologisms and the same sentences with neologisms replaced by carefully chosen words that also fit in the context.  Oracle ensemble selects the best translation from all models.}
\label{fig:mt_histogram_full}
\vspace*{-4mm}
\end{figure*}
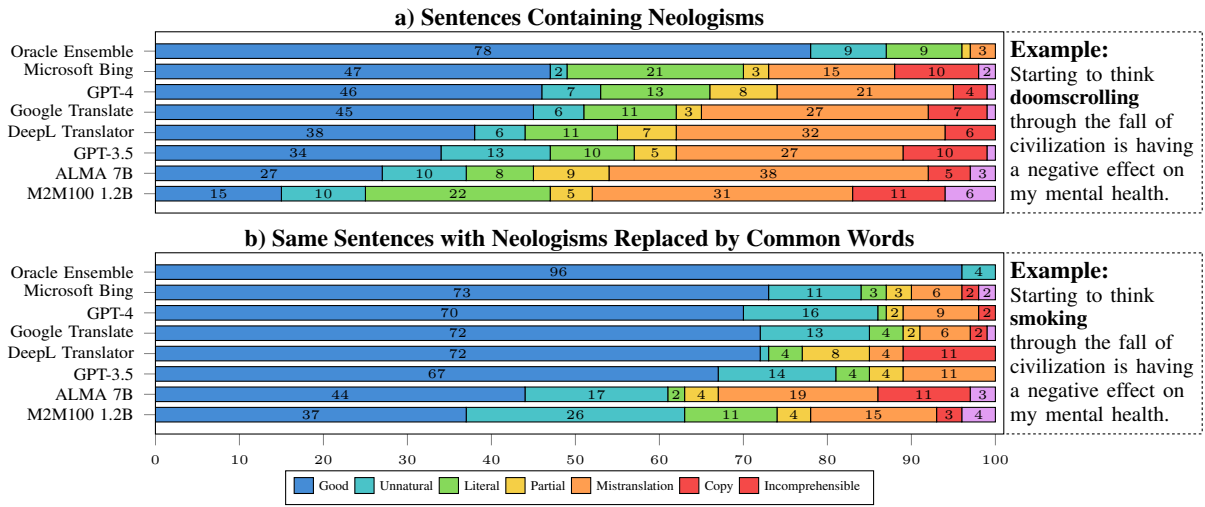



\noindent Neologisms -- recent word forms representing a new meaning, sense, or connotation \cite{cartier-2017-neoveille} -- consistently surface as language changes. Neologisms emerge to describe the ever-changing state of the world, such as new terms created during the COVID-19 pandemic. While humans easily adapt to language change, large language models (LLMs) struggle with the misalignment of training data and new test data distributions \cite{luu-etal-2022-time}.


Prior work on temporal language change \cite{https://doi.org/10.48550/arxiv.2102.01951, onoe-etal-2022-entity, luu-etal-2022-time} observed model degradation when finetuning on older text and evaluating on newer data and named entities \cite{rijhwani-preotiuc-pietro-2020-temporally,agarwal2022temporal,liu-ritter-2023-conll}. However, as far as we are aware there has not been prior work that analyzes the robustness of LLMs on handling neologisms. We show that adding a neologism to text decreases machine translation quality by an average of 43\% in a human evaluation (\S \ref{sec:motivation}), even for popular words emerging before 2020. 

In this paper, we present {\sc Neo-Bench}, a new benchmark designed to test the ability of LLMs to understand and process neologisms. We combine multiple methods and online text corpora to collect a diverse set of 2,505 neologisms based on the linguistic taxonomy devised by \citet{pinter-etal-2020-nytwit}: (i) \textbf{lexical neologisms} -- words representing new concepts, e.g., \textit{``long covid''}; (ii) \textbf{morphological neologisms} -- blends of existing subwords, e.g., \textit{``doomscrolling''}; and (iii) \textbf{semantic neologisms} -- existing words that convey a new meaning or sense, e.g., \textit{``ice''} (a term that refers to petrol- or diesel-powered cars taking electric car charging spots). We estimate word prevalence over time with Google Trends to obtain trending neologisms. We also create 4 benchmark tasks to evaluate the impact of neologisms on LLMs with Perplexity, Cloze Question Answering, Definition Generation, and Machine Translation. 



We show that lower neologism perplexities correlate with higher downstream task performance.
Older LLMs -- BART, T5, GPT-J, and Flan-T5 -- perform much worse with an average of 32.20\% and 12.27\% accuracy in question answering and definition generation, respectively. We also find that automatic metrics do not accurately measure the quality of translated sentences containing neologisms, evidenced by Spearman's $\rho$ rank correlation between COMETKiwi (a state-of-the-art metric) and human judgment, which is 0.491. This is lower than the average $\rho$ of 0.629 for COMETKiwi across 5 language pairs reported in the WMT23 Quality Estimation task \cite{blain-etal-2023-findings}. LLM performance in {\sc Neo-Bench} also differs based on a word's linguistic type, as lexical neologisms without derivations yield the highest perplexities and the most fragmented subword tokenization, while semantic neologisms that repurpose existing words result in literal definitions and translations.

{\sc Neo-Bench} evaluates a diverse set of LLM capabilities on handling neologisms in various tasks. Models must also understand compositionality for morphological neologisms, differentiate between word senses for semantic neologisms, and handle different contexts for lexical neologisms.

\section{Motivation } \label{sec:motivation}

We start by using machine translation as an example to illustrate the significant challenge neologisms pose on state-of-the-art NLP systems. We manually collect 100 neologism words with sentential context from social media, news articles, and dictionaries. GPT-3.5, GPT-4 and commercial translation systems, e.g., Google Translate,\footnote{\href{https://translate.google.com/}{https://translate.google.com/}} Microsoft Bing,\footnote{\href{https://www.bing.com/translator}{https://www.bing.com/translator}} and DeepL Translator,\footnote{\href{https://www.deepl.com/translator}{https://www.deepl.com/translator}} only managed to correctly translate about 34-47\% of these 100 sentences that contain neologisms based on our manual inspection (Figure \ref{fig:mt_histogram_full}; from English to Chinese). In stark contrast, when replacing the neologism with a common word in these sentences, the percentage of correct translations rises substantially to 67-73\%. We observe similar trends in open-source translation models, such as ALMA \cite{xu2023paradigm} and M2M100 \cite{DBLP:journals/corr/abs-2010-11125}. 


One thing to note is that these replacement words are not exact synonyms, but words that have been carefully chosen to create a near-identical, semantically plausible sentence; because new words emerge in areas not occupied by existing words \cite{ryskina-etal-2020-new}, true synonyms would often be verbose and incompatible with the sentence context.  Because the original sentences containing neologisms were collected in the wild, one might assume they would be even more natural in comparison to their modified counterparts, but yet, \textbf{there is a large gap in translation quality between neologism and non-neologism words for all models}.

A closer look reveals that six typical types of errors are made in mistranslated model outputs, which include (ordered by severity): 

\vspace{-5pt}
\begin{itemize}
\itemsep-.35em
\item \textbf{Unnatural}: Imperfect translation of the sentence due to grammatical errors;
\item \textbf{Literal}: Inaccurate output that literally translates the neologism or remaining sequence; 
\item \textbf{Partial}: Part of the sentence is untranslated and left out of the output;
\item \textbf{Mistranslation}: Incorrectly translated sentence portion leads to a poor understanding of the overall sentence meaning;
\item \textbf{Copy}: Part of the output is not translated and copied from the English input;
\item \textbf{Incomprehensible}: Incoherent output that fails to capture any original sentence meaning;
\end{itemize}
\vspace{-3pt}

\noindent Table \ref{tab:translation_examples} in the Appendix shows translations for each error type. The most common errors are mistranslations and literal translations with an average of 27.3\% and 13.7\% respectively. Model output for non-neologism sentences is more likely to have minor errors and be labeled unnatural by annotators. 

\begin{figure}[t!]
\resizebox{0.47\textwidth}{!}{
\hspace{-10pt}
\begin{tikzpicture}
\pgfplotsset{
compat=newest,
every axis plot/.append style={no marks,thick},
every axis/.style={
  width=5.5cm,
  height=2.3cm,
  }
}
\begin{axis}[
    name=bottomplot,
    ymin=3,
    ymax=80,
    extra y ticks={72},
    extra tick style={grid=major,major grid style={color1bg,very thick}},
    extra y tick labels = {},
    date coordinates in=x,
    table/col sep=comma,
    legend style={at={(0.98,0.42)}},
    xtick pos=lower, ytick pos=left,
    label style={inner sep=0pt}, 
    axis x line*=bottom,
    axis y line*=left,
     y post scale=1.9,
    xtick={2020-01-01, 2021-01-01, 2022-01-01,2023-01-01, 2024-01-01},
    xticklabel style={
        rotate=0,
        anchor=center,
    },
    xmin=2019-12-15,
    xmax=2023-10-01,
    xticklabel=\empty,
    extra x tick labels = {},
    grid=major,
    grid style={dashed,gray!30},
    label style={font=\small},
    y tick label style={font=\scriptsize},
    legend style={font=\small},
    xticklabel style={yshift=-12pt, font=\small},
    yticklabel={$\pgfmathprintnumber{\tick}\%$},
    y label style={font=\small},
    title=\textbf{Google Translate},
    title style={font=\small, at={(0.5, 0.82)}}
    ]
\addplot[color1bg,very thick,mark=square*, draw opacity=1.0] table [y= Google,x=Time]{good_translations.csv};
\addplot[color5bg,very thick,mark=square*, draw opacity = 1.0] table [y=Google,x=Time]{mistranslations.csv};
\end{axis}
\hspace{10pt}
\begin{axis}[
    name=bottomplot1,
    at = {($(bottomplot.east)$)},
    anchor=west,
    ymin=3,
    ymax=80,
    extra y ticks={73},
    extra tick style={grid=major,major grid style={color1bg,very thick}},
    extra y tick labels = {},
    date coordinates in=x,
    table/col sep=comma,
    legend style={at={(0.98,0.42)}},
    xtick pos=lower, ytick pos=left,
    label style={inner sep=0pt}, 
    axis x line*=bottom,
    axis y line*=left,
     y post scale=1.9,
    xtick={2020-01-01, 2021-01-01, 2022-01-01,2023-01-01, 2024-01-01},
    xticklabel style={
        rotate=0,
        anchor=center,
    },
    xmin=2019-12-15,
    xmax=2023-10-01,
    xticklabel=\empty,
    extra x tick labels = {},
    grid=major,
    grid style={dashed,gray!30},
    label style={font=\small},
    y tick label style={font=\small},
    legend style={font=\small},
    xticklabel style={yshift=-12pt, font=\small},
   yticklabel=\empty,
    y label style={font=\small},
    title=\textbf{Microsoft Bing},
    title style={font=\small, at={(0.5, 0.82)}}
    ]
\addplot[color1bg,very thick,mark=square*, draw opacity=1.0] table [y= Bing,x=Time]{good_translations.csv};
\addplot[color5bg,very thick,mark=square*, draw opacity = 1.0] table [y=Bing,x=Time]{mistranslations.csv};
\end{axis}
\end{tikzpicture}
}
\vspace*{-12pt}

\resizebox{0.47\textwidth}{!}{
\hspace{-10pt}
\begin{tikzpicture}
\pgfplotsset{
compat=newest,
every axis plot/.append style={no marks,thick},
every axis/.style={
  width=5.5cm,
  height=2.3cm,
  }
}
\begin{axis}[
    name=bottomplot,
    ymin=5,
    ymax=80,
    extra y ticks={72},
    extra tick style={grid=major,major grid style={color1bg,very thick}},
    extra y tick labels = {},
    date coordinates in=x,
    table/col sep=comma,
    legend style={at={(0.98,0.42)}},
    xtick pos=lower, ytick pos=left,
    label style={inner sep=0pt}, 
    axis x line*=bottom,
    axis y line*=left,
     y post scale=1.9,
    xtick={2020-01-01, 2021-01-01, 2022-01-01,2023-01-01, 2024-01-01},
    xticklabel style={
        rotate=0,
        anchor=center,
    },
    xmin=2019-12-15,
    xmax=2023-10-01,
    xticklabel=\empty,
    extra x tick labels = {},
    grid=major,
    grid style={dashed,gray!30},
    label style={font=\small},
    y tick label style={font=\scriptsize},
    legend style={font=\small},
    xticklabels={2020, 2021, 2022, 2023, 2024},
    xticklabel style={yshift=-5pt, font=\scriptsize},
    yticklabel={$\pgfmathprintnumber{\tick}\%$},
    y label style={font=\small},
    title=\textbf{DeepL Translator},
    title style={font=\small, at={(0.5, 0.82)}}
    ]
\addplot[color1bg,very thick,mark=square*, draw opacity=1.0] table [y= DeepL,x=Time]{good_translations.csv};
\addplot[color5bg,very thick,mark=square*, draw opacity = 1.0] table [y=DeepL,x=Time]{mistranslations.csv};
\end{axis}
\hspace{10pt}
\begin{axis}[
    name=bottomplot1,
    at = {($(bottomplot.east)$)},
    anchor=west,
    ymin=5,
    ymax=80,
    extra y ticks={70},
    extra tick style={grid=major,major grid style={color1bg,very thick}},
    extra y tick labels = {},
    date coordinates in=x,
    table/col sep=comma,
    legend style={at={(0.98,0.42)}},
    xtick pos=lower, ytick pos=left,
    label style={inner sep=0pt}, 
    axis x line*=bottom,
    axis y line*=left,
     y post scale=1.9,
    yticklabel=\empty,
    xtick={2020-01-01, 2021-01-01, 2022-01-01,2023-01-01, 2024-01-01},
    xticklabel style={
        rotate=0,
        anchor=center,
    },
    xmin=2019-12-15,
    xmax=2023-10-01,
    xticklabels={2020, 2021, 2022, 2023, 2024},
    extra x tick labels = {},
    grid=major,
    grid style={dashed,gray!30},
    label style={font=\small},
    legend style={font=\small},
    xticklabel style={yshift=-5pt, font=\scriptsize},
    title=\textbf{GPT-4},
    title style={font=\small, at={(0.5, 0.84)}},
    legend columns=-1,
    legend style={at={(0.5,-0.25)}, font=\scriptsize}
    ]
\addplot[color1bg,very thick,mark=square*, draw opacity=1.0] table [y= GPT-4,x=Time]{good_translations.csv};
\addplot[color5bg,very thick,mark=square*, draw opacity = 1.0] table [y=GPT-4,x=Time]{mistranslations.csv};

\legend{Good, Mistranslations}
\end{axis}
\end{tikzpicture}
}
\vspace{-8pt}
\caption{Percentage of good translations and mistranslations of neologism sentences over time. The dashed line represents the percentage of good translations achieved on non-neologism sentences. }
\label{figure:mt_time}
\end{figure}
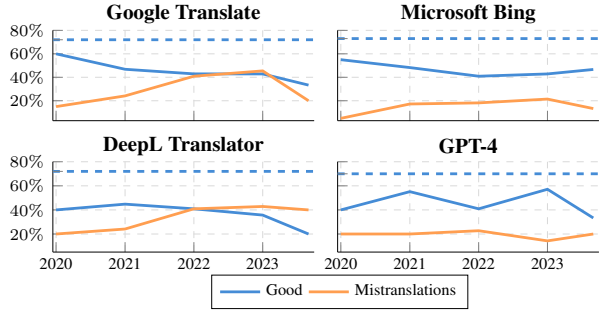

Another interesting observation is that newer neologisms indeed show lower rates of good translations and often higher rates of mistranslations, as one may expect. Figure \ref{figure:mt_time} shows the percentage of good translations and mistranslations over time for varied models. Compared to non-neologism sentences, models still yield lower rates of correct translations for neologisms that emerged before 2020. Many neologisms use existing words to convey meanings, such that the poor performance of models is not wholly explained by the absence of these word forms in training data. We propose a novel benchmark (\S \ref{sec:neo_bench}) to systematically study the impact of neologisms on LLMs (\S \ref{sec:main_results}). 

\section{{\sc Neo-Bench}: A Neologism Benchmark} \label{sec:neo_bench}

We create \textbf{{\sc Neo-Bench}}, a benchmark that consists of 2,505 neologisms (both words and phrases) that newly emerged around 2020--2023 and 4 intrinsic/extrinsic tasks (Table \ref{tab:benchmark_overview}) to evaluate LLMs' abilities to generalize on neologisms. To facilitate continuous research on neologisms and language change, we intend to periodically update {\sc Neo-Bench} with neologisms emerging after 2023. 

%

\subsection{Neologism Collection}\label{sec:neo_collect}
A neologism is a term that represents a new meaning or sense \cite{cartier-2017-neoveille}. Previous datasets \cite{mccrae-2019-identification, ryskina-etal-2020-new, DBLP:journals/corr/abs-2104-05010} only collected specific word types, ignored neologisms conveying new meanings with existing words, and did not utilize word prevalence trends (more in Related Work \S \ref{sec:related_work}). We design a more systematic collection process to quantify the effect of neologisms on a language's data distribution.


\begin{table}[t!]
\setlength{\tabcolsep}{4pt}
\centering
\resizebox{0.47\textwidth}{!}{%
\begin{tabular}{P{0.11\textwidth}|P{0.29\textwidth}M{0.11\textwidth}}
\toprule
\textbf{Task} & \textbf{Dataset} & \textbf{Evaluation}  \\
\midrule
Machine Translation & 240 sentences containing neologisms & BLEU, COMET \\\midrule
Perplexity Ranking & 422 Cloze passages with one-word answers & Word ranking \\\midrule
Cloze Questions & 750 Cloze passages with multiple choice answers & Accuracy \\\midrule
Definition Generation & 750 "What is [neologism]?" questions & Accuracy \\
\bottomrule
\end{tabular}
}
\vspace{-.1in}
\caption{Summary of datasets in {\sc {\sc Neo-Bench}}. }
\label{tab:benchmark_overview} 
\end{table}



\begin{figure}[!t]
    \centering
  \begin{subfigure}[b]{0.47\textwidth}
    \includegraphics[width=\textwidth, height=5.4cm]{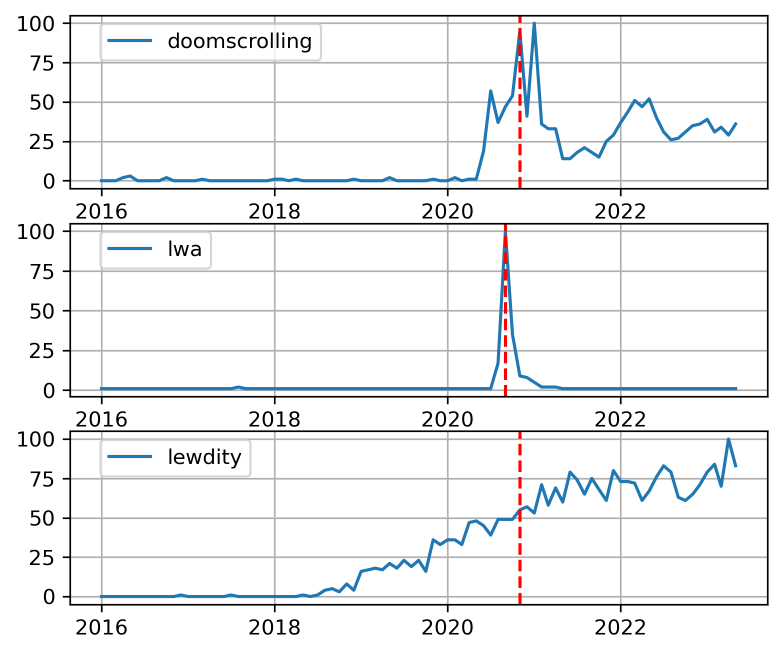}
    \label{fig:f1}
  \end{subfigure}
  \vspace*{-8mm}
  \caption{Example Google Trend lines measuring neologism prevalence. The dashed line estimates the date a neologism becomes popular while not yet conventional.}
    \label{fig:data_collection}
\end{figure}

\vspace{2pt}\noindent\textbf{Filtering Reddit Data based on Google Trends (Method 1).} New words commonly propagate in online communities \cite{DBLP:journals/corr/abs-2104-05010}, thus, we count word frequencies in monthly Reddit data to find single-word neologism candidates. We set a frequency cutoff between 50 and 100 per month to obtain uncommon words and remove misspellings and named entities using SpaCy \cite{ines_montani_2023_10009823}, resulting in 74,542 candidates. We further obtain word search frequencies from 2010 to 2023 on Google Trends\footnote{\url{https://trends.google.com/trends/}} and automatically filter out 87.13\% of neologism candidates based on these trend lines (see Figure \ref{fig:data_collection} for examples) by a combination of curve fitting, argmax detection, and integrals over time. Appendix \S\ref{sec:trends_method_explained} provides more details about trend filtering. From the set of 9,590 remaining candidates, we find that 10.48\% are prevalent neologisms by manual inspection. In total, we collected 1,005 neologism words from Reddit (310 lexical, 588 morphological, and 107 semantic neologisms). 

\vspace{2pt}\noindent\textbf{Retrieving News Articles about Neologisms (Method 2).} As Method 1 is only good at finding single-word neologisms, we turn to news articles that explain the meanings of neologisms to collect multi-word expressions. We first manually get 100 neologisms from news articles, recording news headlines of neologisms. Then, based on the shared text patterns of headlines, we created 16 headline \textbf{templates} (e.g., \textit{``\_\_\_: What is it?''}) to retrieve Google News articles from 2019 to 2023. Using SpaCy, we identify 60,671 noun and verb phrases with a Part-of-Speech tagger and remove duplicates and named entities. We used the same aforementioned filtering method for these phrases using Google Trends. From the remaining 8,039  candidates, we manually extracted 1,100 neologisms (778 lexical, 222 morphological, and 100 semantic neologisms), of which 713 are multiwords. 


\vspace{2pt}\noindent\textbf{Sampling Existing Neologism Datasets (Method 3).} 
To supplement our dataset with additional neologisms, we also sample from two existing open-source resources that contain a lot of rare words, many of which have no Google Trends data available. The NYT First Said Twitter bot (\texttt{@}NYT\_first\_said) tweets out words when they are used for the first time in New York Times articles by using exclusion lists. We retrieve 1,100 of its tweets from 2020 to collect 200 derived neologisms (192 morphological and 8 semantic). We also sample 1,400 entries from another noisy, automatically constructed, dataset of 80,071 new slang dictionary entries \cite{DBLP:journals/corr/abs-2104-05010}. We manually filter the sample and collect 200 derived neologisms (4 lexical, 194 morphological, and 2 semantic).



\vspace{2pt}\noindent\textbf{Overall.} We collected 2,505 neologism words. While semantic neologisms are infrequent in all sources, Google Trends data enables the collection of them, as these words change in baseline prevalence when a new sense is being popularized. Only 5.04\% of words from Reddit, 1.12\% of phrases from news articles, and 3.09\% of entries from previous datasets overlap with candidates from the other two sources --- highlighting the importance of using multiple diverse data sources and methods for neologism collection. We also verified that 44.23\% of these 2,505 words actually appear in the Urban Dictionary, a crowdsourced English-language online dictionary for slang words and phrases.





\begin{table}[t]
\small
\centering

\renewcommand{\arraystretch}{0.7}
\begin{tabular}{p{0.48\textwidth}}
\begin{tabular}{p{0.45\textwidth}}
\toprule
\multicolumn{1}{c}{\fontsize{8}{9.5}\selectfont \textbf{Neologism: \textit{doomscrolling} }}\\\toprule
{\fontsize{8}{8}\it The silver lining of this website no longer functioning as an even vaguely reliable information source is that \_\_\_ has basically been completely undermined. It wouldn't even work now since everything is too geared to outrage clickbait and actual reporting has disappeared, so there is no point staying on the app.}\vspace{4pt}\\
\begin{tabular}{P{0.02cm}P{0.48\textwidth}}
{\hspace{0.5pt}} & {\fontsize{8}{8}\it \selectfont a) misinformation \hspace{21pt} b) surfing} \vspace{2pt}\\
{\hspace{0.5pt}} & {\fontsize{8}{8}\it \selectfont \textbf{c) doomscrolling} \hspace{24pt} d) lying}\vspace{2pt} \\
{\hspace{0.5pt}} & {\fontsize{8}{8}\it \selectfont e) gaming \hspace{47pt} \textbf{f) anti-productivity (distractor)}} 
\end{tabular}\\\bottomrule
\end{tabular}
\end{tabular}
\vspace*{-2mm}
\caption{Example passage in {\sc Neo-Bench} for multiple-choice Cloze Question Answering with correct neologism answers and partially correct distractor answers.}
\label{tab:input_examples}
\end{table}

\begin{table*}[t]
\begin{CJK*}{UTF8}{gbsn}
\small
\centering
\renewcommand{\arraystretch}{0.7}
\setlength{\tabcolsep}{4pt}
\begin{tabular}{lP{0.84\textwidth}}
\toprule
\multicolumn{2}{c}{\fontsize{9}{10.5}\selectfont \textbf{a) Definition Generation Output Examples}}\vspace{2pt}\\\toprule
\multirow{6}{*}{\textbf{Stablecoin}} & {\fontsize{8}{8}\selectfont  {\textbf{Reference Definition:} \it A stablecoin is a type of cryptocurrency where the value of the digital asset is supposed to be pegged to a reference asset, which is either fiat money, exchange-traded commodities, or another cryptocurrency.}} \\\cmidrule{2-2}
& {\fontsize{8}{8}\selectfont \textbf{Model Output} (\hspace{.083333em}\sethlcolor{lightblue}\textbf{\textcolor{black}{\hl{Correct}}\hspace{.083333em}): \it Stablecoins are cryptocurrencies designed to maintain a stable value, typically by pegging their value to a specific asset or basket of assets, such as the US dollar, gold, or a combination of assets.}} \vspace{2pt}\\\toprule
\multirow{3}{*}{\textbf{Angel Shot}} &  {\fontsize{8}{8}\selectfont  {\textbf{Reference Definition:} \it An angel shot is a code to inform a bartender that a customer is not safe and needs assistance. }} \\\cmidrule{2-2}
 & {\fontsize{8}{8}\selectfont \textbf{Model Output} (\hspace{.083333em}\sethlcolor{pink}\textbf{\textcolor{black}{\hl{Incorrect}}\hspace{.083333em}): \it An angel shot is a cocktail made with whiskey and cream, served in a shot glass.}} \vspace{3pt}\\\toprule
 \multicolumn{2}{c}{\fontsize{9}{10.5}\selectfont \textbf{b) Machine Translation Output Examples}}\vspace{2pt}\\\toprule
\multirow{8}{*}{\textbf{Longcovid}} & {\fontsize{8}{8}\selectfont  {\textbf{Input:} Each reinfection increases the risk of \textbf{longcovid}, hospitalization, \& death. }}\\\cmidrule{2-2}
& {\fontsize{8}{8}\selectfont \textbf{Model Output} (\hspace{.083333em}\sethlcolor{lightblue}\textbf{\textcolor{black}{\hl{Correct}}\hspace{.083333em}): 每次再感染都会增加\underline{长新冠}病毒、住院和死亡的风险。}} \vspace{3pt}\\
& \hspace{87pt} {\fontsize{7}{7.5}\selectfont \it (Every reinfection increases the risk of \textbf{long COVID}, hospitalization, and death.)}
\\\cmidrule{2-2}
& {\fontsize{8}{8}\selectfont \textbf{Human Translation:} 每一次新冠感染都会提高出\underline{现后遗症}、住院治疗，甚至死亡的风险。} \vspace{2pt}\\
& \hspace{70pt} {\fontsize{7}{7.5}\selectfont  \it (Each COVID-19 infection increases the risk of developing \textbf{sequelae}, hospitalization, and even death.) } \vspace{3pt}\\\toprule
\multirow{10}{*}{\textbf{Doomscrolling}} & {\fontsize{8}{8}\selectfont  {\textbf{Input:} Starting to think \textbf{doomscrolling} through the fall of civilization is having a negative effect on my mental health. }}\\\cmidrule{2-2}
 & {\fontsize{8}{8}\selectfont \textbf{Model Output} (\hspace{.083333em}\sethlcolor{pink}\textbf{\textcolor{black}{\hl{Incorrect}}\hspace{.083333em}}): 开始认为在文明的衰落中\underline{滚动的厄运}对我的心理健康产生了负面影响。} \vspace{3pt}\\
& \hspace{91pt} {\fontsize{7}{7.5}\selectfont   \it (Start to think that the \textbf{doom rolling} in the decline of civilization is having a negative impact on my} \vspace{2pt}\\
& \hspace{91pt} {\fontsize{7}{7.5}\selectfont  \it mental health.) } \\\cmidrule{2-2}
& {\fontsize{8}{8}\selectfont \textbf{Human Translation:} 开始觉得, \underline{刷}关于文明衰败的\underline{负能量新闻}对我的心理健康产生了负面影响。} \vspace{2pt}\\
& \hspace{70pt} {\fontsize{7}{7.5}\selectfont  \it (Starting to feel that \textbf{scrolling} through negative news about the decline of civilization is having a negative impact} \vspace{2pt}\\
& \hspace{70pt} {\fontsize{7}{7.5}\selectfont  \it on my mental health.) } \\\bottomrule
\end{tabular}
\vspace*{-2mm}
\caption{Example model definitions and translations for {\sc Neo-Bench} tasks. \textit{``Doomscrolling''} is the act of spending an excessive amount of time reading negative news online. (English translations are shown for information only.)}
\vspace*{-4mm}
\label{tab:example_outputs_main_paper}
\end{CJK*}
\end{table*}

\subsection{Benchmark Tasks}\label{sec:benchmark_tasks}
{\sc Neo-Bench} consists of 4 tasks -- 3 downstream and 1 intrinsic metric -- to evaluate models' knowledge of neologisms: (i) Machine Translation with human and automatic evaluation; (ii) Cloze Question Answering to evaluate models in context; (iii) Definition Generation to evaluate models in a context-free setting; and (iv) perplexity to compare single-word neologisms to commonly used words. We describe the setup and result tables/figures in this section, then discuss the key findings based on these results more in-depth in \S \ref{sec:main_results}. 


\vspace{2pt}\noindent \textbf{Machine Translation (Task 1).} We sample from our collected neologisms (\S \ref{sec:neo_collect}) and search for reference sentences containing these words on social media and Google. We construct 240 sentences, including the 100 used in \S\ref{sec:motivation}. We work with in-house native speakers to create reference translations (English to Chinese) and evaluate system outputs in Table \ref{tab:mt_automatic} with BLEU \cite{papineni-etal-2002-bleu}, COMET \cite{rei2020comet}, and MetricX-23 (XXL and XL) \cite{juraska-etal-2023-metricx}. We also use the reference-free metrics COMETKiwi \cite{rei-etal-2022-cometkiwi} and MetricX-23-QE (XXL and XL). Lower MetricX scores indicate higher performance. We report the correlations of metrics with human ratings for the MT models in \S\ref{sec:motivation} using Spearman's $\rho$ in Table \ref{tab:mt_automatic}.

\begin{figure*}[!ht]
\pgfplotsset{
    testbar/.style={
        xbar stacked,
        bar width=7pt,
        xtick pos=lower, ytick pos=left,
        xmin=0,xmax=117,
        ytick = data,
        ytick distance=1,
        yticklabel style={text width=2.1cm, align=right, yshift=0cm, font=\scriptsize},
        ylabel style={font=\tiny},
        ytick style={yshift=0cm},
        legend style={at={(0.3,0.05)},anchor=west, draw=none, fill=none},
        tick align = outside,
        height=0.425\textwidth,
        /pgf/number format/.cd,
        precision=2,
        xlabel = Accuracy,
        xlabel shift = -3pt,
        xlabel style = {font=\small},
        y=3.5mm,
        enlarge y limits={abs=0.7},
        nodes near coords,
        nodes near coords style={/pgf/number format/.cd,precision=2},
        show sum on top/.style={
            /pgfplots/scatter/@post marker code/.append code={%
                \node[
                    at={(normalized axis cs:%
                            \pgfkeysvalueof{/data point/x},%
                            \pgfkeysvalueof{/data point/y})%
                    },
                    anchor=west,
                ]
                {\pgfmathprintnumber[precision=2, fixed zerofill]{\pgfkeysvalueof{/data point/x}}};
            }
        },
    }}
\begin{tikzpicture}
\tikzstyle{every node}=[font=\scriptsize]
\begin{axis}[testbar, name=plot1, xtick=\empty, xlabel=\empty, ylabel = \textbf{Instruction models}, ylabel style={font=\footnotesize}, anchor=north west, yticklabels = {Flan-T5 Large, Alpaca 7B, OLMo Instruct 7B, Mistral Instruct 7B, LLaMA-2 Chat 7B, LLaMA-2 Chat 13B, LLaMA-2 Chat 70B, GPT-4}, legend style={at={(0.68,0.08), draw=none},anchor=west}, title=\textbf{Cloze Question Answering}, title style={at={(0.5,1.10)},font=\small, anchor=north}, nodes near coords style={/pgf/number format/.cd,fixed zerofill,precision=2},
] 
 \addplot[fill=clozecolor2] coordinates{
                                    (42.40, 0)
                                    (47.01, 1)
                                    (54.16, 2)
                                    (58.24, 3)
                                    (60.28, 4)
                                    (64.67, 5)
                                    (68.27, 6)
                                    (82.86, 7)};
 \addplot[fill=clozecolor1, show sum on top] coordinates{
                                   (39.84, 0)
                                   (32.99, 1)
                                   (33.25, 2)
                                   (30.74, 3)
                                   (27.28, 4)
                                   (28.00, 5)
                                   (27.20, 6)
                                   (16.60, 7)};
\end{axis}

\hspace{6pt}
\begin{axis}[testbar, name=plot3,  xtick=\empty, at = {($(plot1.south east)$)}, 
ytick=\empty, yticklabels = {Flan-T5 Large, Alpaca 7B, OLMo Instruct 7B, Mistral Instruct 7B, LLaMA-2 Chat 7B, LLaMA-2 Chat 13B, LLaMA-2 Chat 70B, GPT-4}, 
legend style={at={(0.64,0.12)},anchor=west}, 
title=\textbf{Definition Generation},  xlabel=\empty,
title style={at={(0.5,1.10)},font=\small, anchor=north,yshift=-0.1},  xmax=102, nodes near coords style={/pgf/number format/.cd,fixed zerofill,precision=2},
] 
 \addplot[nodes near coords, fill=clozecolor3] coordinates{
                                    (14.93,0)
                                    (50.13, 1)
                                    (60.40, 2)
                                    (59.07, 3)
                                    (57.34, 4)
                                    (58.27, 5)
                                    (59.87, 6)
                                    (83.5, 7)};
\end{axis}
\end{tikzpicture}

\vspace{-4.5pt}

\begin{tikzpicture}
\tikzstyle{every node}=[font=\scriptsize]
\begin{axis}[testbar, name=plot2, at = {($(plot1.south west)$)}, anchor=north west, yticklabels = {BART Large, T5 Large, GPT-J 6B, LLaMA-1 7B, OLMo 7B, Mistral 7B, LLaMA-2 7B, LLaMA-2 13B, LLaMA-2 70B, GPT-3.5}, legend style={at={(0.52,0.06), font=\footnotesize, draw=none, fill=none},font=\footnotesize,anchor=west}, legend columns=-1,
ylabel = \textbf{Pre-trained models}, ylabel style ={font=\footnotesize}, nodes near coords style={/pgf/number format/.cd,fixed zerofill,precision=2},
] 
 \addplot[fill=clozecolor2] coordinates{
                                    (18.67, 0)
                                    (37.07, 1)
                                    (30.67, 2)
                                    (28.8, 3)
                                    (23.33, 4)
                                    (67.87, 5)
                                    (62.67, 6)
                                    (64.93, 7)
                                    (74.93, 8)
                                    (70.68, 9)};
 \addplot[fill=clozecolor1, show sum on top] coordinates{
                                   (14.27,0)
                                   (13.60, 1) 
                                   (51.87, 2)
                                   (30.5, 3)
                                   (27.60, 4)
                                   (27.07, 5)
                                   (26.8, 6)
                                   (27.2, 7)
                                   (22.93, 8)
                                   (25.97, 9)};
                                   \legend{Neologism,Distractor}
\end{axis}
\hspace{6pt}

\begin{axis}[name=plot4, testbar, at = {($(plot3.south west)$)}, xlabel style={font=\small}, anchor=north west, ytick=\empty, xmax=90,
yticklabels = {GPT-J 6B, LLaMA-1 7B, LLaMA-2 7B, LLaMA-2 13B, LLaMA-2 70B, GPT-3.5}, legend style={at={(0.68,0.08)},anchor=west}, nodes near coords style={/pgf/number format/.cd,fixed zerofill,precision=2},
] 
 \addplot[nodes near coords, fill=clozecolor3] coordinates{
                                    (0, 0)
                                    (0, 1)
                                    (34.13, 2)
                                    (45.60, 3)
                                    (60.40, 4)
                                    (62.67, 5)
                                    (52.67, 6) 
                                    (54.27, 7)
                                    (62.00, 8)
                                    (74.5, 9)};
\node[right] at (-10,-5) {$0.00$};
\node[right] at (-10,90) {$0.00$};
\end{axis}
\end{tikzpicture}
\vspace*{-3mm}
\caption{\textbf{Left}: Results of the Cloze Question Answering task reported by accuracy of selecting the neologism or distractor option. Combined accuracy for selecting either answer is provided.  \textbf{Right}: Results of the Definition Generation task reported with accuracy of correct definitions. 5-shot prompting of models is used for both tasks.}
\vspace{-4mm}
\label{tab:cloze_and_definition_results}
\end{figure*}
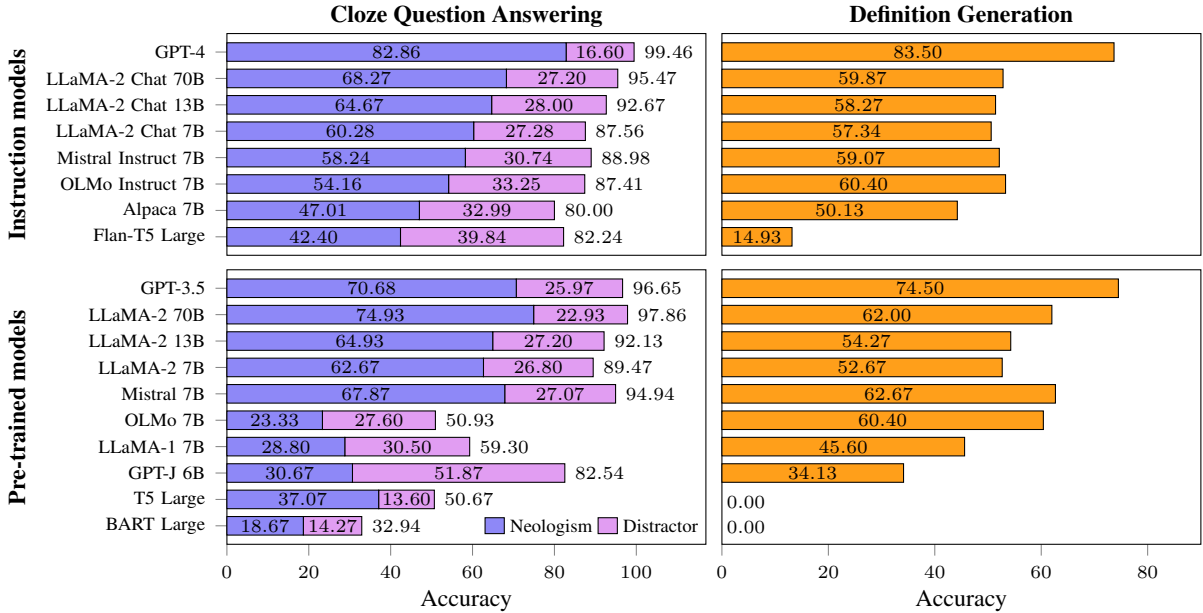

\vspace{2pt}\noindent\textbf{Cloze Question Answering (Task 2).} We sample 750 neologisms to create text passages, where one sentence contains a neologism and the remaining passage serves as preceding or following context (see example in Table \ref{tab:input_examples}). We mask out the neologism and provide four incorrect answers plus one distractor answer, which is a common word or phrase that is feasible in context. We evaluate BART-large \cite{lewis2019bart}, T5-Large \cite{JMLR:v21:20-074}, Flan-T5-Large \cite{chung2022scaling}, GPT-J 6B \cite{gpt-j}, LLaMA-1 7B \cite{touvron2023llama}, Alpaca 7B \cite{alpaca}, LLaMA-2, LLaMA-2-Chat \cite{touvron2023llama2}, OLMo-7B \cite{groeneveld2024olmo}, OLMo-7B-Instruct, Mistral-7B \cite{jiang2023mistral}, Mistral-7B-Instruct, GPT 3.5 \cite{brown2020language}, and GPT-4 in multiple-choice Cloze Question Answering (QA). We experiment with 5-shot prompting and test three sizes of LLaMA-2 models. We show results in Figure \ref{tab:cloze_and_definition_results} with the stratified and combined accuracies of selecting either the neologism or distractor answer.
\begin{table}[t]
\centering
\setlength{\tabcolsep}{4pt}
\scriptsize
\renewcommand{\arraystretch}{0.85}
\resizebox{0.45\textwidth}{!}{
\begin{tabular}{l|rrr}
\toprule
Label & \multicolumn{1}{c}{Complete} & \multicolumn{1}{c}{Partial} & \multicolumn{1}{c}{Unknown} \\\midrule
Good & \textbf{53.13\%} & \textbf{34.78\%} & \textbf{30.77\%}  \\
Unnatural & 9.38\% & 4.35\% & 0.00\% \\
Literal & 10.94\% & 17.39\% & 15.38\% \\
Partial & 4.69\% & 13.04\% & 15.38\% \\
Mistranslation & \textbf{20.31}\% & \textbf{21.74\%} & \textbf{23.09\%} \\
Copy & 1.55\% & 4.35\% & 15.38\% \\
Incomprehensible & 0.00\% & 4.35\% & 0.00\%  \\\midrule
Total & \multicolumn{1}{c}{64} & \multicolumn{1}{c}{23} & \multicolumn{1}{c}{13}\\
\bottomrule
\end{tabular}
}
\vspace{-2pt}
\caption{GPT-4's understanding of a neologism does not result in high machine translation performance. GPT-4 MT output is separated by its performance in Cloze QA, Definition Generation, and Definition Prompting. GPT-4 shows full, partial, and no knowledge of a neologism if zero, one, or multiple tasks are incorrect, respectively.
}\label{tab:gpt_4_finegrained} 
\end{table}

\noindent\textbf{Open-ended Definition Generation (Task 3).} We evaluate the same models from Task 2 for their context-free knowledge of 750 neologisms with question prompts (i.e., \textit{``What is doomscrolling?''}) to obtain neologism definitions. We construct human reference definitions and use GPT-4 to evaluate if model generations are semantically equivalent to the gold reference. We use 5-shot prompting and report results with accuracy in Figure \ref{tab:cloze_and_definition_results}. Table \ref{tab:example_outputs_main_paper} shows example LLM-generated definitions.

\vspace{2pt}\noindent\textbf{Perplexity Rankings (Task 4).} Using 422 Cloze passages that have both singular distractor and neologism answers, we use perplexity to evaluate GPT-J 6B, LLaMA-1 7B, Alpaca 7B, LLaMA-2 7B, LLaMA-2 Chat 7B, OLMo-7B, OLMo-7B-Instruct, Mistral-7B, and Mistral-7B-Instruct. For each passage, we use rank classification \cite{brown2020language}, where we fill in the mask with the neologism and measure the perplexity of the passage. We replace the mask with the distractor answer and the top 5000 singular words from Reddit by frequency and measure perplexities of all 5002 sequences. The mask-filling words are sorted by the corresponding sequence perplexity, and the average rankings of neologisms and distractors are reported in Figure \ref{fig:ranking_results}. Lower neologism rankings represent lower relative perplexities and show that models are likely to complete the passage with a neologism.

\begin{table*}[t]
\centering
\setlength{\tabcolsep}{3pt}
\scriptsize
\renewcommand{\arraystretch}{0.9}
\resizebox{0.92\textwidth}{!}{
\begin{tabular}{lr|cccc|ccc}
\toprule
\multicolumn{2}{l|}{\multirow{2}{*}{\textbf{Model} (human rank)}} & \multicolumn{4}{c|}{\textbf{Reference-Based Metrics}} & \multicolumn{3}{c}{\textbf{Reference-Free Metrics}} \vspace{1pt}\\
& & {\fontsize{6}{6}\selectfont \textbf{BLEU}$\uparrow$} & {\fontsize{6}{6}\selectfont \textbf{COMET}$\uparrow$} & {\fontsize{6}{6}\selectfont \textbf{MX-23}}{\fontsize{4}{4}\selectfont \textbf{XXL}}{\fontsize{6}{6}\selectfont $\downarrow$} & {\fontsize{6}{6}\selectfont \textbf{MX-23}}{\fontsize{4}{4}\selectfont \textbf{XL}}{\fontsize{6}{6}\selectfont $\downarrow$} & {\fontsize{6}{6}\selectfont \textbf{COMET}}{\fontsize{4}{4}\selectfont \textbf{KIWI}}{\fontsize{6}{6}\selectfont $\uparrow$} & {\fontsize{6}{6}\selectfont \textbf{MX-QE}}{\fontsize{4}{4}\selectfont \textbf{XXL}}{\fontsize{6}{6}\selectfont $\downarrow$} & {\fontsize{6}{6}\selectfont \textbf{MX-QE}}{\fontsize{4}{4}\selectfont \textbf{XL}}{\fontsize{6}{6}\selectfont $\downarrow$} \\\midrule

Bing Translator & (1) & \cellcolor[HTML]{7DCEA0}0.452 (2)  & \cellcolor[HTML]{A9DFBF}0.825 (5) & \cellcolor[HTML]{F5B7B1}2.419 (6) & \cellcolor[HTML]{F5B7B1}2.343 (6) & \cellcolor[HTML]{FADBD8}0.788 (5) & \cellcolor[HTML]{FADBD8}1.679 (5)  & \cellcolor[HTML]{FADBD8}2.246 (5)  \\

GPT-4 & (2) & \cellcolor[HTML]{7DCEA0}0.446 (3)  & \cellcolor[HTML]{52BE80}\textbf{0.854} (1) & \cellcolor[HTML]{52BE80}\textbf{1.550} (1) & \cellcolor[HTML]{52BE80}\textbf{1.793} (1) & \cellcolor[HTML]{A9DFBF}0.793 (3) & \cellcolor[HTML]{7DCEA0}1.432 (3) & \cellcolor[HTML]{7DCEA0}2.089 (3) \\

Google Translate & (3) & \cellcolor[HTML]{52BE80}\textbf{0.507 (1)}  & \cellcolor[HTML]{52BE80}0.853 (2) &\cellcolor[HTML]{FADBD8}1.825 (4) & \cellcolor[HTML]{A9DFBF}1.945 (4) & \cellcolor[HTML]{7DCEA0}0.800 (2) & \cellcolor[HTML]{7DCEA0}1.429 (2) & \cellcolor[HTML]{52BE80}\textbf{1.940} (1) \\

DeepL Translator & (4) & \cellcolor[HTML]{A9DFBF}0.406 (4)  & \cellcolor[HTML]{7DCEA0}0.842 (3)  & \cellcolor[HTML]{7DCEA0}1.775 (3)  & \cellcolor[HTML]{7DCEA0}1.901 (3)  & \cellcolor[HTML]{52BE80}\textbf{0.807} (1) & \cellcolor[HTML]{52BE80}\textbf{1.260} (1) & \cellcolor[HTML]{52BE80}1.944 (2) \\

GPT-3.5 & (5) & \cellcolor[HTML]{FADBD8}0.399 (5) & \cellcolor[HTML]{7DCEA0}0.841 (4) & \cellcolor[HTML]{A9DFBF}1.705 (2) & \cellcolor[HTML]{52BE80}1.796 (2)  & \cellcolor[HTML]{A9DFBF}0.792 (4) & \cellcolor[HTML]{A9DFBF}1.467 (4) & \cellcolor[HTML]{A9DFBF}2.157 (4) \\

ALMA 7B (LLaMA-2) & (6) & \cellcolor[HTML]{F1948A}0.285 (7)  & \cellcolor[HTML]{F5B7B1}0.801 (6)  & \cellcolor[HTML]{F5B7B1}2.382 (5) & \cellcolor[HTML]{FADBD8}2.251 (5)  & \cellcolor[HTML]{F1948A}0.746 (6) & \cellcolor[HTML]{F5B7B1}2.038 (6) & \cellcolor[HTML]{F5B7B1}2.462 (6) \\



M2M100 1.2B & (7) &\cellcolor[HTML]{F5B7B1}0.337 (6) & \cellcolor[HTML]{F1948A}0.776 (7)  & \cellcolor[HTML]{F1948A}3.454 (7) & \cellcolor[HTML]{F1948A}3.142 (7) & \cellcolor[HTML]{F1948A}0.745 (7) & \cellcolor[HTML]{F1948A}2.821 (7) & \cellcolor[HTML]{F1948A}2.976 (7) \\\midrule
\multicolumn{2}{l|}{\textbf{Spearman's $\rho$}} & 0.244 & 0.445 & 0.457 & 0.380 & \textbf{0.491}  &  0.451 & 0.445 \\
\bottomrule
\end{tabular}}
\vspace*{-7pt}
\caption{Machine Translation models evaluated on neologisms with BLEU, COMETKiwi, COMET, \textbf{M}etric\textbf{X}-\textbf{23}, and \textbf{M}etric\textbf{X}-23-\textbf{QE}. We use the XXL and XL sizes for MetricX. Rankings of models are provided for metrics and human evaluation for models used in \S\ref{sec:motivation}. Spearman's $\rho$ between each metric and human evaluation is also reported.
}\label{tab:mt_automatic} 
\end{table*}

\section{Key Findings}\label{sec:main_results}
We utilize {\sc Neo-Bench} tasks to evaluate the ability of various LLMs to adapt to neologisms. The following are our key findings:
\vspace{2pt}

\noindent \textbf{Current automatic metrics cannot accurately evaluate MT models that struggle with neologisms.} In \S\ref{sec:motivation}, MT models decrease in performance by 43\% when translating neologisms with Bing being the best model based on human evaluation. However, COMET and COMETKiwi scores are notably high and MetricX-23 error scores are low for all models in Table \ref{tab:mt_automatic}. The best models are Google Translate for BLEU (0.507); DeepL for COMETKiwi (0.807) and MetricX-23-QE (1.260); and GPT-4 for COMET (0.854) and MetricX-23 (1.550), highlighting that automatic metrics show poor system-level correlations with human judgments. For sentence-wise correlation between MT metrics and human evaluations, the average Spearman's $\rho$ of BLEU, COMET, COMETKiwi, MetricX-23 and MetricX-23-QE (XXL) is 0.244, 0.445, 0.491, 0.457, 0.451, respectively. In contrast, COMETKiwi, our highest correlating metric, has an average $\rho$ of 0.629 for five language pairs on the WMT23 Quality Estimation task for direct assessment \cite{blain-etal-2023-findings}. From our reference sentences, translating neologisms often requires paraphrasing, resulting in low $\rho$ for BLEU.\vspace{2pt}

\noindent \textbf{GPT-4's knowledge of neologisms is task specific.} Table \ref{tab:gpt_4_finegrained} presents the human annotations of GPT-4 translations of neologisms, separated by the corresponding performance of neologisms in Cloze QA, Definition Generation, and Definition Prompting, where we ask GPT-4 if the provided human reference definitions of neologisms are correct. GPT-4 shows complete knowledge of a neologism if all tasks are correct; partial knowledge if one task is incorrect; and unknown if multiple tasks are incorrect. There are higher rates of correct translations for neologisms GPT-4 understands, as good translations drop by 20.3\% if GPT-4 does not fully know a neologism's meaning. However, the rate of mistranslations are constant regardless of GPT-4 performance in other tasks, and GPT-4 only correctly translates 53.13\% of neologisms it has complete knowledge of. GPT-4's knowledge of neologisms is compartmentalized and does not result in similarly high performance in machine translation compared to other Neo-Bench tasks, emphasizing the difficulty of translating neologisms. 

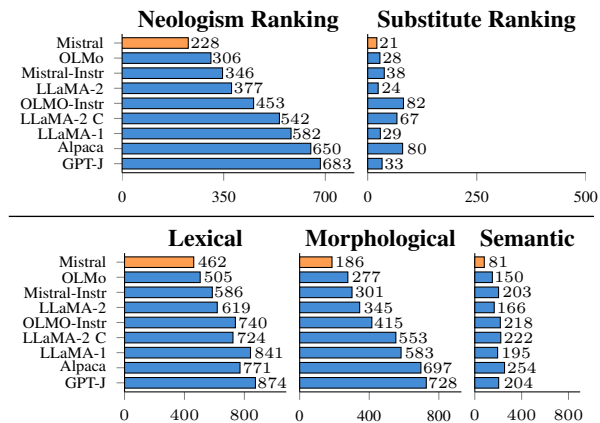
\begin{figure}[!t]
\pgfplotsset{
    testbar/.style={
        xbar stacked,
        bar width=4pt,
        xtick pos=lower, ytick pos=left,
        xmin=0,xmax=770,
        ytick = data,
        ytick distance=1,
        yticklabel style={text width=1.4cm, align=right, yshift=0cm, font=\tiny},
        ytick style={yshift=0cm},
        legend style={at={(0.3,0.05)},font=\tiny, anchor=west, draw=none, fill=none},
        tick align = outside,
        height=0.24\textwidth,
        axis x line*=bottom,
        axis y line*=left,
        xlabel style = {font=\small},
        major tick length =3pt,
        y=2mm,
        enlarge y limits={abs=0.6},
        }
    }
\hspace{-10pt}
\begin{tikzpicture}
\tikzstyle{every node}=[font=\tiny]
\begin{axis}[testbar, xmin=0, xmax=800, height=0.25\textwidth, xticklabels={0,350,700}, xtick={0, 350, 700}, name=plot1, xlabel=\empty, ylabel style={font=\scriptsize}, anchor=north west, yticklabels = {Mistral, OLMo, Mistral-Instr, LLaMA-2, OLMO-Instr, LLaMA-2 C, LLaMA-1, Alpaca, GPT-J}, yticklabel style={font=\tiny},legend style={at={(0.68,0.08), font=\tiny,draw=none},anchor=west}, title=\textbf{Neologism Ranking}, title style={at={(0.5,1.13)},font=\small, anchor=north},
] 
\node[right] at (200, 800) {$228$};
\node[right] at (275, 700) {$306$};
\node[right] at (320, 600) {$346$};
\node[right] at (350, 500) {$377$};
\node[right] at (425, 400) {$453$};
\node[right] at (510, 300) {$542$};
\node[right] at (550, 200) {$582$};
\node[right] at (620, 100) {$650$};
\node[right] at (655, 0) {$683$};

 \addplot[fill=color5bg] coordinates{
                                    (228, 8)
                                    (0, 7)
                                    (0, 6)
                                    (0, 5)
                                    (0, 4)
                                    (0, 3)
                                    (0, 2)
                                    (0, 1)
                                    (0, 0)
                                    };
                                
 \addplot[fill=color1bg] coordinates{
                                    (0, 8)
                                    (306, 7)
                                    (346, 6)
                                    (377, 5)
                                    (453, 4)
                                    (542, 3)
                                    (582, 2)
                                    (650, 1)
                                    (683, 0)};

\end{axis}

\hspace{5pt}
\begin{axis}[testbar, name=plot3,  at = {($(plot1.south east)$)}, 
ytick=\empty, yticklabels = {}, 
legend style={at={(0.64,0.12), draw=none},anchor=west}, 
title=\textbf{Substitute Ranking},  xlabel=\empty, xticklabels={0,250,500}, xtick={0, 250,500},
title style={at={(0.5,1.13)},font=\small, anchor=north,yshift=-0.1}, xmax=500,
]
\node[right] at (3, 800) {$21$};
\node[right] at (10, 700) {$28$};
\node[right] at (19, 600) {$38$};
\node[right] at (8, 500) {$24$};
\node[right] at (65, 400) {$82$};
\node[right] at (50, 300) {$67$};
\node[right] at (10, 200) {$29$};
\node[right] at (67, 100) {$80$};
\node[right] at (15, 0) {$33$};

 \addplot[fill=color5bg] coordinates{
                                    (21, 8)
                                    (0, 7)
                                    (0, 6)
                                    (0, 6)
                                    (0, 4)
                                    (0, 3)
                                    (0, 2)
                                    (0, 1)
                                    (0, 0)
                                    };
 \addplot[fill=color1bg] coordinates{
                                    (0, 8)
                                    (28, 7)
                                    (38, 6)
                                    (24, 5)
                                    (82, 4)
                                    (67, 3)
                                    (29, 2)
                                    (80, 1)
                                    (33, 0)
                                    };
\end{axis}
\end{tikzpicture}

\vspace{-19pt}
\rule[0ex]{\linewidth}{0.5pt}
\begin{tikzpicture}
\tikzstyle{every node}=[font=\tiny]
\begin{axis}[testbar, height=0.20\textwidth, xtick={0, 400, 800}, name=plot1, yticklabels = {Mistral, OLMo, Mistral-Instr, LLaMA-2, OLMO-Instr, LLaMA-2 C, LLaMA-1, Alpaca, GPT-J}, xlabel=\empty, ylabel style={font=\small}, anchor=north west, legend style={at={(0.68,0.08), draw=none},anchor=west}, title=\textbf{Lexical}, title style={at={(0.5,1.13)},font=\footnotesize, anchor=north}, xmin=0, xmax=1080,
] 
\node[right] at (42, 800) {$462$};
\node[right] at (46, 700) {$505$};
\node[right] at (53, 600) {$586$};
\node[right] at (58, 500) {$619$};
\node[right] at (69, 400) {$740$};
\node[right] at (68, 300) {$724$};
\node[right] at (80, 200) {$841$};
\node[right] at (72, 100) {$771$};
\node[right] at (82, 0) {$874$};
\hspace{-7pt}
 \addplot[fill=color5bg] coordinates{
                                    (462, 8)
                                    (0, 7)
                                    (0, 6)
                                    (0, 5)
                                    (0, 4)
                                    (0, 3)
                                    (0, 2)
                                    (0, 1)
                                    (0, 0)
                                    };
 \addplot[fill=color1bg] coordinates{
                                    (0, 8)
                                    (505, 7)
                                    (586, 6)
                                    (619, 5)
                                    (740, 4)
                                    (724, 3)
                                    (841, 2)
                                    (771, 1)
                                    (874, 0)};

\end{axis}

\hspace{5pt}
\begin{axis}[testbar, height=0.20\textwidth,name=plot3,  at = {($(plot1.south east)$)}, 
ytick=\empty, yticklabels = {}, 
legend style={at={(0.64,0.12), draw=none},anchor=west}, 
title=\textbf{Morphological},  xticklabels={0,400,800}, xtick={0, 400, 800},
title style={at={(0.5,1.13)},font=\footnotesize, anchor=north,yshift=-0.1},  xmin=0, xmax=930,
]
\node[right] at (145, 800) {$186$};
\node[right] at (240, 700) {$277$};
\node[right] at (260, 600) {$301$};
\node[right] at (310, 500) {$345$};
\node[right] at (370, 400) {$415$};
\node[right] at (510, 300) {$553$};
\node[right] at (545, 200) {$583$};
\node[right] at (650, 100) {$697$};
\node[right] at (680, 0) {$728$};

 \addplot[fill=color5bg] coordinates{
                                    (186, 8)
                                    (0, 7)
                                    (0, 6)
                                    (0, 5)
                                    (0, 4)
                                    (0, 3)
                                    (0, 2)
                                    (0, 1)
                                    (0, 0)
                                    };
 \addplot[fill=color1bg] coordinates{
                                    (0, 8)
                                    (277, 7)
                                    (301, 6)
                                    (345, 5)
                                    (415,4)
                                    (553, 3)
                                    (583, 2)
                                    (697, 1)
                                    (728, 0)};
\end{axis}

\hspace{5pt}
\begin{axis}[testbar, name=plot4, height=0.16\textwidth, at = {($(plot3.south east)$)}, 
ytick=\empty, yticklabels = {}, 
legend style={at={(0.64,0.12), draw=none},anchor=west}, 
title=\textbf{Semantic},  xlabel=\empty, xtick={0, 400, 800},
title style={at={(0.5,1.13)},font=\footnotesize, anchor=north,yshift=-0.1}, xmin=0, xmax=900,
]
\node[right] at (30, 800) {$81$};
\node[right] at (80, 700) {$150$};
\node[right] at (150, 600) {$203$};
\node[right] at (100, 500) {$166$};
\node[right] at (160, 400) {$218$};
\node[right] at (160, 300) {$222$};
\node[right] at (140, 200) {$195$};
\node[right] at (190, 100) {$254$};
\node[right] at (160, 0) {$204$};
 \addplot[fill=color5bg] coordinates{
                                    (81, 8)
                                    (0, 7)
                                    (0, 6)
                                    (0, 5)
                                    (0, 4)
                                    (0, 3)
                                    (0, 2)
                                    (0, 1)
                                    (0, 0)
                                    };
 \addplot[fill=color1bg] coordinates{
                                    (0, 8)
                                    (150, 7)
                                    (203, 6)
                                    (166, 5)
                                    (218, 4)
                                    (222, 3)
                                    (195, 2)
                                    (254, 1)
                                    (204, 0)};
\end{axis}
\end{tikzpicture}
\vspace*{-6mm}
\caption{Average rankings of neologisms and pre-existing substitute terms compared to 5000 common words, sorted by model perplexities of texts filled in with each word. Neologisms are separated by linguistic type: lexical, morphological, and semantic.}
\label{fig:ranking_results}
\end{figure}

\noindent \textbf{Models perform worse on neologisms than pre-existing words.} For Cloze questions in Figure \ref{tab:cloze_and_definition_results}, neologism answers are designed to be more natural as the original passages contained these neologisms, yet all models select a large portion (27.99\% on average) of distractor answers. Neologisms also have an average perplexity rank of 463 compared to distractor rankings of 45 in Figure \ref{fig:ranking_results}, indicating much lower perplexity for pre-existing words.


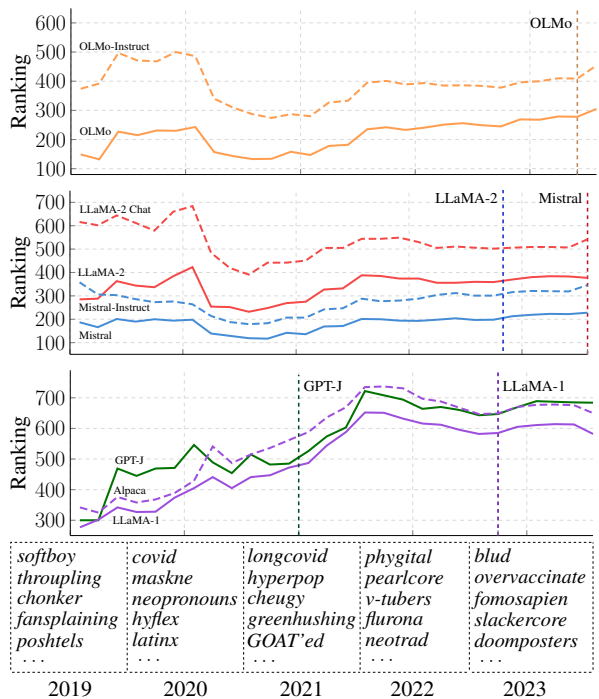
\begin{figure}[t!]
\resizebox{0.485\textwidth}{!}{
\hspace{-2.5pt}
\begin{tikzpicture}
\pgfplotsset{
compat=newest,
every axis plot/.append style={no marks,thick},
every axis/.style={
  width=16cm,
  height=4cm,
  }
}
\begin{axis}[
    ymin=80,
    ymax=650,
    date coordinates in=x,
    table/col sep=comma,
    legend style={at={(0.98,0.3)}},
    xtick pos=lower, ytick pos=left,
    label style={inner sep=0pt}, 
    axis x line*=bottom,
    axis y line*=left,
     y post scale=1.9,
    xtick={2019-01-01, 2020-01-01, 2021-01-01, 2022-01-01,2023-01-01, 2024-01-01},
    xticklabel style={
        rotate=45,
        anchor=east,
    },
    ytick={100,200,300,400,500,600},
    xmin=2019-1-01,
    xmax=2023-08-01,
    xticklabels=\empty,
    extra x ticks={2023-06-01},
    extra tick style={grid=major,major grid style={orange1,very thick}},
    extra x tick labels = {},
    grid=major,
    grid style={dashed,gray!30},
    label style={font=\LARGE},
    y tick label style={font=\LARGE},
    legend style={font=\normalsize},
    xticklabel style={yshift=-3pt},
    ylabel=Ranking,
    y label style={font=\LARGE}
    ]
\addplot[color5bg,ultra thick,mark=square*,dashed, dash pattern=on 6pt off 3pt, draw opacity=1.0] table [y=Part12,x=Time]{ranking.csv};
\addplot[color5bg,ultra thick,mark=square*, draw opacity=1.0] table [y=Part13,x=Time]{ranking.csv};
\node[] at (axis cs: 2023-3-10,600) {\large OLMo};
\node[] at (axis cs: 2019-5-20,520) {\small OLMo-Instruct};
\node[] at (axis cs: 2019-3-20,230) {\small OLMo};
\end{axis}
\end{tikzpicture}}

\resizebox{0.485\textwidth}{!}{
\hspace{-2.5pt}
\begin{tikzpicture}
\pgfplotsset{
compat=newest,
every axis plot/.append style={no marks,thick},
every axis/.style={
  width=16cm,
  height=4cm,
  }
}
\begin{axis}[
    ymin=50,
    ymax=750,
    date coordinates in=x,
    table/col sep=comma,
    legend style={at={(0.98,0.3)}},
    xtick pos=lower, ytick pos=left,
    label style={inner sep=0pt}, 
    axis x line*=bottom,
    axis y line*=left,
     y post scale=1.9,
    xtick={2019-01-01, 2020-01-01, 2021-01-01, 2022-01-01,2023-01-01, 2024-01-01},
    xticklabel style={
        rotate=45,
        anchor=east,
    },
    ytick={100,200,300,400,500,600,700},
    xmin=2019-1-01,
    xmax=2023-08-01,
    xticklabels=\empty,
    extra x ticks={2023-08-01},
    extra tick style={grid=major,major grid style={red1,very thick}},
    extra x tick labels = {},
    grid=major,
    grid style={dashed,gray!30},
    label style={font=\LARGE},
    y tick label style={font=\LARGE},
    legend style={font=\normalsize},
    xticklabel style={yshift=-3pt},
    ylabel=Ranking,
    y label style={font=\LARGE}
    ]
\addplot[color6bg,ultra thick,mark=square*, draw opacity=1.0] table [y= Part2,x=Time]{ranking.csv};
\addplot[color6bg,ultra thick,mark=square*,dash pattern=on 6pt off 3pt, draw opacity=1.0] table [y=Part1,x=Time]{ranking.csv};
\addplot[color1bg,ultra thick,dashed,dash pattern=on 6pt off 3pt,mark=square*, draw opacity=1.0] table [y=Part10,x=Time]{ranking.csv};
\addplot[color1bg,ultra thick,mark=square*, draw opacity=1.0] table [y=Part11,x=Time]{ranking.csv};

\node[] at (axis cs: 2022-07-6,720) {\large LLaMA-2};
\node[] at (axis cs: 2023-05-6,720) {\large Mistral};
\node[] at (axis cs: 2019-5-28,670) {\small LLaMA-2 Chat};
\node[] at (axis cs: 2019-4-13,400) {\small LLaMA-2};
\node[] at (axis cs: 2019-5-25,250) {\small Mistral-Instruct};
\node[] at (axis cs: 2019-3-25,130) {\small Mistral};
\draw[very thick, dashed, line1color] (2022-11-01,0) -- (2022-11-01,775);
\end{axis}
\end{tikzpicture}
}

\resizebox{0.485\textwidth}{!}{
\begin{tikzpicture}
\pgfplotsset{
compat=newest,
every axis plot/.append style={no marks,thick},
every axis/.style={
  width=16cm,
  height=4cm,
  }
}
\begin{axis}[
    name=bottomplot,
    ymin=250,
    ymax=790,
    date coordinates in=x,
    table/col sep=comma,
    legend style={at={(0.98,0.42)}},
    xtick pos=lower, ytick pos=left,
    label style={inner sep=0pt}, 
    axis x line*=bottom,
    axis y line*=left,
     y post scale=1.9,
    xtick={2019-01-01, 2020-01-01, 2021-01-01, 2022-01-01,2023-01-01, 2024-01-01},
    xticklabel style={
        rotate=0,
        anchor=center,
    },
    xmin=2019-1-01,
    xmax=2023-08-01,
    xticklabel=\empty,
    extra x ticks={2022-10-01},
    extra tick style={grid=major,major grid style={purple1,very thick}},
    extra x tick labels = {},
    grid=major,
    ytick={300,400,500,600,700},
    grid style={dashed,gray!30},
    label style={font=\LARGE},
    y tick label style={font=\LARGE},
    legend style={font=\normalsize},
    xticklabel style={yshift=-12pt, font=\Large},
    ylabel=Ranking,
    y label style={font=\LARGE}
    ]
\addplot[green1,ultra thick,mark=square*, draw opacity=1.0] table [y= Part5,x=Time]{ranking.csv};
\addplot[purple2,ultra thick,mark=square*, draw opacity = 1.0] table [y= Part4,x=Time]{ranking.csv};
\addplot[purple2,ultra thick,mark=square*,dashed,dash pattern=on 6pt off 3pt, draw opacity = 1.0] table [y= Part3,x=Time]{ranking.csv};
\draw[very thick, dashed, green3] (2021-1-01,250) -- (2021-1-01,775);
\node[] at (axis cs: 2023-01-25,740) {\large LLaMA-1};
\node[] at (axis cs: 2021-03-20,740) {\large GPT-J};
\node[] at (axis cs: 2019-8-1,300) {\small LLaMA-1};
\node[] at (axis cs: 2019-7-10,495) {\small GPT-J};
\node[] at (axis cs: 2019-7-10,395) {\small Alpaca};
\end{axis}
\end{tikzpicture}
}

\vspace*{-6pt}

\resizebox{0.485\textwidth}{!}{
\begin{tikzpicture}
\begin{axis}[dash pattern=on 2pt off 2pt, dash phase=2pt, height=5.2cm,width=4.8cm, name=text1, xtick=0.51, xticklabel style={font=\LARGE, yshift=-3pt},  xticklabels=2019, xtick pos=bottom, xtick style={draw=none}, anchor=north,
ytick=\empty, yticklabels = {}, xmin=0, xmax=1, ymin=0, ymax=1,
legend style={at={(0.64,0.12)},anchor=west},  xlabel=\empty,
title style={at={(0.5,1.08)},font=\small, anchor=north,yshift=-0.1}] 
\node[font=\small, text width=3cm, align=left] at (axis cs: 0.52,0.5) {\it {\fontsize{16}{15}\selectfont {softboy \vspace{2pt}\newline throupling \vspace{1pt}\newline chonker \vspace{2pt}\newline fansplaining\vspace{2pt}\newline poshtels\newline\textcolor{white}{.}$\cdots$

}}};
\end{axis}
\begin{axis}[dash pattern=on 2pt off 2pt, dash phase=2pt,height=5.2cm,width=4.8cm, name=text2, xtick style={draw=none}, xtick=0.5, xticklabel style={font=\LARGE, yshift=-3pt},  xticklabels=2020, xtick pos=bottom, at = {($(text1.east)$)}, anchor=west, 
ytick=\empty, yticklabels = {}, xmin=0, xmax=1, ymin=0, ymax=1,
legend style={at={(0.64,0.12)},anchor=west},axis y line*=right,  xlabel=\empty,
title style={at={(0.5,1.08)},font=\small, anchor=north,yshift=-0.1}] 
\node[font=\small, text width=3cm, align=left] at (axis cs: 0.51,0.5) {\it {\fontsize{16}{15}\selectfont {covid\vspace{2pt}\newline maskne \vspace{1pt}\newline neopronouns\vspace{1pt}\newline hyflex \vspace{2pt}\newline latinx \newline\textcolor{white}{.}$\cdots$

}}};
\end{axis}
\begin{axis}[dash pattern=on 2pt off 2pt, dash phase=2pt,height=5.2cm,width=4.8cm, name=text3,xtick style={draw=none}, xtick=0.5, xticklabel style={font=\LARGE, yshift=-3pt},  xticklabels=2021, xtick pos=bottom, at = {($(text2.east)$)}, anchor=west, 
ytick=\empty, yticklabels = {}, xmin=0, xmax=1, ymin=0, ymax=1,
legend style={at={(0.64,0.12)},anchor=west},axis y line*=right,  xlabel=\empty,
title style={at={(0.5,1.08)},font=\small, anchor=north,yshift=-0.1}] 
\node[font=\small, text width=3cm, align=left] at (axis cs: 0.5,0.5) {\it {\fontsize{16}{15}\selectfont {longcovid\vspace{2pt}\newline hyperpop\vspace{1pt}\newline cheugy \vspace{2pt}\newline greenhushing\vspace{2pt}\newline GOAT'ed \newline \textcolor{white}{.}$\cdots$

}}};
\end{axis}
\begin{axis}[dash pattern=on 2pt off 2pt, dash phase=2pt,height=5.2cm,width=4.6cm, name=text4, xtick=0.48, xticklabel style={font=\LARGE, yshift=-3pt},  xticklabels=2022, xtick pos=bottom, xtick style={draw=none}, at = {($(text3.east)$)}, anchor=west, 
ytick=\empty, yticklabels = {}, xmin=0, xmax=1, ymin=0, ymax=1,
legend style={at={(0.64,0.12)},anchor=west},axis y line*=right,  xlabel=\empty,
title style={at={(0.5,1.08)},font=\small, anchor=north,yshift=-0.1}] 
\node[font=\small, text width=3cm, align=left] at (axis cs: 0.545,0.5) {\it {\fontsize{16}{15}\selectfont {phygital\vspace{2pt}\newline pearlcore \vspace{1pt}\newline v-tubers\vspace{1pt}\newline flurona\vspace{2pt}\newline neotrad \newline \textcolor{white}{.}$\cdots$

}}};
\end{axis}
\begin{axis}[dash pattern=on 2pt off 2pt, dash phase=2pt,height=5.2cm,width=5cm, name=text5, xtick=0.45, xticklabel style={font=\LARGE, yshift=-3pt},  xticklabels=2023, xtick pos=bottom, xtick style={draw=none}, at = {($(text4.east)$)}, anchor=west, 
ytick=\empty, yticklabels = {}, xmin=0, xmax=1, ymin=0, ymax=1,
legend style={at={(0.64,0.12)},anchor=west},axis y line*=right,  xlabel=\empty,
title style={at={(0.5,1.08)},font=\small, anchor=north,yshift=-0.1}] 
\node[font=\small, text width=3cm, align=left] at (axis cs: 0.48,0.5) { \it {\fontsize{16}{15}\selectfont {blud\vspace{2pt}\newline overvaccinate\vspace{2pt}\newline  fomosapien\vspace{2pt}\newline slackercore \vspace{2pt}\newline doomposters \newline \textcolor{white}{.}$\cdots$

}}};
\end{axis}
\end{tikzpicture}
}
\vspace{-22pt}
\caption[ranking caption]{Rankings of neologisms over time compared to 5000 common words. Newer models are plotted separately. Dashed lines show model knowledge cutoffs\footnotemark. Example neologisms from each year are provided, and neologisms without trendlines are reported at the end.}
\label{figure:time_rankings}
\end{figure}

\footnotetext{Mistral AI does not reveal training data details, so we provide our best estimation for the model's knowledge cutoff.}

\vspace{2pt}
\noindent \textbf{Older LLMs perform significantly worse.} In Figure \ref{tab:cloze_and_definition_results}, the average performance of GPT-J, BART, T5, and Flan-T5 is 26.99\% lower in Cloze QA and 47.78\% lower in Definition Generation than other models. In Figure \ref{fig:ranking_results}, GPT-J and LLaMA-1 models exhibit higher neologism rankings than newer open-source models, correlating with lower downstream performance. Newer models -- GPT-3.5, GPT-4, LLaMA-2, OLMo, and Mistral -- perform better as they are trained on data containing newer neologisms, generally have algorithmic improvements, and are trained with more resources than older models. 
\vspace{1pt}

\noindent  \textbf{Perplexity rankings of older models increase drastically from 2019 until 2021.} While neologisms are often gradually worked into a vocabulary \cite{DBLP:journals/corr/abs-2104-05010}, we use trend lines to best estimate the date when a neologism becomes popular and report perplexity over time in Figure \ref{figure:time_rankings}. Newer models dip in 2020 but increase afterward as 52\% of neologisms from this period are now conventionalized terms related to COVID-19.

\vspace{2pt}

\noindent \textbf{Larger models handle neologisms better.} Increasing the sizes of LLaMa-2 and LLaMA-2 Chat leads to consistent improvements across both Cloze Question Answering and Definition Generation. On average, LaMA-2 70B and LLaMA-2 Chat 70B yield 10.13\% higher accuracy in Cloze QA and 5.93\% higher accuracy in Definition Generation than LLaMA-2 7B and LLaMA-2 Chat 7B.
\vspace{2pt}

\noindent \textbf{Instruction-tuning results in high neologism perplexities.} In Figure \ref{fig:ranking_results}, LLaMA-1, LLaMA-2, OLMo, and Mistral models have, on average, 125 lower neologism rankings than their instruct-tuned counterparts. Instruct models are trained with dialogue \cite{wei2022finetuned}, so uncommon generation is less desired.

\section{Linguistic Taxonomy Analysis} \label{sec:ling_results}
\begin{figure}[!t]
\pgfplotsset{
    testbar11/.style={
        ybar stacked,
        bar width=3.6pt,
        axis x line*=bottom,
        axis y line*=left,
        xtick pos=lower, ytick pos=left,
        ymin=0,ymax=101.5,
        xtick = data,
        xtick distance=1,
        xticklabel style={text width=1.25cm, align=right, font=\fontsize{6}{5}\selectfont, anchor=east, scale=0.5, yshift=0cm, rotate=35},
        ytick style={yshift=0cm},
        yticklabel ={$\pgfmathprintnumber{\tick}\%$},
        yticklabel style={font=\tiny, scale=0.5, align=right},
        major tick length =2pt,
        legend style={at={(0.3,0.05)}, font=\tiny, anchor=west},
        tick align = outside,
        height=0.23\textwidth,
        xlabel style = {font=\small},
        x=1.8mm,
        enlarge x limits={abs=0.65},
    }}
\hspace{-20pt}
\resizebox{0.52\textwidth}{!}{
\begin{tikzpicture}
\tikzstyle{every node}=[font=\tiny]
\begin{axis}[testbar11, name=bing, xtick pos=lower, ytick pos=left, anchor=north west, xticklabels = {Semantic, Morphological, Lexical, Semantic, Morphological, Lexical, Semantic,Morphological, Lexical, Semantic, Morphological, Lexical, Semantic, Morphological, Lexical, Semantic, Morphological, Lexical, Semantic, Morphological, Lexical, Semantic, Morphological, Lexical, Semantic, Morphological, Lexical}, legend style={at={(0.5,-0.2), font=\tiny,nodes={scale=0.2}},fill=none,draw=none,anchor=north, scale=0.02},
title style={at={(0,0.93)}, font=\fontsize{6}{6}\selectfont,anchor=west,yshift=-0.1, scale=0.7},
title = \textbf{\hspace{3pt}Bing\hspace{10pt}GPT-4\hspace{8pt}Google\hspace{9pt}DeepL\hspace{6pt}GPT-3.5\hspace{5pt}ALMA\hspace{5pt}M2M100}, legend columns=-1, enlarge y limits={abs=0.6},
] 
 \addplot[fill=color1bg] coordinates{
                                    (2, 35)
                                    (1, 38.6)
                                    (0, 63.9)
                                    (5.5, 40)
                                    (4.5, 36.4)
                                    (3.5, 61.1)
                                    (9, 35)
                                    (8,34.1)
                                    (7,63.9)
                                    (12.5, 25)
                                    (11.5, 27.3)
                                    (10.5, 58.3)
                                    (16, 55.6)
                                    (15, 18.2)
                                    (14, 30)
                                    (19.5, 10)
                                    (18.5, 25)
                                    (17.5, 38.9)
                                    (23, 27.8)
                                    (22, 13.6)
                                    (21, 5)
                                    };
 \addplot[fill=color2bg] coordinates{
                                   (2, 0)
                                   (1, 4.6)
                                   (0, 0)
                                   (5.5, 5)
                                   (4.5, 6.8)
                                   (3.5, 8.3)
                                   (9,10)
                                   (8,6.8)
                                   (7,2.8)
                                   (12.5, 5)
                                   (11.5, 6.8)
                                   (10.5, 5.6)
                                   (16, 16.7)
                                   (15, 9.1)
                                   (14, 15)
                                   (19.5, 5)
                                   (18.5, 13.6)
                                   (17.5, 8.3)
                                   (23, 11.1)
                                   (22, 9.1)
                                   (21, 10)
                                   };

 \addplot[fill=color3bg] coordinates{
                                   (2, 30)
                                   (1, 18.2)
                                   (0, 19.4)
                                   (5.5, 30)
                                   (4.5, 9.1)
                                   (3.5, 8.3)
                                   (9,25)
                                   (8,9.1)
                                   (7,5.6)
                                   (12.5, 30)
                                   (11.5, 9.1)
                                   (10.5, 2.8)
                                   (16, 13.9)
                                   (15, 4.6)
                                   (14, 15)
                                   (19.5, 10)
                                   (18.5, 9.1)
                                   (17.5, 5.6)
                                   (23, 8.3)
                                   (22, 0)
                                   (21, 5)
                                   };
                                   
 \addplot[fill=color4bg] coordinates{
                                   (2, 0)
                                   (1, 4.6)
                                   (0, 2.8)
                                   (5.5, 10)
                                   (4.5, 13.6)
                                   (3.5, 0)
                                   (9,5)
                                   (8,2.3)
                                   (7,2.8)
                                   (12.5, 5)
                                   (11.5, 11.4)
                                   (10.5, 2.8)
                                   (16, 0)
                                   (15, 6.8)
                                   (14, 10)
                                   (19.5, 15)
                                   (18.5, 2.3)
                                   (17.5, 13.9)
                                   (23, 2.8)
                                   (22, 2.3)
                                   (21, 10)
                                   };
                                   
 \addplot[fill=color5bg] coordinates{
                                   (2, 15)
                                   (1, 20.5)
                                   (0, 8.3)
                                   (5.5, 10)
                                   (4.5, 29.5)
                                   (3.5, 16.7)
                                   (9,20)
                                   (8,34.1)
                                   (7,22.2)
                                   (12.5, 30)
                                   (11.5, 36.4)
                                   (10.5, 27.8)
                                   (16, 11.1)
                                   (15, 43.2)
                                   (14, 20)
                                   (19.5, 55)
                                   (18.5, 40.9)
                                   (17.5, 25)
                                   (23, 36.1)
                                   (22, 56.8)
                                   (21, 50)
                                   };

 \addplot[fill=color6bg] coordinates{
                                   (2, 20)
                                   (1, 11.4)
                                   (0, 2.8)
                                   (5.5, 5)
                                   (4.5, 2.3)
                                   (3.5, 5.6)
                                   (9,5)
                                   (8,11.4)
                                   (7,2.8)
                                   (12.5, 5)
                                   (11.5, 9.1)
                                   (10.5, 2.8)
                                   (16, 2.8)
                                   (15, 15.9)
                                   (14, 10)
                                   (19.5, 0)
                                   (18.5, 9.1)
                                   (17.5, 2.8)
                                   (23, 5.6)
                                   (22, 4.6)
                                   (21, 10)
                                   };

 \addplot[fill=color7bg] coordinates{
                                   (2, 0)
                                   (1, 2.3)                                   
                                   (0, 2.8)
                                   (5.5, 0)
                                   (4.5, 2.3)
                                   (3.5, 0)
                                   (9,0)
                                   (8,2.3)                                   
                                   (7,0)
                                   (12.5, 0)
                                   (11.5, 0)                                   
                                   (10.5, 0)
                                   (16, 0)
                                   (15,2.3)
                                   (14,0)
                                   (19.5, 5)
                                   (18.5, 0)                                   
                                   (17.5, 5.6)
                                   (23, 8.3)
                                   (22, 13.6)
                                   (21, 10)
                                   };
\end{axis}
\end{tikzpicture}

}
\vspace*{-8pt}
\centering
\includegraphics[width=0.45\textwidth]{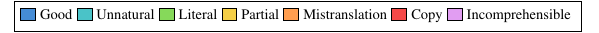}
\caption{Results of the human-annotated MT models for each linguistic type of neologism.}
\label{fig:mt_linguistic_main}
\end{figure}
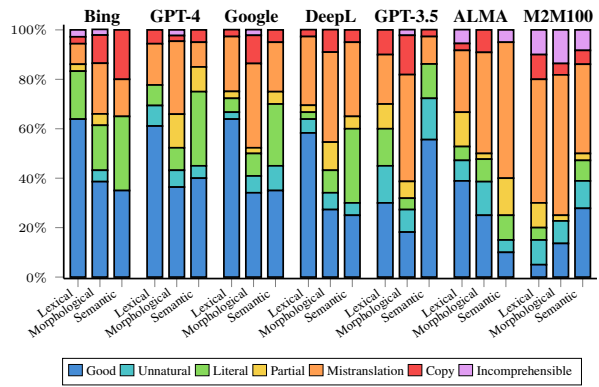

\begin{figure}[!tbp]
    \centering
    \includegraphics[width=0.48\textwidth]{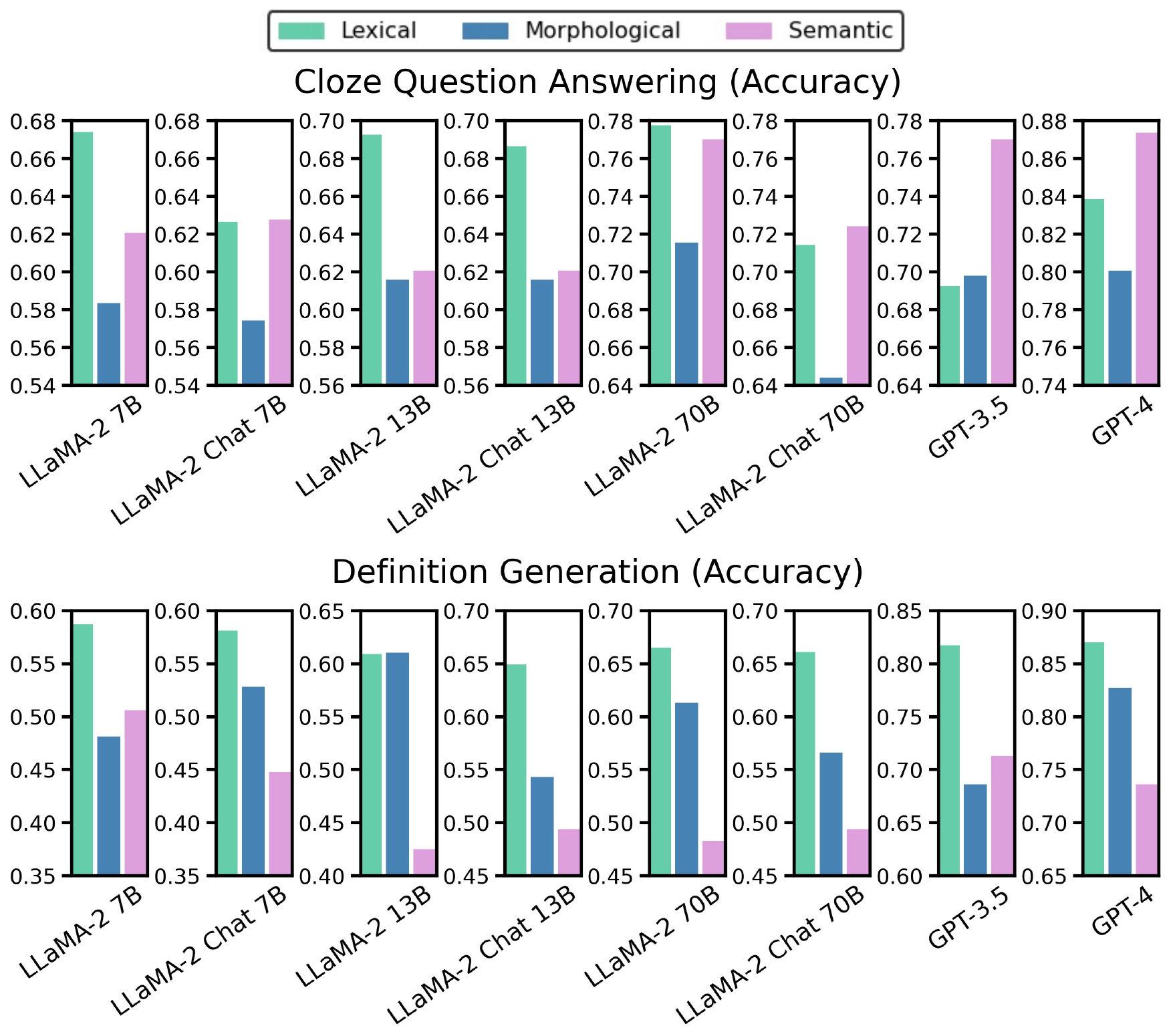}
\vspace*{-9mm}
  \caption{Results of Cloze QA and Definition Generation stratified by linguistic types of neologisms.}
    \label{fig:ling_main_paper}
\end{figure}

We separate {\sc Neo-Bench} task results by neologism linguistic structure: lexical, morphological, and semantic. Figure \ref{fig:ranking_results} presents perplexity rankings, Figure \ref{fig:mt_linguistic_main} reports human evaluation for MT, and Figure \ref{fig:ling_main_paper} shows the results for the best models on Cloze QA and Definition Generation.

\vspace{2pt}

\begin{figure}[!tbp]
    \centering
    \includegraphics[width=0.48\textwidth]{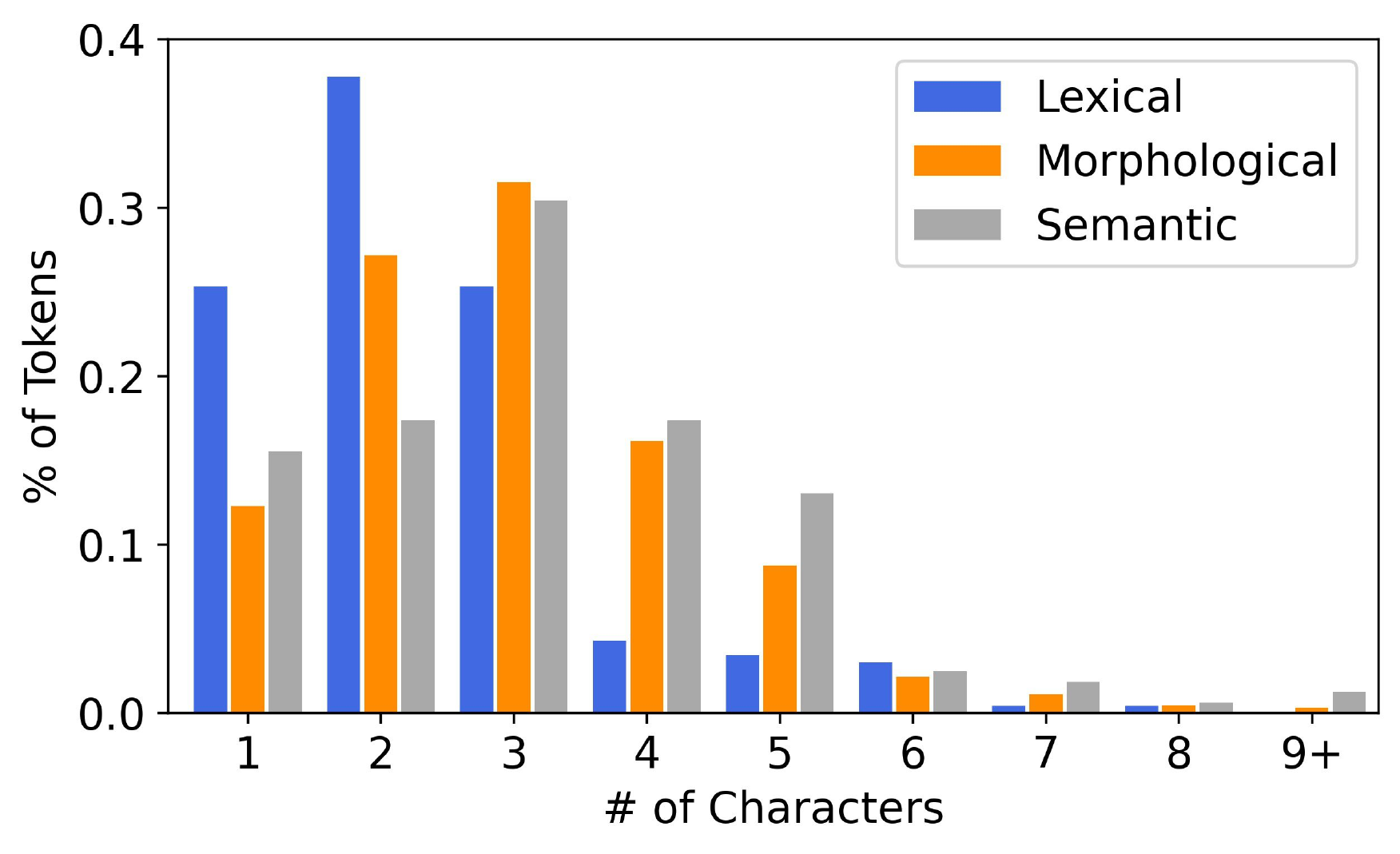}
\vspace*{-9mm}
\caption{Distributions of characters per subword token of each neologism type, reported by the proportion of tokens that have a certain number of characters.}\label{fig:Label}
\label{fig:linguistic_histogram}
\end{figure}

\noindent \textbf{Lexical neologisms produce the highest perplexities, but yield the best downstream results.} On average, lexical neologisms have 226 higher rankings than other words, indicating higher relative model perplexities. Figure \ref{fig:linguistic_histogram} shows the distribution of characters per token of neologisms using the LLaMA tokenizer, and the average number of characters per token for lexical, morphological, and semantic neologisms is 2.36, 2.98, and 3.24 respectively. Lexical neologisms have more fragmented tokenizations, as these words have the highest proportion of 1-2 character tokens. Lexical neologisms are less likely to be separated into long, common word roots or segments representing subword information, instead producing uncommon token sequences that result in higher neologism rankings and perplexity. In downstream tasks, however, lexical neologisms yield 0.6\% higher Cloze accuracy, 8.8\% more correct definitions, and 21.5\% more good translations than other neologisms. 

\vspace{2pt}

\noindent \textbf{Morphological neologisms produce low perplexities but yield poor downstream performance.} Compared to lexical neologisms, morphological neologisms are, on average, segmented into longer tokens and constructed with common subwords, resulting in lower perplexity rankings. However, they yield 4.2\% lower Cloze accuracy, 9.1\% more incorrect definitions, and 26.1\% less good translations than lexical neologisms. 76.8\% of neologisms without trend lines are morphological. Compared to lexical and semantic neologisms that require prevalence to be differentiated from incoherent strings, morphological neologisms are created with polynomial combinations of common subwords. Many of these intelligible combinations are largely unused, resulting in lower downstream performance. 
\vspace{2pt}

\noindent \textbf{Semantic neologisms produce the lowest perplexities and the worst performance in generation tasks.} Since these neologisms use existing word forms, they have an average of 266 lower perplexity rankings than other neologism words. While semantic neologisms yield high Cloze QA accuracy, they also achieve the lowest percentages of correct definitions and translations. Models produce popular definitions and literal translations based on a word's most common meaning, as the new sense of semantic neologisms is often nuanced and difficult to capture. 

\section{Related Work}
\label{sec:related_work}
\paragraph{Temporal Drift in LLMs.} Prior work has explored temporal data drift by creating temporal splits of training data data \cite{loureiro2022timelms, luu-etal-2022-time, rottger-pierrehumbert-2021-temporal-adaptation, jin-etal-2022-lifelong, luu-etal-2022-time, https://doi.org/10.48550/arxiv.2102.01951}. New factual updates of concepts are studied with temporal splits of text corpora and QA datasets \cite{jang-etal-2022-temporalwiki, margatina-etal-2023-dynamic, zhao-etal-2022-impact, vu2023freshllms}. Other work has observed model degradation from new named entities \cite{onoe-etal-2022-entity, rijhwani-preotiuc-pietro-2020-temporally, chen2021mitigating}. Temporal degradation occurs during short-term crisis events where information changes quickly \cite{pramanick-etal-2022-challenges}. Studies have consistently found model degradation with perplexity and downstream tasks. There are no studies on model degradation from language change of neologisms, so we create a benchmark to evaluate models on neologisms with similar tasks. 

\paragraph{Neologism Collection.} Using reference texts as exclusion lists to filter common words from target corpora is the most documented method of neologism detection. Target texts and exclusion lists include news articles \cite{pinter-etal-2020-nytwit, falk-etal-2014-non}, dictionaries \cite{Kerremans2018, Langemets2020, DBLP:conf/rocling/LiuHP13, dhuliawala-etal-2016-slangnet}, social media \cite{korean-neologisms, zalmout-etal-2019-unsupervised, Megerdoomian2010MiningAC} and other corpora \cite{cartier-2017-neoveille, lejeune-cartier-2017-character}. These texts are slow to curate, and semantic neologisms are filtered out. Moreover, no resource collects general semantic neologisms. \newline\indent
gSome resources measure word prevalence with time-series data of search queries to collect single-word neologism candidates \cite{broad-etal-2018-candidate} or cybersecurity neologisms \cite{Li2021}. They are limited in scope by collecting only one type of neologism based on rising popularity. Other methods collect neologisms with new slang dictionary entries \cite{dhuliawala-etal-2016-slangnet, DBLP:journals/corr/abs-2104-05010}. Dictionaries are slow to update, so new entries are often conventionalized words. There is no resource that uses time-series data to collect a variety of neologisms rising in prevalence.

Previous work has utilized search templates of explanation patterns to collect automatically neologisms \cite{breen-etal-2018-company}. A few efforts have used neural methods to automatically detect one specific type of neologism, such as adjective-noun neologism pairs \cite{mccrae-2019-identification}, blend words \cite{Megerdoomian2010MiningAC}, and grammatical neologisms \cite{janssen-2012-neotag, falk-etal-2014-non}, which are existing words with new parts of speech. No automatic resource collects neologisms from various topics and linguistic backgrounds. To address these limitations, {\sc Neo-Bench} uses multiple methods to semi-automatically collect a variety of neologisms, including multiword, semantic, and prevalent neologisms. 

\paragraph{Unseen Words.} Rare words or typos are typically unseen in data when training a model but may show up during inference. Prior work has measured model degradation from unseen words \cite{chirkova2021simple, nayak-etal-2020-domain} and used contextual subword embeddings \cite{garneau-etal-2018-predicting, DBLP:journals/corr/abs-1903-10635, hu-etal-2019-shot, DBLP:journals/corr/abs-1909-03341, araabi-etal-2022-effective}, similar surface forms of common words \cite{chen2022imputing, fukuda-etal-2020-robust}, and morphological structure \cite{DBLP:journals/corr/abs-2007-07318, LOCHTER2022108911} to represent unseen words. Comparatively, the neologism lifecycle follows three stages: emergence, dissemination, and conventionalization \cite{cartier-2017-neoveille}. New words often become prevalent and drastically shift a language's distribution. Semantic neologisms also use existing word forms and are not classified as unseen words.

\section{Conclusion}
In this paper, we present {\sc Neo-Bench}, a new benchmark to test the ability of LLMs to generalize on neologisms. We use several methods to collect a variety of neologisms, including prevalent, multiword, and semantic neologisms. In our experiments, we find that models struggle with neologisms in both perplexity and downstream tasks. Machine Translation is especially difficult, as translating neologisms often requires paraphrasing the sentence. Current automatic metrics cannot measure translation quality, and human evaluation is still needed. Neologisms also affect models differently based on linguistic structure, indicating that this phenomenon is complex for LLMs to address.

\section*{Limitations}
Most of our neologisms largely originated in US and UK English, as we collect textual data from news articles from this region. We do not restrict our locations for Reddit data, but the majority of English-speaking Reddit users are also from the same regions. Given our limited expertise in other English dialects, especially regions whose English variations are largely influenced by other languages, we do not collect many neologisms from English-speaking regions outside these regions. However, our computational framework for collecting neologisms can be applied to any language or local variation. For temporal drift of multilingual language modeling, we leave multilingual neologism collection and temporal drift analysis up for future work. Additionally, {\sc Neo-Bench} is static as we collect neologisms from mostly 2020-2023, which will become outdated over time as newer models will be exposed to new language in context. However, the semi-automatic collection methods require minimal human supervision and can be dynamically updated to continuously obtain neologisms. These methods require time-series data of words and online text corpora without needing human-curated information like updated dictionaries to filter words. The time-independent filters can collect recent neologisms without needing the time-consuming, manual curation of temporal splits of textual data. Provided that the Google and Reddit Terms of Service and API access enable {\sc Neo-Bench} collection, we intend on periodically updating the set of neologisms in our dataset. 

\section*{Ethical Considerations}
We utilize Reddit monthly dumps to obtain uncommon words, which often include sensitive information such as account usernames. We take the appropriate measures to ensure that no personally identifiable information (PII) is included in our dataset. We use a named-entity recognition model via SpaCy to identify named entities that are potentially PII and largely remove this information automatically when filtering for neologism candidates. We also manually inspect all the candidates to ensure that no PII is included in our dataset. As we use natural references from Google to construct our model inputs, we also review our hand-crafted sentences to ensure that there is no PII contained in these sentences. Many neologism entries in our work emerge from slang, and some slang words have expletive or offensive meanings. The purpose of our dataset and benchmark is to obtain a representative sample of neologisms and comprehensively evaluate the impact of neologisms on Large Language Models. We present examples that do not contain such offensive information, but these offensive entries are nonetheless a consistent source of neologisms. For expletive neologisms, we strive to create input sentences that capture the meaning of the neologism while not perpetuating gender, racial, and other potential biases. We do not collect any neologisms that directly reference stereotypes and demographic biases.

\section*{Acknowledgments}
We would like to thank Piranava Abeyakaran, Kaushik Sriram, Vishnesh Jayanthi Ramanathan, Yao Dou, Chao Jiang, Cai Yang, and Xiaofeng Wu for their help in constructing model inputs and translating reference sentences; Qihang Yao and Sam Stevens for their scripts for processing Google Trends and Urban Dictionary data, respectively; Graham Neubig and Paul Michel for valuable discussions. We would also like to thank Azure’s Accelerate Foundation Models Research Program for graciously providing access to API-based models, such as GPT-4.  This research is supported in part by the NSF CAREER awards IIS-2144493 and IIS-2052498, ODNI and IARPA via the HIATUS program (contract 2022-22072200004). The views and conclusions contained herein are those of the authors and should not be interpreted as necessarily representing the official policies, either expressed or implied, of NSF, ODNI, IARPA, or the U.S. Government. The U.S. Government is authorized to reproduce and distribute reprints for governmental purposes notwithstanding any copyright annotation therein.

\bibliography{acl_latex}

\begin{thebibliography}{67}
\expandafter\ifx\csname natexlab\endcsname\relax\def\natexlab#1{#1}\fi

\bibitem[{Agarwal and Nenkova(2022)}]{agarwal2022temporal}
Oshin Agarwal and Ani Nenkova. 2022.
\newblock Temporal effects on pre-trained models for language processing tasks.
\newblock \emph{Transactions of the Association for Computational Linguistics}.

\bibitem[{Araabi et~al.(2022)Araabi, Monz, and
  Niculae}]{araabi-etal-2022-effective}
Ali Araabi, Christof Monz, and Vlad Niculae. 2022.
\newblock \href {https://aclanthology.org/2022.amta-research.9} {How effective
  is byte pair encoding for out-of-vocabulary words in neural machine
  translation?}
\newblock In \emph{Proceedings of the 15th biennial conference of the
  Association for Machine Translation in the Americas (Volume 1: Research
  Track)}, pages 117--130, Orlando, USA. Association for Machine Translation in
  the Americas.

\bibitem[{Bird et~al.(2009)Bird, Klein, and Loper}]{bird2009natural}
Steven Bird, Ewan Klein, and Edward Loper. 2009.
\newblock \emph{Natural language processing with Python: analyzing text with
  the natural language toolkit}.
\newblock " O'Reilly Media, Inc.".

\bibitem[{Blain et~al.(2023)Blain, Zerva, Ribeiro, Guerreiro, Kanojia, C.~de
  Souza, Silva, Vaz, Jingxuan, Azadi, Orasan, and
  Martins}]{blain-etal-2023-findings}
Frederic Blain, Chrysoula Zerva, Ricardo Ribeiro, Nuno~M. Guerreiro, Diptesh
  Kanojia, Jos{\'e}~G. C.~de Souza, Beatriz Silva, T{\^a}nia Vaz, Yan Jingxuan,
  Fatemeh Azadi, Constantin Orasan, and Andr{\'e} Martins. 2023.
\newblock \href {https://doi.org/10.18653/v1/2023.wmt-1.52} {Findings of the
  {WMT} 2023 shared task on quality estimation}.
\newblock In \emph{Proceedings of the Eighth Conference on Machine
  Translation}, pages 629--653, Singapore. Association for Computational
  Linguistics.

\bibitem[{Breen et~al.(2018)Breen, Baldwin, and Bond}]{breen-etal-2018-company}
James Breen, Timothy Baldwin, and Francis Bond. 2018.
\newblock \href {https://aclanthology.org/2018.gwc-1.19} {The company they
  keep: Extracting {J}apanese neologisms using language patterns}.
\newblock In \emph{Proceedings of the 9th Global Wordnet Conference}, pages
  163--171, Nanyang Technological University (NTU), Singapore. Global Wordnet
  Association.

\bibitem[{Broad et~al.(2018)Broad, Langone, and
  Brizan}]{broad-etal-2018-candidate}
Claire Broad, Helen Langone, and David~Guy Brizan. 2018.
\newblock \href {https://aclanthology.org/L18-1134} {Candidate ranking for
  maintenance of an online dictionary}.
\newblock In \emph{Proceedings of the Eleventh International Conference on
  Language Resources and Evaluation ({LREC} 2018)}, Miyazaki, Japan. European
  Language Resources Association (ELRA).

\bibitem[{Brown et~al.(2020)Brown, Mann, Ryder, Subbiah, Kaplan, Dhariwal,
  Neelakantan, Shyam, Sastry, Askell et~al.}]{brown2020language}
Tom~B Brown, Benjamin Mann, Nick Ryder, Melanie Subbiah, Jared Kaplan, Prafulla
  Dhariwal, Arvind Neelakantan, Pranav Shyam, Girish Sastry, Amanda Askell,
  et~al. 2020.
\newblock Language models are few-shot learners.
\newblock \emph{arXiv preprint arXiv:2005.14165}.

\bibitem[{Cartier(2017)}]{cartier-2017-neoveille}
Emmanuel Cartier. 2017.
\newblock \href {https://aclanthology.org/E17-3024} {Neoveille, a web platform
  for neologism tracking}.
\newblock In \emph{Proceedings of the Software Demonstrations of the 15th
  Conference of the {E}uropean Chapter of the Association for Computational
  Linguistics}, pages 95--98, Valencia, Spain. Association for Computational
  Linguistics.

\bibitem[{Chen et~al.(2022)Chen, Varoquaux, and Suchanek}]{chen2022imputing}
Lihu Chen, Gaël Varoquaux, and Fabian~M. Suchanek. 2022.
\newblock \href {http://arxiv.org/abs/2203.07860} {Imputing out-of-vocabulary
  embeddings with love makes language models robust with little cost}.

\bibitem[{Chen et~al.(2019)Chen, Mathews, Ouyang, and
  Beaufays}]{DBLP:journals/corr/abs-1903-10635}
Mingqing Chen, Rajiv Mathews, Tom Ouyang, and Fran{\c{c}}oise Beaufays. 2019.
\newblock \href {http://arxiv.org/abs/1903.10635} {Federated learning of
  out-of-vocabulary words}.
\newblock \emph{CoRR}, abs/1903.10635.

\bibitem[{Chen et~al.(2021)Chen, Neves, and Solorio}]{chen2021mitigating}
Shuguang Chen, Leonardo Neves, and Thamar Solorio. 2021.
\newblock \href {http://arxiv.org/abs/2104.09742} {Mitigating temporal-drift: A
  simple approach to keep ner models crisp}.

\bibitem[{Chirkova and Troshin(2021)}]{chirkova2021simple}
Nadezhda Chirkova and Sergey Troshin. 2021.
\newblock \href {http://arxiv.org/abs/2010.12663} {A simple approach for
  handling out-of-vocabulary identifiers in deep learning for source code}.

\bibitem[{Chung et~al.(2022)Chung, Hou, Longpre, Zoph, Tay, Fedus, Li, Wang,
  Dehghani, Brahma, Webson, Gu, Dai, Suzgun, Chen, Chowdhery, Castro-Ros,
  Pellat, Robinson, Valter, Narang, Mishra, Yu, Zhao, Huang, Dai, Yu, Petrov,
  Chi, Dean, Devlin, Roberts, Zhou, Le, and Wei}]{chung2022scaling}
Hyung~Won Chung, Le~Hou, Shayne Longpre, Barret Zoph, Yi~Tay, William Fedus,
  Yunxuan Li, Xuezhi Wang, Mostafa Dehghani, Siddhartha Brahma, Albert Webson,
  Shixiang~Shane Gu, Zhuyun Dai, Mirac Suzgun, Xinyun Chen, Aakanksha
  Chowdhery, Alex Castro-Ros, Marie Pellat, Kevin Robinson, Dasha Valter,
  Sharan Narang, Gaurav Mishra, Adams Yu, Vincent Zhao, Yanping Huang, Andrew
  Dai, Hongkun Yu, Slav Petrov, Ed~H. Chi, Jeff Dean, Jacob Devlin, Adam
  Roberts, Denny Zhou, Quoc~V. Le, and Jason Wei. 2022.
\newblock \href {http://arxiv.org/abs/2210.11416} {Scaling
  instruction-finetuned language models}.

\bibitem[{Dhuliawala et~al.(2016)Dhuliawala, Kanojia, and
  Bhattacharyya}]{dhuliawala-etal-2016-slangnet}
Shehzaad Dhuliawala, Diptesh Kanojia, and Pushpak Bhattacharyya. 2016.
\newblock \href {https://aclanthology.org/L16-1686} {{S}lang{N}et: A
  {W}ord{N}et like resource for {E}nglish slang}.
\newblock In \emph{Proceedings of the Tenth International Conference on
  Language Resources and Evaluation ({LREC}'16)}, pages 4329--4332,
  Portoro{\v{z}}, Slovenia. European Language Resources Association (ELRA).

\bibitem[{Falk et~al.(2014)Falk, Bernhard, and G{\'e}rard}]{falk-etal-2014-non}
Ingrid Falk, Delphine Bernhard, and Christophe G{\'e}rard. 2014.
\newblock \href
  {http://www.lrec-conf.org/proceedings/lrec2014/pdf/288_Paper.pdf} {From non
  word to new word: Automatically identifying neologisms in {F}rench
  newspapers}.
\newblock In \emph{Proceedings of the Ninth International Conference on
  Language Resources and Evaluation ({LREC}'14)}, pages 4337--4344, Reykjavik,
  Iceland. European Language Resources Association (ELRA).

\bibitem[{Fan et~al.(2020)Fan, Bhosale, Schwenk, Ma, El{-}Kishky, Goyal,
  Baines, Celebi, Wenzek, Chaudhary, Goyal, Birch, Liptchinsky, Edunov, Grave,
  Auli, and Joulin}]{DBLP:journals/corr/abs-2010-11125}
Angela Fan, Shruti Bhosale, Holger Schwenk, Zhiyi Ma, Ahmed El{-}Kishky,
  Siddharth Goyal, Mandeep Baines, Onur Celebi, Guillaume Wenzek, Vishrav
  Chaudhary, Naman Goyal, Tom Birch, Vitaliy Liptchinsky, Sergey Edunov,
  Edouard Grave, Michael Auli, and Armand Joulin. 2020.
\newblock \href {http://arxiv.org/abs/2010.11125} {Beyond english-centric
  multilingual machine translation}.
\newblock \emph{CoRR}, abs/2010.11125.

\bibitem[{Freitag et~al.(2021)Freitag, Foster, Grangier, Ratnakar, Tan, and
  Macherey}]{Freitag_2021}
Markus Freitag, George Foster, David Grangier, Viresh Ratnakar, Qijun Tan, and
  Wolfgang Macherey. 2021.
\newblock \href {https://doi.org/10.1162/tacl_a_00437} {Experts, errors, and
  context: A large-scale study of human evaluation for machine translation}.
\newblock \emph{Transactions of the Association for Computational Linguistics},
  9:1460–1474.

\bibitem[{Fukuda et~al.(2020)Fukuda, Yoshinaga, and
  Kitsuregawa}]{fukuda-etal-2020-robust}
Nobukazu Fukuda, Naoki Yoshinaga, and Masaru Kitsuregawa. 2020.
\newblock \href {https://doi.org/10.18653/v1/2020.findings-emnlp.434} {Robust
  {B}acked-off {E}stimation of {O}ut-of-{V}ocabulary {E}mbeddings}.
\newblock In \emph{Findings of the Association for Computational Linguistics:
  EMNLP 2020}, pages 4827--4838, Online. Association for Computational
  Linguistics.

\bibitem[{Garneau et~al.(2018)Garneau, Leboeuf, and
  Lamontagne}]{garneau-etal-2018-predicting}
Nicolas Garneau, Jean-Samuel Leboeuf, and Luc Lamontagne. 2018.
\newblock \href {https://doi.org/10.18653/v1/W18-5439} {Predicting and
  interpreting embeddings for out of vocabulary words in downstream tasks}.
\newblock In \emph{Proceedings of the 2018 {EMNLP} Workshop {B}lackbox{NLP}:
  Analyzing and Interpreting Neural Networks for {NLP}}, pages 331--333,
  Brussels, Belgium. Association for Computational Linguistics.

\bibitem[{Groeneveld et~al.(2024)Groeneveld, Beltagy, Walsh, Bhagia, Kinney,
  Tafjord, Jha, Ivison, Magnusson, Wang, Arora, Atkinson, Authur, Chandu,
  Cohan, Dumas, Elazar, Gu, Hessel, Khot, Merrill, Morrison, Muennighoff, Naik,
  Nam, Peters, Pyatkin, Ravichander, Schwenk, Shah, Smith, Strubell, Subramani,
  Wortsman, Dasigi, Lambert, Richardson, Zettlemoyer, Dodge, Lo, Soldaini,
  Smith, and Hajishirzi}]{groeneveld2024olmo}
Dirk Groeneveld, Iz~Beltagy, Pete Walsh, Akshita Bhagia, Rodney Kinney, Oyvind
  Tafjord, Ananya~Harsh Jha, Hamish Ivison, Ian Magnusson, Yizhong Wang, Shane
  Arora, David Atkinson, Russell Authur, Khyathi~Raghavi Chandu, Arman Cohan,
  Jennifer Dumas, Yanai Elazar, Yuling Gu, Jack Hessel, Tushar Khot, William
  Merrill, Jacob Morrison, Niklas Muennighoff, Aakanksha Naik, Crystal Nam,
  Matthew~E. Peters, Valentina Pyatkin, Abhilasha Ravichander, Dustin Schwenk,
  Saurabh Shah, Will Smith, Emma Strubell, Nishant Subramani, Mitchell
  Wortsman, Pradeep Dasigi, Nathan Lambert, Kyle Richardson, Luke Zettlemoyer,
  Jesse Dodge, Kyle Lo, Luca Soldaini, Noah~A. Smith, and Hannaneh Hajishirzi.
  2024.
\newblock \href {http://arxiv.org/abs/2402.00838} {Olmo: Accelerating the
  science of language models}.

\bibitem[{Heineman et~al.(2023)Heineman, Dou, and Xu}]{heineman2023thresh}
David Heineman, Yao Dou, and Wei Xu. 2023.
\newblock \href {http://arxiv.org/abs/2308.06953} {Thresh: A unified,
  customizable and deployable platform for fine-grained text evaluation}.

\bibitem[{Honnibal and Montani(2017)}]{spacy2}
Matthew Honnibal and Ines Montani. 2017.
\newblock {spaCy 2}: Natural language understanding with {B}loom embeddings,
  convolutional neural networks and incremental parsing.
\newblock To appear.

\bibitem[{Hu et~al.(2019)Hu, Chen, Chang, and Sun}]{hu-etal-2019-shot}
Ziniu Hu, Ting Chen, Kai-Wei Chang, and Yizhou Sun. 2019.
\newblock \href {https://doi.org/10.18653/v1/P19-1402} {Few-shot representation
  learning for out-of-vocabulary words}.
\newblock In \emph{Proceedings of the 57th Annual Meeting of the Association
  for Computational Linguistics}, pages 4102--4112, Florence, Italy.
  Association for Computational Linguistics.

\bibitem[{Jang et~al.(2022)Jang, Ye, Lee, Yang, Shin, Han, Kim, and
  Seo}]{jang-etal-2022-temporalwiki}
Joel Jang, Seonghyeon Ye, Changho Lee, Sohee Yang, Joongbo Shin, Janghoon Han,
  Gyeonghun Kim, and Minjoon Seo. 2022.
\newblock \href {https://aclanthology.org/2022.emnlp-main.418}
  {{T}emporal{W}iki: A lifelong benchmark for training and evaluating
  ever-evolving language models}.
\newblock In \emph{Proceedings of the 2022 Conference on Empirical Methods in
  Natural Language Processing}, pages 6237--6250, Abu Dhabi, United Arab
  Emirates. Association for Computational Linguistics.

\bibitem[{Janssen(2012)}]{janssen-2012-neotag}
Maarten Janssen. 2012.
\newblock \href
  {http://www.lrec-conf.org/proceedings/lrec2012/pdf/1098_Paper.pdf}
  {{N}eo{T}ag: a {POS} tagger for grammatical neologism detection}.
\newblock In \emph{Proceedings of the Eighth International Conference on
  Language Resources and Evaluation ({LREC}'12)}, pages 2118--2124, Istanbul,
  Turkey. European Language Resources Association (ELRA).

\bibitem[{Jiang et~al.(2023)Jiang, Sablayrolles, Mensch, Bamford, Chaplot,
  de~las Casas, Bressand, Lengyel, Lample, Saulnier, Lavaud, Lachaux, Stock,
  Scao, Lavril, Wang, Lacroix, and Sayed}]{jiang2023mistral}
Albert~Q. Jiang, Alexandre Sablayrolles, Arthur Mensch, Chris Bamford,
  Devendra~Singh Chaplot, Diego de~las Casas, Florian Bressand, Gianna Lengyel,
  Guillaume Lample, Lucile Saulnier, Lélio~Renard Lavaud, Marie-Anne Lachaux,
  Pierre Stock, Teven~Le Scao, Thibaut Lavril, Thomas Wang, Timothée Lacroix,
  and William~El Sayed. 2023.
\newblock \href {http://arxiv.org/abs/2310.06825} {Mistral 7b}.

\bibitem[{Jin et~al.(2022)Jin, Zhang, Zhu, Xiao, Li, Wei, Arnold, and
  Ren}]{jin-etal-2022-lifelong}
Xisen Jin, Dejiao Zhang, Henghui Zhu, Wei Xiao, Shang-Wen Li, Xiaokai Wei,
  Andrew Arnold, and Xiang Ren. 2022.
\newblock \href {https://doi.org/10.18653/v1/2022.bigscience-1.1} {Lifelong
  pretraining: Continually adapting language models to emerging corpora}.
\newblock In \emph{Proceedings of BigScience Episode {\#}5 -- Workshop on
  Challenges {\&} Perspectives in Creating Large Language Models}, pages 1--16,
  virtual+Dublin. Association for Computational Linguistics.

\bibitem[{Juraska et~al.(2023)Juraska, Finkelstein, Deutsch, Siddhant,
  Mirzazadeh, and Freitag}]{juraska-etal-2023-metricx}
Juraj Juraska, Mara Finkelstein, Daniel Deutsch, Aditya Siddhant, Mehdi
  Mirzazadeh, and Markus Freitag. 2023.
\newblock \href {https://doi.org/10.18653/v1/2023.wmt-1.63} {{M}etric{X}-23:
  The {G}oogle submission to the {WMT} 2023 metrics shared task}.
\newblock In \emph{Proceedings of the Eighth Conference on Machine
  Translation}, pages 756--767, Singapore. Association for Computational
  Linguistics.

\bibitem[{Kerremans et~al.(2018)Kerremans, Proki{\'{c}}, W\"{u}rschinger, and
  Schmid}]{Kerremans2018}
Daphn{\'{e}} Kerremans, Jelena Proki{\'{c}}, Quirin W\"{u}rschinger, and
  Hans-J\"{o}rg Schmid. 2018.
\newblock \href {https://doi.org/10.1075/pc.00006.ker} {Using data-mining to
  identify and study patterns in lexical innovation on the web}.
\newblock \emph{Pragmatics and Cognition}, 25(1):174--200.

\bibitem[{Langemets et~al.(2020)Langemets, Kallas, Norak, and
  Hein}]{Langemets2020}
Margit Langemets, Jelena Kallas, Kaisa Norak, and Indrek Hein. 2020.
\newblock \href {https://doi.org/10.1353/dic.2020.0005} {New estonian words and
  senses: Detection and description}.
\newblock \emph{Dictionaries: Journal of the Dictionary Society of North
  America}, 41(1):69--82.

\bibitem[{Lazaridou et~al.(2021)Lazaridou, Kuncoro, Gribovskaya, Agrawal,
  Liska, Terzi, Gimenez, d'Autume, Kocisky, Ruder, Yogatama, Cao, Young, and
  Blunsom}]{https://doi.org/10.48550/arxiv.2102.01951}
Angeliki Lazaridou, Adhiguna Kuncoro, Elena Gribovskaya, Devang Agrawal, Adam
  Liska, Tayfun Terzi, Mai Gimenez, Cyprien de~Masson d'Autume, Tomas Kocisky,
  Sebastian Ruder, Dani Yogatama, Kris Cao, Susannah Young, and Phil Blunsom.
  2021.
\newblock \href {https://doi.org/10.48550/ARXIV.2102.01951} {Mind the gap:
  Assessing temporal generalization in neural language models}.

\bibitem[{Lejeune and Cartier(2017)}]{lejeune-cartier-2017-character}
Ga{\"e}l Lejeune and Emmanuel Cartier. 2017.
\newblock \href {https://doi.org/10.18653/v1/W17-4103} {Character based pattern
  mining for neology detection}.
\newblock In \emph{Proceedings of the First Workshop on Subword and Character
  Level Models in {NLP}}, pages 25--30, Copenhagen, Denmark. Association for
  Computational Linguistics.

\bibitem[{Lewis et~al.(2019)Lewis, Liu, Goyal, Ghazvininejad, Mohamed, Levy,
  Stoyanov, and Zettlemoyer}]{lewis2019bart}
Mike Lewis, Yinhan Liu, Naman Goyal, Marjan Ghazvininejad, Abdelrahman Mohamed,
  Omer Levy, Ves Stoyanov, and Luke Zettlemoyer. 2019.
\newblock \href {http://arxiv.org/abs/1910.13461} {Bart: Denoising
  sequence-to-sequence pre-training for natural language generation,
  translation, and comprehension}.

\bibitem[{Li et~al.(2021)Li, Cheng, Huang, Chen, and Niu}]{Li2021}
Ying Li, Jiaxing Cheng, Cheng Huang, Zhouguo Chen, and Weina Niu. 2021.
\newblock \href {https://doi.org/10.1016/j.jisa.2021.102784} {{NEDetector}:
  Automatically extracting cybersecurity neologisms from hacker forums}.
\newblock \emph{Journal of Information Security and Applications}, 58:102784.

\bibitem[{Liu and Ritter(2023)}]{liu-ritter-2023-conll}
Shuheng Liu and Alan Ritter. 2023.
\newblock \href {https://aclanthology.org/2023.acl-long.459} {Do {C}o{NLL}-2003
  named entity taggers still work well in 2023?}
\newblock In \emph{Proceedings of the 61st Annual Meeting of the Association
  for Computational Linguistics (Volume 1: Long Papers)}, Toronto, Canada.
  Association for Computational Linguistics.

\bibitem[{Liu et~al.(2013)Liu, Hsieh, and
  Pr{\'{e}}vot}]{DBLP:conf/rocling/LiuHP13}
Tsun{-}Jui Liu, Shu{-}Kai Hsieh, and Laurent Pr{\'{e}}vot. 2013.
\newblock \href {https://aclanthology.org/O13-1025/} {Observing features of
  {PTT} neologisms: {A} corpus-driven study with n-gram model}.
\newblock In \emph{Proceedings of the 25th Conference on Computational
  Linguistics and Speech Processing, {ROCLING} 2015, National Sun Yat-sen
  University, Kaohsiung, Taiwan, October 4-5, 2013}. Association for
  Computational Linguistics and Chinese Language Processing (ACLCLP), Taiwan.

\bibitem[{Lochter et~al.(2022)Lochter, Silva, and Almeida}]{LOCHTER2022108911}
Johannes~V. Lochter, Renato~M. Silva, and Tiago~A. Almeida. 2022.
\newblock \href {https://doi.org/https://doi.org/10.1016/j.knosys.2022.108911}
  {Multi-level out-of-vocabulary words handling approach}.
\newblock \emph{Knowledge-Based Systems}, 251:108911.

\bibitem[{Lochter et~al.(2020)Lochter, Silva, and
  Almeida}]{DBLP:journals/corr/abs-2007-07318}
Johannes~V. Lochter, Renato~Moraes Silva, and Tiago~A. Almeida. 2020.
\newblock \href {http://arxiv.org/abs/2007.07318} {Deep learning models for
  representing out-of-vocabulary words}.
\newblock \emph{CoRR}, abs/2007.07318.

\bibitem[{Loureiro et~al.(2022)Loureiro, Barbieri, Neves, Anke, and
  Camacho-Collados}]{loureiro2022timelms}
Daniel Loureiro, Francesco Barbieri, Leonardo Neves, Luis~Espinosa Anke, and
  Jose Camacho-Collados. 2022.
\newblock \href {http://arxiv.org/abs/2202.03829} {Timelms: Diachronic language
  models from twitter}.

\bibitem[{Luu et~al.(2022)Luu, Khashabi, Gururangan, Mandyam, and
  Smith}]{luu-etal-2022-time}
Kelvin Luu, Daniel Khashabi, Suchin Gururangan, Karishma Mandyam, and Noah~A.
  Smith. 2022.
\newblock \href {https://doi.org/10.18653/v1/2022.naacl-main.435} {Time waits
  for no one! analysis and challenges of temporal misalignment}.
\newblock In \emph{Proceedings of the 2022 Conference of the North American
  Chapter of the Association for Computational Linguistics: Human Language
  Technologies}, pages 5944--5958, Seattle, United States. Association for
  Computational Linguistics.

\bibitem[{Margatina et~al.(2023)Margatina, Wang, Vyas, Anna~John, Benajiba, and
  Ballesteros}]{margatina-etal-2023-dynamic}
Katerina Margatina, Shuai Wang, Yogarshi Vyas, Neha Anna~John, Yassine
  Benajiba, and Miguel Ballesteros. 2023.
\newblock \href {https://doi.org/10.18653/v1/2023.eacl-main.211} {Dynamic
  benchmarking of masked language models on temporal concept drift with
  multiple views}.
\newblock In \emph{Proceedings of the 17th Conference of the European Chapter
  of the Association for Computational Linguistics}, pages 2881--2898,
  Dubrovnik, Croatia. Association for Computational Linguistics.

\bibitem[{McCrae(2019)}]{mccrae-2019-identification}
John~Philip McCrae. 2019.
\newblock \href {https://doi.org/10.18653/v1/W19-5116} {Identification of
  adjective-noun neologisms using pretrained language models}.
\newblock In \emph{Proceedings of the Joint Workshop on Multiword Expressions
  and WordNet (MWE-WN 2019)}, pages 135--141, Florence, Italy. Association for
  Computational Linguistics.

\bibitem[{Megerdoomian and Hadjarian(2010)}]{Megerdoomian2010MiningAC}
Karine Megerdoomian and Ali Hadjarian. 2010.
\newblock Mining and classification of neologisms in persian blogs.
\newblock In \emph{HLT-NAACL 2010}.

\bibitem[{Montani et~al.(2023)Montani, Honnibal, Honnibal, Boyd, Landeghem, and
  Peters}]{ines_montani_2023_10009823}
Ines Montani, Matthew Honnibal, Matthew Honnibal, Adriane Boyd, Sofie~Van
  Landeghem, and Henning Peters. 2023.
\newblock \href {https://doi.org/10.5281/zenodo.10009823} {{explosion/spaCy:
  v3.7.2: Fixes for APIs and requirements}}.

\bibitem[{Nayak et~al.(2020)Nayak, Timmapathini, Ponnalagu, and
  Gopalan~Venkoparao}]{nayak-etal-2020-domain}
Anmol Nayak, Hariprasad Timmapathini, Karthikeyan Ponnalagu, and Vijendran
  Gopalan~Venkoparao. 2020.
\newblock \href {https://doi.org/10.18653/v1/2020.insights-1.1} {Domain
  adaptation challenges of {BERT} in tokenization and sub-word representations
  of out-of-vocabulary words}.
\newblock In \emph{Proceedings of the First Workshop on Insights from Negative
  Results in NLP}, pages 1--5, Online. Association for Computational
  Linguistics.

\bibitem[{Onoe et~al.(2022)Onoe, Zhang, Choi, and
  Durrett}]{onoe-etal-2022-entity}
Yasumasa Onoe, Michael Zhang, Eunsol Choi, and Greg Durrett. 2022.
\newblock \href {https://doi.org/10.18653/v1/2022.findings-naacl.52} {Entity
  cloze by date: What {LM}s know about unseen entities}.
\newblock In \emph{Findings of the Association for Computational Linguistics:
  NAACL 2022}, pages 693--702, Seattle, United States. Association for
  Computational Linguistics.

\bibitem[{Papineni et~al.(2002)Papineni, Roukos, Ward, and
  Zhu}]{papineni-etal-2002-bleu}
Kishore Papineni, Salim Roukos, Todd Ward, and Wei-Jing Zhu. 2002.
\newblock \href {https://doi.org/10.3115/1073083.1073135} {{B}leu: a method for
  automatic evaluation of machine translation}.
\newblock In \emph{Proceedings of the 40th Annual Meeting of the Association
  for Computational Linguistics}, pages 311--318, Philadelphia, Pennsylvania,
  USA. Association for Computational Linguistics.

\bibitem[{Pinter et~al.(2020)Pinter, Jacobs, and
  Bittker}]{pinter-etal-2020-nytwit}
Yuval Pinter, Cassandra~L. Jacobs, and Max Bittker. 2020.
\newblock \href {https://doi.org/10.18653/v1/2020.coling-main.572} {{NYTWIT}: A
  dataset of novel words in the {N}ew {Y}ork {T}imes}.
\newblock In \emph{Proceedings of the 28th International Conference on
  Computational Linguistics}, pages 6509--6515, Barcelona, Spain (Online).
  International Committee on Computational Linguistics.

\bibitem[{Pramanick et~al.(2022)Pramanick, Beck, Stowe, and
  Gurevych}]{pramanick-etal-2022-challenges}
Aniket Pramanick, Tilman Beck, Kevin Stowe, and Iryna Gurevych. 2022.
\newblock \href {https://aclanthology.org/2022.findings-emnlp.195} {The
  challenges of temporal alignment on {T}witter during crises}.
\newblock In \emph{Findings of the Association for Computational Linguistics:
  EMNLP 2022}, pages 2658--2672, Abu Dhabi, United Arab Emirates. Association
  for Computational Linguistics.

\bibitem[{Pyo(2023)}]{korean-neologisms}
Jiyoon Pyo. 2023.
\newblock Detection and replacement of neologisms for translation.
\newblock Master's thesis, The Cooper Union for the Advancement of Science and
  Art.

\bibitem[{Raffel et~al.(2020)Raffel, Shazeer, Roberts, Lee, Narang, Matena,
  Zhou, Li, and Liu}]{JMLR:v21:20-074}
Colin Raffel, Noam Shazeer, Adam Roberts, Katherine Lee, Sharan Narang, Michael
  Matena, Yanqi Zhou, Wei Li, and Peter~J. Liu. 2020.
\newblock \href {http://jmlr.org/papers/v21/20-074.html} {Exploring the limits
  of transfer learning with a unified text-to-text transformer}.
\newblock \emph{Journal of Machine Learning Research}, 21(140):1--67.

\bibitem[{Rei et~al.(2020)Rei, Stewart, Farinha, and Lavie}]{rei2020comet}
Ricardo Rei, Craig Stewart, Ana~C Farinha, and Alon Lavie. 2020.
\newblock \href {http://arxiv.org/abs/2009.09025} {Comet: A neural framework
  for mt evaluation}.

\bibitem[{Rei et~al.(2022)Rei, Treviso, Guerreiro, Zerva, Farinha, Maroti,
  C.~de Souza, Glushkova, Alves, Coheur, Lavie, and
  Martins}]{rei-etal-2022-cometkiwi}
Ricardo Rei, Marcos Treviso, Nuno~M. Guerreiro, Chrysoula Zerva, Ana~C Farinha,
  Christine Maroti, Jos{\'e}~G. C.~de Souza, Taisiya Glushkova, Duarte Alves,
  Luisa Coheur, Alon Lavie, and Andr{\'e} F.~T. Martins. 2022.
\newblock \href {https://aclanthology.org/2022.wmt-1.60} {{C}omet{K}iwi:
  {IST}-unbabel 2022 submission for the quality estimation shared task}.
\newblock In \emph{Proceedings of the Seventh Conference on Machine Translation
  (WMT)}, pages 634--645, Abu Dhabi, United Arab Emirates (Hybrid). Association
  for Computational Linguistics.

\bibitem[{Rijhwani and
  Preotiuc-Pietro(2020)}]{rijhwani-preotiuc-pietro-2020-temporally}
Shruti Rijhwani and Daniel Preotiuc-Pietro. 2020.
\newblock Temporally-informed analysis of named entity recognition.
\newblock In \emph{Proceedings of the 58th Annual Meeting of the Association
  for Computational Linguistics}, Online. Association for Computational
  Linguistics.

\bibitem[{R{\"o}ttger and
  Pierrehumbert(2021)}]{rottger-pierrehumbert-2021-temporal-adaptation}
Paul R{\"o}ttger and Janet Pierrehumbert. 2021.
\newblock \href {https://doi.org/10.18653/v1/2021.findings-emnlp.206} {Temporal
  adaptation of {BERT} and performance on downstream document classification:
  Insights from social media}.
\newblock In \emph{Findings of the Association for Computational Linguistics:
  EMNLP 2021}, pages 2400--2412, Punta Cana, Dominican Republic. Association
  for Computational Linguistics.

\bibitem[{Ryskina et~al.(2020)Ryskina, Rabinovich, Berg-Kirkpatrick, Mortensen,
  and Tsvetkov}]{ryskina-etal-2020-new}
Maria Ryskina, Ella Rabinovich, Taylor Berg-Kirkpatrick, David Mortensen, and
  Yulia Tsvetkov. 2020.
\newblock \href {https://aclanthology.org/2020.scil-1.43} {Where new words are
  born: Distributional semantic analysis of neologisms and their semantic
  neighborhoods}.
\newblock In \emph{Proceedings of the Society for Computation in Linguistics
  2020}, pages 367--376, New York, New York. Association for Computational
  Linguistics.

\bibitem[{Taori et~al.(2023)Taori, Gulrajani, Zhang, Dubois, Li, Guestrin,
  Liang, and Hashimoto}]{alpaca}
Rohan Taori, Ishaan Gulrajani, Tianyi Zhang, Yann Dubois, Xuechen Li, Carlos
  Guestrin, Percy Liang, and Tatsunori~B. Hashimoto. 2023.
\newblock Stanford alpaca: An instruction-following llama model.
\newblock \url{https://github.com/tatsu-lab/stanford_alpaca}.

\bibitem[{Touvron et~al.(2023{\natexlab{a}})Touvron, Lavril, Izacard, Martinet,
  Lachaux, Lacroix, Rozière, Goyal, Hambro, Azhar, Rodriguez, Joulin, Grave,
  and Lample}]{touvron2023llama}
Hugo Touvron, Thibaut Lavril, Gautier Izacard, Xavier Martinet, Marie-Anne
  Lachaux, Timothée Lacroix, Baptiste Rozière, Naman Goyal, Eric Hambro,
  Faisal Azhar, Aurelien Rodriguez, Armand Joulin, Edouard Grave, and Guillaume
  Lample. 2023{\natexlab{a}}.
\newblock \href {http://arxiv.org/abs/2302.13971} {Llama: Open and efficient
  foundation language models}.

\bibitem[{Touvron et~al.(2023{\natexlab{b}})Touvron, Martin, Stone, Albert,
  Almahairi, Babaei, Bashlykov, Batra, Bhargava, Bhosale, Bikel, Blecher,
  Ferrer, Chen, Cucurull, Esiobu, Fernandes, Fu, Fu, Fuller, Gao, Goswami,
  Goyal, Hartshorn, Hosseini, Hou, Inan, Kardas, Kerkez, Khabsa, Kloumann,
  Korenev, Koura, Lachaux, Lavril, Lee, Liskovich, Lu, Mao, Martinet, Mihaylov,
  Mishra, Molybog, Nie, Poulton, Reizenstein, Rungta, Saladi, Schelten, Silva,
  Smith, Subramanian, Tan, Tang, Taylor, Williams, Kuan, Xu, Yan, Zarov, Zhang,
  Fan, Kambadur, Narang, Rodriguez, Stojnic, Edunov, and
  Scialom}]{touvron2023llama2}
Hugo Touvron, Louis Martin, Kevin Stone, Peter Albert, Amjad Almahairi, Yasmine
  Babaei, Nikolay Bashlykov, Soumya Batra, Prajjwal Bhargava, Shruti Bhosale,
  Dan Bikel, Lukas Blecher, Cristian~Canton Ferrer, Moya Chen, Guillem
  Cucurull, David Esiobu, Jude Fernandes, Jeremy Fu, Wenyin Fu, Brian Fuller,
  Cynthia Gao, Vedanuj Goswami, Naman Goyal, Anthony Hartshorn, Saghar
  Hosseini, Rui Hou, Hakan Inan, Marcin Kardas, Viktor Kerkez, Madian Khabsa,
  Isabel Kloumann, Artem Korenev, Punit~Singh Koura, Marie-Anne Lachaux,
  Thibaut Lavril, Jenya Lee, Diana Liskovich, Yinghai Lu, Yuning Mao, Xavier
  Martinet, Todor Mihaylov, Pushkar Mishra, Igor Molybog, Yixin Nie, Andrew
  Poulton, Jeremy Reizenstein, Rashi Rungta, Kalyan Saladi, Alan Schelten, Ruan
  Silva, Eric~Michael Smith, Ranjan Subramanian, Xiaoqing~Ellen Tan, Binh Tang,
  Ross Taylor, Adina Williams, Jian~Xiang Kuan, Puxin Xu, Zheng Yan, Iliyan
  Zarov, Yuchen Zhang, Angela Fan, Melanie Kambadur, Sharan Narang, Aurelien
  Rodriguez, Robert Stojnic, Sergey Edunov, and Thomas Scialom.
  2023{\natexlab{b}}.
\newblock \href {http://arxiv.org/abs/2307.09288} {Llama 2: Open foundation and
  fine-tuned chat models}.

\bibitem[{Vu et~al.(2023)Vu, Iyyer, Wang, Constant, Wei, Wei, Tar, Sung, Zhou,
  Le, and Luong}]{vu2023freshllms}
Tu~Vu, Mohit Iyyer, Xuezhi Wang, Noah Constant, Jerry Wei, Jason Wei, Chris
  Tar, Yun-Hsuan Sung, Denny Zhou, Quoc Le, and Thang Luong. 2023.
\newblock \href {http://arxiv.org/abs/2310.03214} {Freshllms: Refreshing large
  language models with search engine augmentation}.

\bibitem[{Wang and Komatsuzaki(2021)}]{gpt-j}
Ben Wang and Aran Komatsuzaki. 2021.
\newblock {GPT-J-6B: A 6 Billion Parameter Autoregressive Language Model}.
\newblock \url{https://github.com/kingoflolz/mesh-transformer-jax}.

\bibitem[{Wang et~al.(2019)Wang, Cho, and
  Gu}]{DBLP:journals/corr/abs-1909-03341}
Changhan Wang, Kyunghyun Cho, and Jiatao Gu. 2019.
\newblock \href {http://arxiv.org/abs/1909.03341} {Neural machine translation
  with byte-level subwords}.
\newblock \emph{CoRR}, abs/1909.03341.

\bibitem[{Wei et~al.(2022)Wei, Bosma, Zhao, Guu, Yu, Lester, Du, Dai, and
  Le}]{wei2022finetuned}
Jason Wei, Maarten Bosma, Vincent~Y. Zhao, Kelvin Guu, Adams~Wei Yu, Brian
  Lester, Nan Du, Andrew~M. Dai, and Quoc~V. Le. 2022.
\newblock \href {http://arxiv.org/abs/2109.01652} {Finetuned language models
  are zero-shot learners}.

\bibitem[{Xu et~al.(2023)Xu, Kim, Sharaf, and Awadalla}]{xu2023paradigm}
Haoran Xu, Young~Jin Kim, Amr Sharaf, and Hany~Hassan Awadalla. 2023.
\newblock \href {http://arxiv.org/abs/2309.11674} {A paradigm shift in machine
  translation: Boosting translation performance of large language models}.

\bibitem[{Zalmout et~al.(2019)Zalmout, Thadani, and
  Pappu}]{zalmout-etal-2019-unsupervised}
Nasser Zalmout, Kapil Thadani, and Aasish Pappu. 2019.
\newblock \href {https://doi.org/10.18653/v1/D19-5555} {Unsupervised neologism
  normalization using embedding space mapping}.
\newblock In \emph{Proceedings of the 5th Workshop on Noisy User-generated Text
  (W-NUT 2019)}, pages 425--430, Hong Kong, China. Association for
  Computational Linguistics.

\bibitem[{Zhao et~al.(2022)Zhao, Chrysostomou, Bontcheva, and
  Aletras}]{zhao-etal-2022-impact}
Zhixue Zhao, George Chrysostomou, Kalina Bontcheva, and Nikolaos Aletras. 2022.
\newblock \href {https://doi.org/10.18653/v1/2022.findings-emnlp.298} {On the
  impact of temporal concept drift on model explanations}.
\newblock In \emph{Findings of the Association for Computational Linguistics:
  EMNLP 2022}, pages 4039--4054, Abu Dhabi, United Arab Emirates. Association
  for Computational Linguistics.

\bibitem[{Zhu and Jurgens(2021)}]{DBLP:journals/corr/abs-2104-05010}
Jian Zhu and David Jurgens. 2021.
\newblock \href {http://arxiv.org/abs/2104.05010} {The structure of online
  social networks modulates the rate of lexical change}.
\newblock \emph{CoRR}, abs/2104.05010.

\end{thebibliography}

\appendix
\begin{table*}[t]
\setlength{\tabcolsep}{3pt}
\centering
\small
\resizebox{0.99\textwidth}{!}{%
\begin{tabular}{lccccccccc}
\toprule
& \multicolumn{4}{c}{\textbf{Neologism Type}} & & \multicolumn{4}{c}{\textbf{Collection Method}} \\\cmidrule{2-5} \cmidrule{7-10}
\textbf{Dataset} & \textbf{Emerging?} & \textbf{Multiword?} & \textbf{Semantic?} & \textbf{Generalized?} & & \textbf{Exclusion Lists} & \textbf{Dictionaries} & \textbf{Time-Series} & \textbf{Templates} \\
\midrule
\cite{pinter-etal-2020-nytwit} & {\textcolor{red} {\xmark}} & {\textcolor{red} {\xmark}} & {\textcolor{red} {\xmark}} & {\textcolor{lightgreen} {\cmark}} & & {\textcolor{lightgreen} {\cmark}} &  {\textcolor{red} {\xmark}} & {\textcolor{red} {\xmark}} & {\textcolor{red} {\xmark}} \\
\cite{Kerremans2018} & {\textcolor{red} {\xmark}} & {\textcolor{red} {\xmark}} & {\textcolor{red} {\xmark}} & {\textcolor{lightgreen} {\cmark}} & & {\textcolor{lightgreen} {\cmark}} &  {\textcolor{red} {\xmark}} & {\textcolor{red} {\xmark}} & {\textcolor{red} {\xmark}} \\
\cite{zalmout-etal-2019-unsupervised} & {\textcolor{red} {\xmark}} & {\textcolor{red} {\xmark}} & {\textcolor{red} {\xmark}} & {\textcolor{lightgreen} {\cmark}} & & {\textcolor{lightgreen} {\cmark}} &  {\textcolor{red} {\xmark}} & {\textcolor{red} {\xmark}} & {\textcolor{red} {\xmark}} \\
\cite{janssen-2012-neotag} & {\textcolor{red} {\xmark}} & {\textcolor{red} {\xmark}} & {\textcolor{lightgreen} {\cmark}} & {\textcolor{red} {\xmark}} & &  {\textcolor{lightgreen} {\cmark}} &  {\textcolor{red} {\xmark}} & {\textcolor{red} {\xmark}} & {\textcolor{red} {\xmark}} \\
\cite{dhuliawala-etal-2016-slangnet} & {\textcolor{red} {\xmark}} & {\textcolor{lightgreen} {\cmark}} & {\textcolor{red} {\xmark}} & {\textcolor{red} {\xmark}} & & {\textcolor{red} {\xmark}} & {\textcolor{lightgreen} {\cmark}} & {\textcolor{red} {\xmark}} & {\textcolor{red} {\xmark}} \\
\cite{DBLP:journals/corr/abs-2104-05010} & {\textcolor{red} {\xmark}} & {\textcolor{lightgreen} {\cmark}} & {\textcolor{red} {\xmark}} & {\textcolor{lightgreen} {\cmark}}&   & {\textcolor{red} {\xmark}} & {\textcolor{lightgreen} {\cmark}} & {\textcolor{red} {\xmark}} & {\textcolor{red} {\xmark}}\\
\cite{mccrae-2019-identification} & {\textcolor{red} {\xmark}} & {\textcolor{lightgreen} {\cmark}} & {\textcolor{red} {\xmark}} & {\textcolor{lightgreen} {\cmark}}  & & {\textcolor{red} {\xmark}} & {\textcolor{lightgreen} {\cmark}} & {\textcolor{red} {\xmark}} & {\textcolor{red} {\xmark}} \\
\cite{broad-etal-2018-candidate} & {\textcolor{lightgreen} {\cmark}} & {\textcolor{red} {\xmark}} & {\textcolor{red} {\xmark}} & {\textcolor{lightgreen} {\cmark}} & & {\textcolor{red} {\xmark}} & {\textcolor{red} {\xmark}} & {\textcolor{lightgreen} {\cmark}} & {\textcolor{red} {\xmark}} \\
\cite{Li2021} & {\textcolor{red} {\xmark}} & {\textcolor{lightgreen} {\cmark}} & {\textcolor{red} {\xmark}} & {\textcolor{red} {\xmark}} &  & {\textcolor{red} {\xmark}} & {\textcolor{red} {\xmark}} & {\textcolor{lightgreen} {\cmark}} & {\textcolor{red} {\xmark}} \\
\midrule
{\sc Neo-Bench} (this work) & {\textcolor{lightgreen} {\cmark}} & {\textcolor{lightgreen} {\cmark}} & {\textcolor{lightgreen} {\cmark}} & {\textcolor{lightgreen} {\cmark}} & &  {\textcolor{lightgreen} {\cmark}} & {\textcolor{lightgreen} {\cmark}} &  {\textcolor{lightgreen} {\cmark}} & {\textcolor{lightgreen} {\cmark}} \\
\bottomrule
\end{tabular}}
\vspace{-.1in}
\caption{Comparison of English Neologism resources by the types of neologisms collected and the collection method used. {\sc Neo-Bench} covers more types of neologisms by using more methods than prior neologism datasets.}
\vspace{-.4cm}
\label{tab:survey} 
\end{table*}

\section{Related Work}
Table \ref{tab:survey} provides an overview and comparison of English neologism resources, including the types of neologisms and collection methods used in each dataset. {\sc Neo-Bench} uses more collection methods and covers more types of neologisms, including multiword and semantic neologisms.

\section{Data Collection}
\begin{table}
\centering
\small
\setlength{\tabcolsep}{4pt}
\resizebox{0.48\textwidth}{!}{
\begin{tabular}{l|cccc}
\toprule
 \textbf{Source} & \textbf{Reddit} & \textbf{News} & \textbf{NYTimes} & \textbf{Dictionary}\\ 
\textbf{\# Candidates} & 74542 & 60671 & 6908 & 80071 \\\midrule
\textbf{\% Reddit} & - & 0.93\% & 0.94\% & 3.91\% \\
\textbf{\% News} & 0.76\% & - & 0.26\% & 0.12\% \\
\textbf{\% NYTimes} & 0.09\% & 0.03\% & - & 0.15\% \\
\textbf{\% Dictionary} & 4.19\% & 0.16\% & 1.80\% & - \\\midrule
\textbf{\% Total} & 5.04\% & 1.12\% & 3.00\% & 4.18 \% \\
\bottomrule 
\end{tabular}
}
\caption{Number of shared neologism candidates for each method pair. The overlap is reported as a percent of the total number of candidates for each method.}
\label{tab:overlap}
\end{table}
We base the categories of neologisms on the linguistic taxonomy used in previous literature. Based on our empirical studies, we observe that all neologisms fall under three broad categories: lexical, morphological, and semantic.  We additionally label neologisms based on subcategories of these broad linguistic classes (e.g., word blends, derivations, acronyms, and novel phrases).

For Reddit neologism candidates, we collected 500 million utterances in December 2021 and 200 million utterances from January to May 2022. We tokenize the utterances with the NLTK package \cite{bird2009natural} to get individual word counts and update a generic word counter. Neologism candidates are selected by filtering out typos and extremely rare words with less than a frequency of 50. We further filter out named entities by utilizing a SpaCy named entity recognition (NER) model \cite{spacy2} (\texttt{en\_core\_web\_sm}) to detect proper nouns and update a named entity counter. We compare the counts of words from the general counter and the named entity counter and filter out the word if the proportion that a general word is in a named entity is greater than 0.5. The remaining words with the lowest frequencies are the list of uncommon words that we treat as neologism candidates for a given month. In total, we collect 74,542 neologism candidates.

\begin{figure}[!h]
\resizebox{0.47\textwidth}{!}{
\begin{tikzpicture}
\pie [rotate = 0,
text = legend,
sum=auto,
every only number node/.style={text=white},
color={color1bg,color3bg,color5bg,color6bg,color7bg},] {
    1005/Reddit,
    1000/News Articles,
    200/NYTimes,
    200/Dictionaries,
    100/Handpicked}
\node[above] at (0, 3) {\textbf{{\sc Neo-Bench}}};
\end{tikzpicture}
}

\resizebox{0.47\textwidth}{!}{
\begin{tikzpicture}
\pie [name=plot1, scale=0.5, rotate = 0,
text = \empty,
sum=auto,
every only number node/.style={text=white},
color={pie1,pie2,pie3},] {
    310,
    588,
    107}

\node[above] at (0, 1.5) {\textbf{Reddit}};
\node[above] at (3.5, 1.5) {\textbf{News Articles}};
\node[above] at (7, 1.5) {\textbf{NYTimes}};

\pie [scale=0.5, pos={7,0}, rotate = 0,
text = \empty,
sum=auto,
every only number node/.style={text=white},
color={pie1,pie2,pie3},] {
    750,
    182,
    68}
\pie [scale=0.5, rotate = 0, pos={14,0},
text = \empty,
sum=auto,
every only number node/.style={text=white},
color={pie2,pie3},] {
    192,
    8}
\pie[sum=200, scale=0.5, pos={14,0}, color={pie2,pie3}, hide number]{192/, 8/8}
\pie [scale=0.5, sum=200, pos={14,0}, color={pie2,pie3}] {
    192/}

\end{tikzpicture}
}

\resizebox{0.46\textwidth}{!}{
\begin{tikzpicture}
\pie [name=plot1, scale=0.5, rotate = 0,
text = \empty,
sum=auto,
every only number node/.style={text=white},
color={pie1,pie2,pie3},] {
    4,
    194,
    2}

\pie[sum=200, scale=0.5, color={pie1,pie2,pie3}, hide number]{4/4, 194/, 2/2}
\pie[sum=200, rotate=7, scale=0.5, color={pie2}]{194/}

\pie [scale=0.5, pos={7,0}, rotate = 0,
text = legend,
sum=auto,
every only number node/.style={text=white},
color={pie1,pie2,pie3},] {
    28/Lexical,
    40/Morphological,
    32/Semantic}
\node[above] at (0, 1.5) {\textbf{Dictionaries}};
\node[above] at (3.5, 1.4) {\textbf{Handpicked}};
\end{tikzpicture}
}
\caption{Breakdown of {\sc Neo-Bench} by collection method. Each method is further stratified by the linguistic type of neologisms.}
\label{fig:neobench_pie_chart}
\end{figure}
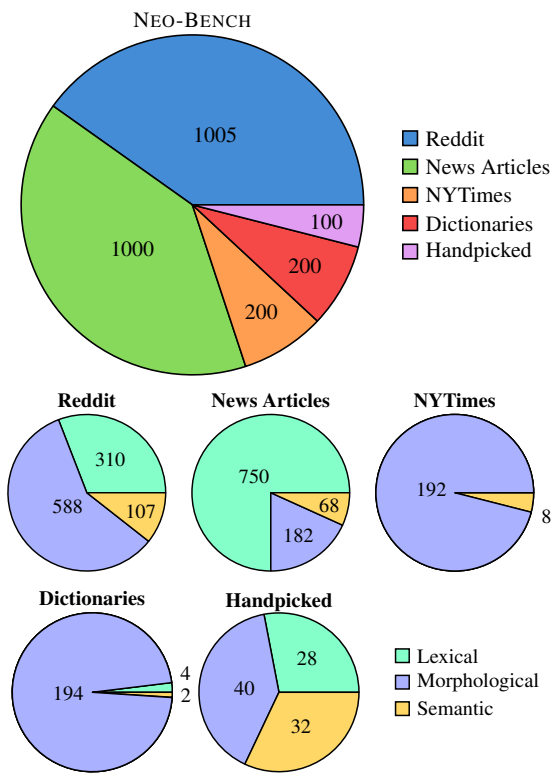

For news articles, we use a script to collect 11,412 headlines from Google News from 2019-2023. In total, we get 60,671 noun and verb phrases with a Part-of-Speech Tagger via SpaCy (\texttt{en\_core\_web\_sm}) that we treat as neologism candidates. We use an old dataset of 80,071 neologisms obtained from two slang dictionaries \cite{DBLP:journals/corr/abs-2104-05010} and sample 200 neologisms with interesting or no trend lines. Table \ref{tab:overlap} provides the breakdown of method overlap between each method pair in {\sc Neo-Bench}. Instead of the sample of 1,100 data points, we compare a total of all 6,908 words tweeted out by the NYT First Said bot from 2020 to 2023 with the other methods. Figure \ref{fig:neobench_pie_chart} provides the breakdown of {\sc Neo-Bench} by collection method and linguistic type.

\subsection{Google Trends Filtering}\label{sec:trends_method_explained}
We collect Google Trends monthly data from January 2010 to July 2023. While Google Trends provides data from 2004, there are inconsistencies in word usage frequencies until 2010.
To compare word prevalence between neologisms, we make a pairwise comparison of a neologism candidate with the misspelling `dangrous', which provides a consistent baseline comparison for word usage data. We then use this normalized trend line for neologism candidate filtering.

In total, we create five differing methods that use a combination of filtering criteria, including curve-fitting, argmax detection, integral, line of best-fit, and maximum trend data values, to evaluate words as neologism candidates. We select the best combination based on which method yields both high precision and estimated recall in collecting neologisms. Using 20,000 words collected in February 2022, we filter this set through all five methods which filter out almost 90\% of words. We sample each method for 100 candidates and manually annotate the samples for neologism classification, obtaining a sampled precision of each method. We combine all the neologisms from the manually-annotated samples to obtain a computationally derived neologism set. We evaluate each method for its estimated recall based on the proportion of words from the computational neologism sample that is not filtered out. The sample precision is particularly low given the sparsity of neologisms that appear at a specific point in time, so we select the method with the highest precision of 0.2 and an estimated recall of 0.625 to reduce the amount of manual annotation required.

\subsection{Dataset Analysis}
With the best-performing filter method, we also estimate its recall with a set of 100 handpicked neologisms in our dataset. The estimated recall of our method is 0.55. This estimate is slightly lower than the estimated recall we used with the neologism candidates we computationally gathered, but our collection methods remain consistent. 

Analyzing the overlap of words in our dataset with Urban Dictionary by collection method, we find that 44.37\% of Reddit neologisms, 38.4\% of News neologisms, and 65.25\% of neologisms obtained from dictionaries and exclusion lists overlap.

\subsection{Emergence Date Labeling for Perplexity-Based Rankings}
\label{sec:emergence_dates}

Using Google Trends to record the month and year where word usage spikes, we estimate the date for when a neologism enters the dissemination stage of its lifecycle. We find that 68\% of neologisms emerged during 2020-2023, and 17\% of all words have no dissemination date or trend line. The remaining words were prevalent before 2020, but a new connotation or usage has recently emerged. These dates are potentially inaccurate as a trend line is collapsed into a single date. Compared to using the entire graph to evaluate long-term word trends when filtering for neologisms, a single date may not perfectly capture neologism growth for words that exhibit a steady rise in growth.

\subsection{{\sc Neo-Bench} Tasks}
We sample 750 neologisms from our set of 2505 total neologisms. We stratify the sample based on the 5 collection methodologies used: handpicked, Reddit, News articles, NYT First Said Bot, and Slang dictionaries. We collect 500 neologisms from Reddit and News Articles, 100 from NYT First Said Bot, 100 from slang dictionaries, and 50 from our handpicked set.

We work with 3 in-house native English speakers to construct inputs for Cloze question answering, definition generation, and human-evaluated machine translation. Using a script, annotators are provided with Google Search Descriptions containing neologisms. With these reference sentences, annotators were instructed to create 750 multi-sentence Cloze passages and to create 4 multiple-choice options based on topical relevance to the neologism and 1 distractor answer that is a feasible substitute in the passage. Annotators also constructed 750 sentences containing neologisms and 750 questions asking for the definition of neologisms. From a sub-sample of 100 neologisms, annotators constructed minimal pair sentences using Google Search references for human-evaluated machine translation. Annotators were instructed to create any feasible substitute word, regardless of topic, to replace the neologism in the minimal pair sentence. For perplexity rankings, we use 422 Cloze passages that have both a single-word neologism and distractor answer. 

We additionally subsample 240 sentences containing neologisms for automatic machine translation. We work with 3 in-house native Chinese speakers to construct 240 reference translations for automatic MT evaluation. Annotators were provided the input sentence, the neologism, and the definition and were instructed to create fluent and accurate translations. If no associated term in Chinese exists, annotators were instructed to paraphrase sentences or use substitute terms with minimal information loss. Annotators were then instructed to retranslate the Chinese sentence back into English to enable further study on accurate translation techniques of neologisms, including paraphrasing the sentence. All 6 annotators we work with have a college-level education. 

\section{Experimental Details}
All models are evaluated on two NVIDIA A40 GPUs for a single run since models are not finetuned for {\sc Neo-Bench} tasks.

\subsection{Machine Translation}\label{sec:appendix_mt}
For human-evaluated MT, our error categorization is partially adapted from the widely-used MQM framework \cite{Freitag_2021}, which is adjusted based on the pilot studies we conducted on translating sentences containing neologism words. Translation errors are ordered by severity in affecting the understanding of the sentence, and annotators label sentences based on the most severe translation error. For instance, if there are grammatical mistakes and English words in the translation output, the output will be labeled “Copy” since that is the most severe translation error. Table \ref{tab:translation_examples} provides example outputs for each translation category. 

\begin{figure*}[!tbp]
    \centering
    \includegraphics[width=0.9\textwidth]{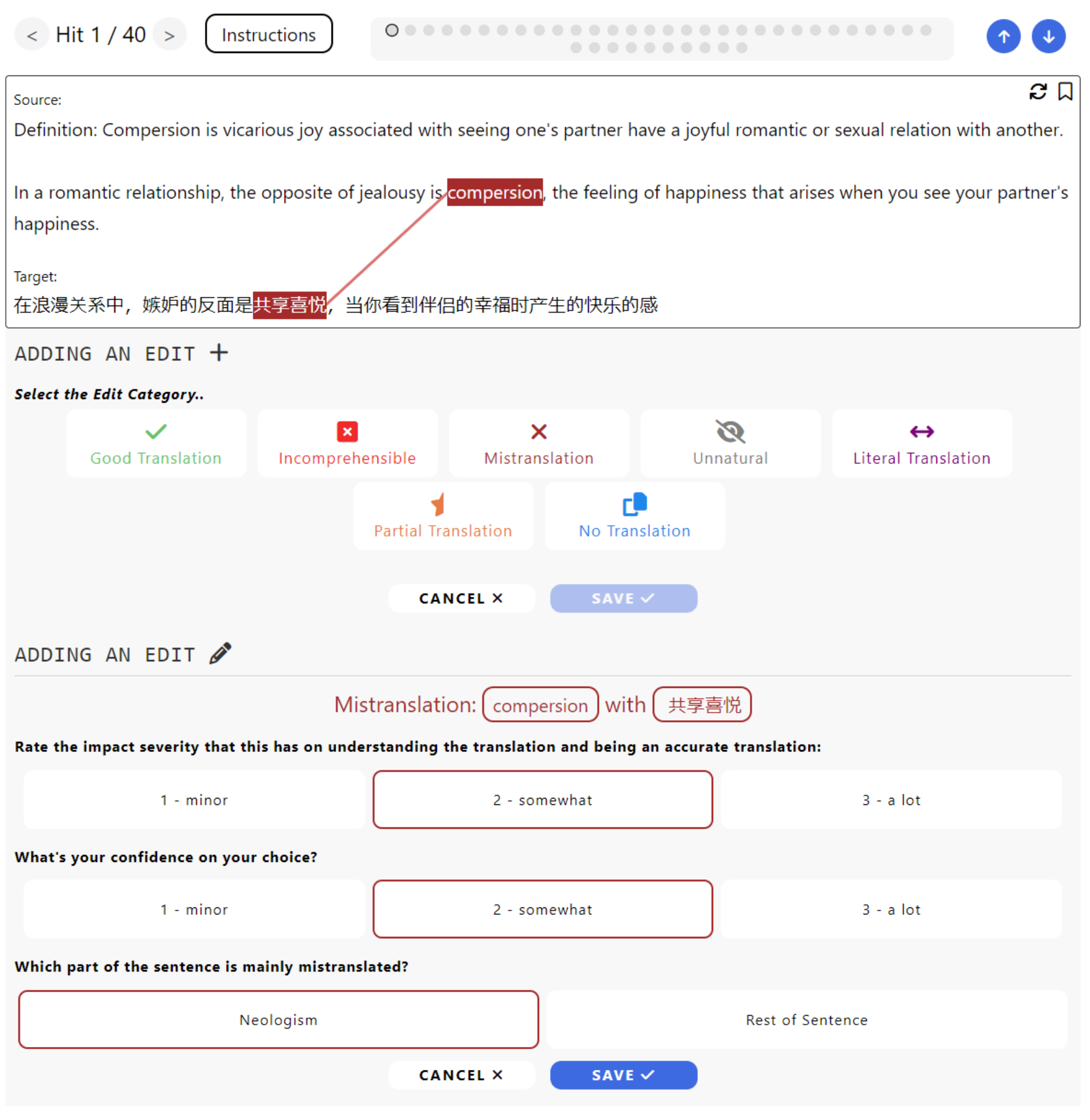}
\vspace*{-2mm}
  \caption{Thresh Interface used to crowdsource human annotations of Machine Translation Output.}
    \label{fig:thresh_interface}
    \vspace*{-2mm}
\end{figure*}

We crowdsource annotations by screening for 5 fluent Mandarin speakers on Prolific\footnote{https://www.prolific.com}, a crowdsourcing website. All 5 annotators are fluent in English and Mandarin and reside in the United States and United Kingdom. All of the annotators have a college education and are informed about the nature of the study. Each annotator was given the same set of neologism sentences across all 5 models evaluated to ensure a standard comparison between models. Annotators were provided with neologism definitions and were instructed to select the label corresponding to the worst translation error in the sentence and highlight the corresponding spans in the input sentence and MT output. Annotators were also instructed to label the error severity and the confidence in their selection of translation categories on 3-point Likert scales. Finally, annotators marked if the translation error occurred from the neologism or from another portion of the sentence. The average time to annotate 20 minimal pairs was 80 minutes, and each annotator was paid \$12.00 an hour, which is on the high end for standard pay on Prolific. We use the Thresh \cite{heineman2023thresh} interface for annotating translation sentences, and Figure \ref{fig:thresh_interface} provides a screenshot of the interface. 
\newline\indent
Based on our human reference translations of neologisms, we find that translation is difficult. For neologisms representing new concepts, there may be information loss if there is not an associated word in the target language (e.g. boyflux $\rightarrow$ nonbinary). There are often no exact words associated with neologisms in the target language, so translation requires providing the neologism definition and rephrasing the entire sentence (e.g. fossilflation $\rightarrow$ fossil fuel price increases). However, since neologisms are created with the same linguistic origins as common words, the novelty of neologisms results in lower MT performance.


\subsection{Rank Classification with Perplexity}
\label{rank_appendix}

We also tested T5 and Flan-T5 for perplexity ranking and find that Flan-T5 exhibits higher neologism rankings than T5. However, when sorting words by lowest perplexity and filling in the mask, we find that these models produced entirely incoherent sequences, so we do not report these models. Given the computational intensity of evaluating 5,002 sequence perplexities, we only evaluate the base size of models.

\subsection{Cloze Question Answering}
\label{sec:cloze_appendix}

Rank classification is used to select the lowest perplexity answer for BART, T5, and GPT-J.  For other models, we shuffle the order of answers and conduct experiments with 5-shot prompting with the following format:
\newline

\noindent\texttt{Fill in the blank with the options below:}

\noindent\texttt{Question: [EXAMPLE CLOZE PASSAGE]}

\noindent\texttt{a) [EXAMPLE INCORRECT ANSWER]}

\noindent\texttt{b) [EXAMPLE DISTRACTOR ANSWER]}

\noindent\texttt{c) [EXAMPLE INCORRECT ANSWER]}

\noindent\texttt{d) [EXAMPLE INCORRECT ANSWER]}

\noindent\texttt{e) [EXAMPLE NEOLOGISM ANSWER]}

\noindent\texttt{f) [EXAMPLE INCORRECT ANSWER]}

\noindent\texttt{Answer: e) [EXAMPLE NEOLOGISM ANSWER]}

\noindent\texttt{...}

\noindent\texttt{Fill in the blank with the options below:}

\noindent\texttt{Question: [TEST CLOZE PASSAGE]}

\noindent\texttt{a) [TEST INCORRECT ANSWER]}

\noindent\texttt{b) [TEST INCORRECT ANSWER]}

\noindent\texttt{c) [TEST NEOLOGISM ANSWER]}

\noindent\texttt{d) [TEST DISTRACTOR ANSWER]}

\noindent\texttt{e) [EXAMPLE INCORRECT ANSWER]}

\noindent\texttt{f) [EXAMPLE INCORRECT ANSWER]}

\noindent\texttt{Answer:}

\subsection{Definition Generation}
\label{sec:opendef_appendix}
We conduct experiments with 5-shot prompting with the following prompt:\newline\newline
\noindent\texttt{Answer the question.}

\noindent\texttt{Question: [EXAMPLE DEFINITION QUESTION]}

\noindent\texttt{Answer: [EXAMPLE DEFINITION ANSWER]}

\noindent\texttt{...}

\noindent\texttt{Answer the question.}

\noindent\texttt{Question: [TEST CLOZE PASSAGE]}

\noindent\texttt{Question: [TEST DEFINITION QUESTION]}

\noindent\texttt{Answer:}
\newline\newline One of the paper's authors manually annotates 100 outputs. We measure the Cohen's Kappa between automatic GPT-4 and human evaluation, and we obtain an average Cohen's Kappa of 0.744, indicating high agreement between human judgment and GPT-4.

For automatic evaluation, we additionally use GPT-4 to determine if a correct model definition is better or worse than the reference definition provided by human input. For incorrect answers, we separate between incorrect and omitted generations, which are model outputs that are either left blank or, for GPT models, outputs where the model acknowledges that it does not recognize the neologism. 

Table \ref{tab:opendomain_full} provides the full results of the open-domain question-answering experiments, including the average length of definitions, manual evaluation and model-wise Cohen's Kappa. While LLaMA-2 70B outperforms GPT-3.5 in Cloze QA, GPT 3.5 produces more correct definitions than LLaMA-2-70B. Instruction-tuned models produce a higher proportion of correct answers that are deemed better than the human reference sentence. We report the average length of generations for each model and conclude that GPT-4 prefers instruction model outputs because the human reference sentences are on average 19.20 words long, which is more concise compared to the more elaborative responses of instruct-tuned models. We find that 81.07\% of preferred answers across all models are longer than the alternative correct answer. Table \ref{tab:length_instruct} provides examples of instruct-tuned model responses that are evaluated as better than the human reference and examples of GPT-omitted responses. Even when prompted with shortened answers, instruction-tuned models produce longer-form responses. We did not test for restraining the length of the model output as the complexity and reference definition length for each neologism varies extensively. Instruction-tuned models are less likely to produce omitted answers, and only GPT-3.5 and GPT-4 have generations that acknowledge that a term is unrecognized. 

\begin{table}[t]
\centering
\setlength{\tabcolsep}{4pt}
\scriptsize
\renewcommand{\arraystretch}{0.85}
\resizebox{0.48\textwidth}{!}{
\begin{tabular}{l|ccc}
\toprule
& \multicolumn{3}{c}{\textbf{Pre-trained}} \\
Model & Lexical & Morphological & Semantic \\ \midrule
BART-Large & 19.88 & 18.48 & 14.94 \\
T5-Large & 33.54 & 39.88 & 39.08 \\
GPTJ 6B & 34.78 & 25.22 & 36.78 \\
LLaMA-1 7B & 29.50 & 27.86 & 29.89 \\
OLMo 7B & 26.40 & 28.45 & 28.74 \\
Mistral 7B & 69.57 & 65.69 & 70.11 \\
LLaMA-2 7B & 67.39 & 58.36 & 62.07 \\
LLaMA-2 13B & 69.25 & 61.58 & 62.07 \\
LLaMA-2 70B & \textbf{77.95} & \textbf{71.55} & \underline{77.01} \\
GPT 3.5 & 69.25 & 69.79 & \underline{77.01} \\
\midrule
& \multicolumn{3}{c}{\textbf{Instruction-Tuned}} \\
Model & Lexical & Morphological & Semantic \\ \midrule
Flan-T5 Large & 41.99 & 42.99 & 41.61 \\
Alpaca 7B & 50.37 & 44.87 & 42.99 \\
OLMo Instruct 7B & 60.45 & 47.99 & 55.07 \\
Mistral Instruct 7B & 61.18 & 57.16 & 51.62 \\
LLaMA-2 Chat 7B & 62.64 & 57.42 & 62.76 \\ 
LLaMA-2 Chat 13B & 68.63 & 61.58 & 62.07 \\ 
LLaMA-2 Chat 70B & 71.43 & 64.22 & 72.41 \\
GPT-4 & \textbf{83.85} & \textbf{80.06} & \textbf{87.36} \\\bottomrule
\end{tabular}
}
\caption{Neologism accuracies of models for the Cloze Question Answering task separated by linguistic type: lexical (322), morphological (341), and semantic (87). Best performing accuracy is presented in \textbf{bold}, while highest shared accuracy is reported in \underline{underline}.
}\label{tab:cloze_linguistic} 
\end{table}

\section{Linguistic Taxonomy}
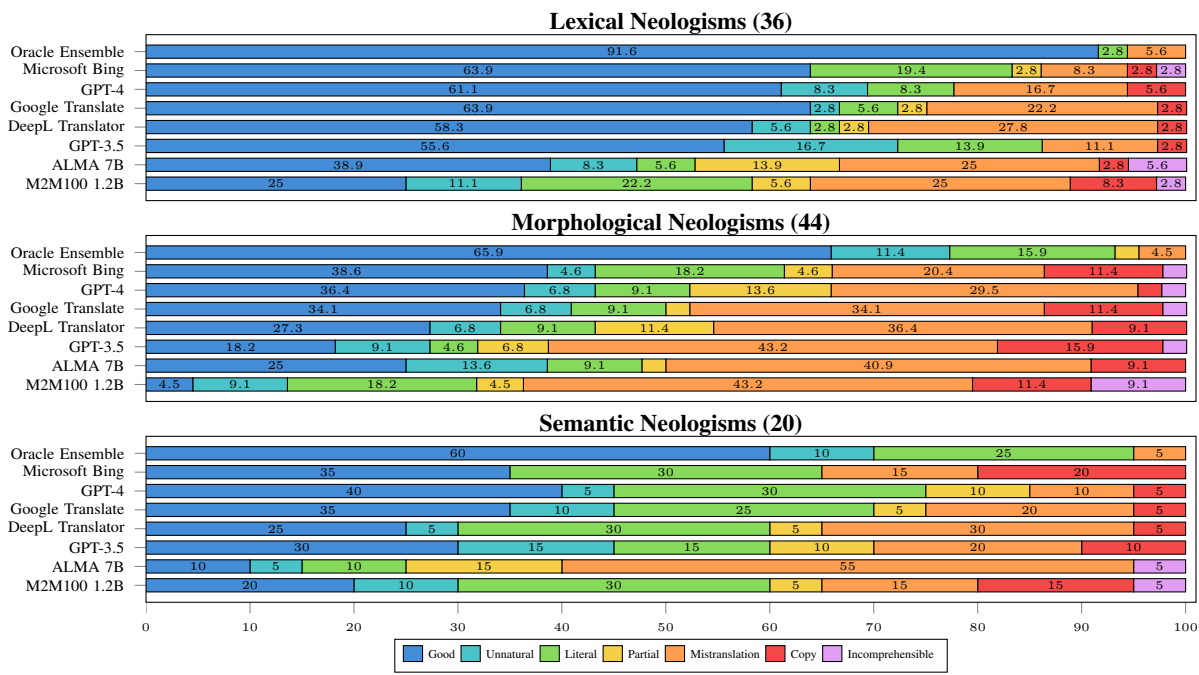
\begin{figure*}[!ht]
\pgfplotsset{
    testbar11/.style={
        xbar stacked,
        bar width=5pt,
        xtick pos=lower, ytick pos=left,
        xmin=0,xmax=101.5,
        ytick = data,
        ytick distance=1,
        yticklabel style={text width=1.6cm, align=right, yshift=0cm, font=\fontsize{6}{6}\selectfont},
        ytick style={yshift=0cm},
        legend style={at={(0.3,0.05)},anchor=west},
        tick align = outside,
        height=0.83\textwidth,
        xlabel style = {font=\small},
        y=2.5mm,
        enlarge y limits={abs=0.9},
    }}
\begin{tikzpicture}
\tikzstyle{every node}=[font=\fontsize{4.5}{5}\selectfont]
\begin{axis}[testbar11, xmax=101, name=plot1, xtick=\empty,  anchor=north west, yticklabels = {M2M100 1.2B, ALMA 7B, GPT-3.5, DeepL Translator, Google Translate, GPT-4, Microsoft Bing, Oracle Ensemble}, legend style={at={(0.5,0.125), font=\footnotesize,nodes={scale=0.5}},fill=none,draw=none,anchor=north},
title style={at={(0.5,1.10)}, font=\small,anchor=north,yshift=-0.1},
title = \textbf{Lexical Neologisms (36)}, legend columns=-1, enlarge y limits={abs=0.9},
] 
 \addplot[fill=color1bg, nodes near coords] coordinates{
                                    (25, 0)
                                    (38.9, 1)
                                    (55.6, 2)
                                    (58.3, 3)
                                    (63.9, 4)
                                    (61.1, 5)
                                    (63.9, 6)
                                    (91.6, 7)
                                    };
 \addplot[fill=color2bg, nodes near coords] coordinates{
                                   (11.1, 0)
                                   (8.3, 1)
                                   (16.7, 2)
                                   (5.6, 3)
                                   (2.8, 4)
                                   (8.3, 5)
                                   (0, 6)
                                   (0, 7)
                                   };

 \addplot[fill=color3bg, nodes near coords] coordinates{
                                   (22.2, 0)
                                   (5.6, 1)
                                   (13.9, 2)
                                   (2.8, 3)
                                   (5.6, 4)
                                   (8.3, 5)
                                   (19.4, 6)  
                                   (2.8, 7)
                                   };
                                   
 \addplot[fill=color4bg, nodes near coords] coordinates{
                                   (5.6, 0)
                                   (13.9, 1)
                                   (0, 2)
                                   (2.8, 3)
                                   (2.8, 4)
                                   (0, 5)
                                   (2.8, 6)
                                   (0, 7)
                                   };
                                   
 \addplot[fill=color5bg, nodes near coords] coordinates{
                                   (25, 0)
                                   (25, 1)
                                   (11.1, 2)
                                   (27.8, 3)
                                   (22.2, 4)
                                   (16.7, 5)
                                   (8.3, 6)
                                   (5.6, 7)
                                   };

 \addplot[fill=color6bg, nodes near coords] coordinates{
                                   (8.3, 0)
                                   (2.8, 1)
                                   (2.8, 2)
                                   (2.8, 3)
                                   (2.8, 4)
                                   (5.6, 5)
                                   (2.8, 6)
                                   (0, 7)
                                   };

 \addplot[fill=color7bg, nodes near coords] coordinates{
                                   (2.8, 0)
                                   (5.6, 1)
                                   (0, 2)
                                   (0, 3)
                                   (0, 4)
                                   (0, 5)
                                   (2.8, 6)
                                   (0, 7)
                                   };
\end{axis}
\end{tikzpicture}
\begin{tikzpicture}
\tikzstyle{every node}=[font=\fontsize{4.5}{5}\selectfont]
\begin{axis}[testbar11,xmax=101, name=plot2, anchor=north west,xtick=\empty,  yticklabels = {M2M100 1.2B, ALMA 7B, GPT-3.5, DeepL Translator, Google Translate, GPT-4, Microsoft Bing, Oracle Ensemble}, legend style={at={(0.5,0.125), font=\footnotesize,nodes={scale=0.5}},fill=none,draw=none,anchor=north},
title style={at={(0.5,1.10)}, font=\small,anchor=north,yshift=-0.1},
title = \textbf{Morphological Neologisms (44)}, legend columns=-1, enlarge y limits={abs=0.9},
] 
 \addplot[fill=color1bg, nodes near coords] coordinates{
                                    (4.5, 0)
                                    (25, 1)
                                    (18.2, 2)
                                    (27.3, 3)
                                    (34.1, 4)
                                    (36.4, 5)
                                    (38.6, 6)
                                    (65.9, 7)
                                    };
 \addplot[fill=color2bg, nodes near coords] coordinates{
                                   (9.1, 0)
                                   (13.6, 1)
                                   (9.1, 2)
                                   (6.8, 3)
                                   (6.8, 4)
                                   (6.8, 5)
                                   (4.6, 6)
                                   (11.4, 7)
                                   };

 \addplot[fill=color3bg, nodes near coords] coordinates{
                                   (18.2, 0)
                                   (9.1, 1)
                                   (4.6, 2)
                                   (9.1, 3)
                                   (9.1, 4)
                                   (9.1, 5)
                                   (18.2, 6)
                                   (15.9, 7)
                                   };
                                   
 \addplot[fill=color4bg, nodes near coords] coordinates{
                                   (4.5, 0)
                                   (0, 1)
                                   (6.8, 2)
                                   (11.4, 3)
                                   (0, 4)
                                   (13.6, 5)
                                   (4.6, 6)
                                   (0, 7)
                                   };

 \addplot[fill=color4bg] coordinates{
                                   (0, 0)
                                   (2.3, 1)
                                   (0, 2)
                                   (0, 3)
                                   (2.3, 4)
                                   (0, 5)
                                   (0, 6)
                                   (2.3, 7)
                                   };
                                   
 \addplot[fill=color5bg, nodes near coords] coordinates{
                                   (43.2, 0)
                                   (40.9, 1)
                                   (43.2, 2)
                                   (36.4, 3)
                                   (34.1, 4)
                                   (29.5, 5)
                                   (20.4, 6)
                                   (4.5, 7)
                                   };

 \addplot[fill=color6bg, nodes near coords] coordinates{
                                   (11.4, 0)
                                   (9.1, 1)
                                   (15.9, 2)
                                   (9.1, 3)
                                   (11.4, 4)
                                   (0, 5)
                                   (11.4, 6)
                                   (0, 7)
                                   };

 \addplot[fill=color6bg] coordinates{
                                   (0, 0)
                                   (0, 1)
                                   (0, 2)
                                   (0, 3)
                                   (0, 4)
                                   (2.3, 5)
                                   (0, 6)
                                   (0, 7)
                                   };

 \addplot[fill=color7bg, nodes near coords] coordinates{
                                   (9.1, 0)
                                   (0, 1)
                                   (0, 2)
                                   (0, 3)
                                   (0, 4)
                                   (0, 5)
                                   (0, 6)
                                   (0, 7)
                                   };

 \addplot[fill=color7bg] coordinates{
                                   (0, 0)
                                   (0, 1)
                                   (2.3, 2)
                                   (0, 3)
                                   (2.3, 4)
                                   (2.3, 5)
                                   (2.3, 6)
                                   (0, 7)
                                   };
\end{axis}
\end{tikzpicture}

\begin{tikzpicture}
\tikzstyle{every node}=[font=\fontsize{4.5}{5}\selectfont]
\begin{axis}[testbar11, xmax=101, name=plot3, anchor=north west, yticklabels = {M2M100 1.2B, ALMA 7B, GPT-3.5, DeepL Translator, Google Translate, GPT-4, Microsoft Bing, Oracle Ensemble}, legend style={at={(0.5,-0.22), font=\footnotesize,nodes={scale=0.5}},anchor=north},
title style={at={(0.5,1.10)}, font=\small,anchor=north,yshift=-0.1},
title = \textbf{Semantic Neologisms (20)}, legend columns=-1, enlarge y limits={abs=0.9},
] 
 \addplot[fill=color1bg, nodes near coords] coordinates{
                                    (20, 0)
                                    (10, 1)
                                    (30, 2)
                                    (25, 3)
                                    (35, 4)
                                    (40, 5)
                                    (35, 6)
                                    (60, 7)
                                    };
 \addplot[fill=color2bg, nodes near coords] coordinates{
                                   (10, 0)
                                   (5, 1)
                                   (15, 2)
                                   (5, 3)
                                   (10, 4)
                                   (5, 5)
                                   (0, 6)
                                   (10,7)
                                   };

 \addplot[fill=color3bg, nodes near coords] coordinates{
                                   (30, 0)
                                   (10, 1)
                                   (15, 2)
                                   (30, 3)
                                   (25, 4)
                                   (30, 5)
                                   (30, 6)
                                   (25, 7)
                                   };
                                   
 \addplot[fill=color4bg, nodes near coords] coordinates{
                                   (5, 0)
                                   (15, 1)
                                   (10, 2)
                                   (5, 3)
                                   (5, 4)
                                   (10, 5)
                                   (0, 6)
                                   (0, 7)
                                   };
                                   
 \addplot[fill=color5bg, nodes near coords] coordinates{
                                   (15, 0)
                                   (55, 1)
                                   (20, 2)
                                   (30, 3)
                                   (20, 4)
                                   (10, 5)
                                   (15, 6)
                                   (5, 7)
                                   };

 \addplot[fill=color6bg, nodes near coords] coordinates{
                                   (15, 0)
                                   (0, 1)
                                   (10, 2)
                                   (5, 3)
                                   (5, 4)
                                   (5, 5)
                                   (20, 6)
                                   (0, 7)
                                   };

 \addplot[fill=color7bg, nodes near coords] coordinates{
                                   (5, 0)
                                   (5, 1)
                                   (0, 2)
                                   (0, 3)
                                   (0, 4)
                                   (0, 5)
                                   (0, 6)
                                   (0, 7)
                                   };

\legend{Good, Unnatural, Literal, Partial, Mistranslation, Copy, Incomprehensible}
\end{axis}
\end{tikzpicture}

\caption{Results of the Machine Translation task with human-annotated labels for each linguistic type of neologism. Results are reported as percentages of the total number of neologisms of each linguistic category (provided in the titles). A Human Oracle Ensemble selecting the best model translation for each sentence is provided.}
\label{fig:mt_linguistic}
\end{figure*}

Table \ref{tab:cloze_linguistic} provides the linguistic breakdown of neologism accuracies for each model in Cloze QA. Table \ref{tab:definition_linguistic} provides the stratified results of definition generation by linguistic type of neologism. Tables \ref{tab:mt_automatic_linguistic_xmetric} and \ref{tab:mt_automatic_linguistic_comet} provides the breakdown of automatic machine translation evaluation, and Figure \ref{fig:mt_linguistic} provides the stratified results of manually labeling translations by linguistic type. We report each category as a proportion to the total amount of neologisms for a certain linguistic type. Model performance discrepancy between the open source models and commercial systems is highest for lexical neologisms. For automatic metrics, lexical neologisms do not yield higher BLEU scores but yield higher COMET and COMETKiwi scores. Morphological neologisms yield 5.3 lower BLEU scores, 3.4\% lower COMET scores, and 4.4\% lower COMETKiwi scores than lexical neologisms. BLEU scores for semantic neologisms are high as there is high token overlap between different senses of the same word form. However, semantic neologisms often yield similarly low COMET and COMETKiwi scores as Morphological neologisms.

\begin{table*}[t]
\setlength{\tabcolsep}{5pt}
\centering
\small
\resizebox{0.99\textwidth}{!}{%
\begin{tabular}{lccccccccccccccc}
\toprule
& \multicolumn{3}{c}{\textbf{MetricX-23$_{\text{XXL}}$}} & & \multicolumn{3}{c}{\textbf{MetricX-23-QE$_{\text{XXL}}$}} & & \multicolumn{3}{c}{\textbf{MetricX-23$_{\text{XL}}$}} & & \multicolumn{3}{c}{\textbf{MetricX-23-QE$_{\text{XL}}$}}  \\\cmidrule{2-4} \cmidrule{6-8} \cmidrule{10-12} \cmidrule{14-16}
\textbf{Model} & \textbf{Lex.} & \textbf{Morph.} & \textbf{Sem.} & & \textbf{Lex.} & \textbf{Morph.} & \textbf{Sem.} & & \textbf{Lex.} & \textbf{Morph.} & \textbf{Sem.} & & \textbf{Lex.} & \textbf{Morph.} & \textbf{Sem.} \\
\midrule
Google Translate & 1.494 & 2.087 & 1.864 & & 1.126 & 1.762 & \textbf{1.180} & & 1.642 & 2.145 & 2.099 & & 1.742 & \textbf{2.129} & \textbf{1.867} \\
Bing Translator & 1.889 & 2.725 & 2.821 & & 1.265 & 2.011 & 1.711 & & 1.999 & 2.501 & 2.722 & & 1.987 & 2.450 & 2.278 \\
DeepL Translator & 1.507 & 1.892 & 2.089 & & 1.042 & \textbf{1.455} & 1.220 & & 1.815 & \textbf{1.846} & 2.282 & & \textbf{1.648} & 2.195 & 1.924 \\
GPT-4  & 1.428 & \textbf{1.612} & \textbf{1.667} & & 1.275 & 1.532 & 1.520 & & 1.739 & 1.776 & 1.978 & & 1.804 & 2.278 & 2.297 \\
GPT-3.5 & \textbf{1.197} & 1.995 & 2.095 & & \textbf{1.035} & 1.779 & 1.606 & & \textbf{1.528} & 1.915 & 2.109 & & 1.656 & 2.481 & 2.430 \\
ALMA-7 B & 2.025 & 2.553 & 2.758 & & 1.710 & 2.246 & 2.233 & & 1.950 & 2.489 & 2.284 & & 2.172 & 2.719 & 2.413 \\
M2M100 1.2B & 2.764 & 3.917 & 3.779 & & 2.204 & 3.370 & 2.704 & & 2.647 & 3.553 & 3.138 & & 2.487 & 3.373 & 3.004 \\
\bottomrule
\end{tabular}}
\vspace{-.1in}
\caption{MetricX-23 and MetricX-23-QE scores of Machine Translation models evaluated on neologisms, separated by linguistic type of neologisms and aggregate score.}
\label{tab:mt_automatic_linguistic_xmetric} 
\end{table*}

\begin{table*}[t]
\setlength{\tabcolsep}{9pt}
\centering
\small
\resizebox{0.99\textwidth}{!}{%
\begin{tabular}{lccccccccccc}
\toprule
& \multicolumn{3}{c}{\textbf{COMET}} & & \multicolumn{3}{c}{\textbf{COMETKiwi}} & & \multicolumn{3}{c}{\textbf{BLEU}}  \\\cmidrule{2-4} \cmidrule{6-8} \cmidrule{10-12}
\textbf{Model} & \textbf{Lex.} & \textbf{Morph.} & \textbf{Sem.} & & \textbf{Lex.} & \textbf{Morph.} & \textbf{Sem.} & & \textbf{Lex.} & \textbf{Morph.} & \textbf{Sem.} \\
\midrule
Google Translate & \textbf{0.870} & 0.842 & 0.849 & & 0.820 & 0.782 & 0.805 & & \textbf{0.530} & \textbf{0.487} & \textbf{0.507} \\
Bing Translator & 0.852 & 0.806 & 0.812 & & 0.812 & 0.769 & 0.786 & & 0.484 & 0.418 & 0.467 \\
DeepL Translator & 0.856 & 0.833 & 0.833 & & 0.823 & \textbf{0.792} & \textbf{0.814} & & 0.434 & 0.373 & 0.429 \\
GPT-4 & 0.866 & \textbf{0.845} & \textbf{0.852} & & 0.814 & 0.782 & 0.776 & & 0.466 & 0.414 & 0.491 \\
GPT-3.5 & 0.859 & 0.833 & 0.820 & & \textbf{0.824} & 0.773 & 0.769 & & 0.425 & 0.365 & 0.425 \\
ALMA-7 B & 0.814 & 0.795 & 0.786 & & 0.770 & 0.731 & 0.730 & & 0.303 & 0.262 & 0.287 \\
M2M100 1.2B & 0.816 & 0.743 & 0.774 & & 0.786 & 0.715 & 0.730 & & 0.357 & 0.310 & 0.361 \\
\bottomrule
\end{tabular}}
\vspace{-.1in}
\caption{COMET and BLEU scores of Machine Translation models evaluated on neologisms, separated by linguistic type of neologisms and aggregate score.}
\label{tab:mt_automatic_linguistic_comet} 
\end{table*}

\begin{table*}[t]
\centering
\scriptsize
\setlength{\tabcolsep}{3pt}
\renewcommand{\arraystretch}{0.93}
\resizebox{0.88\textwidth}{!}{
\begin{tabular}{l|crr|crr|crr}
\toprule
\multirow{2}{*}{\textbf{Pre-trained Models}} & \multicolumn{3}{c|}{\textbf{Lexical (322)}} & \multicolumn{3}{c|}{\textbf{Morphological (341)}} & \multicolumn{3}{c}{\textbf{Semantic (87)}} \\
 & Correct & \% Worse & \% Better & Correct & \% Worse & \% Better & Correct & \% Worse & \% Better \\ \midrule
GPTJ 6B & 0.367 & 78.7\% & 21.3\% & 0.323 & 68.1\% & 31.9\% & 0.322 & 78.6\% & 21.4\% \\
LLaMA-1 7B & 0.506 & 72.9\% & 27.1\% & 0.437 & 68.4\% & 31.6\% & 0.345 & 63.2\% & 36.8\% \\
OLMo 7B & 0.674 & 52.1\% & 47.9\% & 0.557 & 49.5\% & 50.5\% & 0.529 & 58.7\% & 41.3\% \\
Mistral 7B & 0.661 & 59.6\% & 40.4\% & 0.628 & 62.6\% & 37.4\% & 0.494 & 69.8\% & 30.2\% \\
LLaMA-2 7B & 0.587 & 65.1\% & 34.9\% & 0.481 & 58.0\% & 42.0\% & 0.506 & 72.7\% & 27.3\% \\
LLaMA-2 13B & 0.609 & 62.7\% & 37.3\% & 0.510 & 62.2\% & 37.8\% & 0.425 & 64.9\% & 35.1\% \\
LLaMA-2 70B & 0.665 & 57.4\% & 42.6\% & 0.613 & 54.0\% & 46.0\% & 0.483 & 57.1\% & 42.9\% \\
GPT 3.5 & 0.817 & 6.1\% & 93.9\% & 0.686 & 9.8\% & 90.2\% & 0.713 & 11.4\% & 88.6\% \\
\midrule
\multirow{2}{*}{\textbf{Instruct-tuned Models}} & \multicolumn{3}{c|}{\textbf{Lexical (322)}} & \multicolumn{3}{c|}{\textbf{Morphological (341)}} & \multicolumn{3}{c}{\textbf{Semantic (87)}} \\
 & Correct & \% Worse & \% Better & Correct & \% Worse & \% Better & Correct & \% Worse & \% Better \\ \midrule
Flan-T5 Large & 0.158 & 96.2\% & 3.8\% & 0.158 & 83.5\% & 16.5\% & 0.080 & 86.3\% & 13.7\%  \\
Alpaca 7B & 0.565 & 71.0\% & 29.0\% & 0.463 & 70.8\% & 29.2\% & 0.414 & 63.8\% & 36.2\% \\
OLMo Instruct 7B & 0.804 & 41.7\% & 58.3\% & 0.581 & 34.3\% & 65.7\% & 0.471 & 31.7\% & 68.3\% \\
Mistral Instruct 7B  & 0.767 & 47.8\% & 52.5\% & 0.581 & 39.4\% & 60.6\% & 0.471 & 29.3\% & 70.7\% \\
LLaMA-2 Chat 7B & 0.581 & 59.4\% & 40.6\% & 0.528 & 63.3\% & 36.7\% & 0.448 & 66.7\% & 33.3\% \\ 
LLaMA-2 Chat 13B & 0.649 & 40.7\% & 59.3\% & 0.543 & 41.1\% & 58.9\% & 0.494 & 39.5\% & 60.5\% \\ 
LLaMA-2 Chat 70B & 0.661 &  46.0\% & 54.0\% & 0.566 & 44.5\% & 55.5\% & 0.494 & 37.2\% & 62.8\% \\
GPT-4 & 0.870 & 8.3\% & 91.7\% & 0.827 & 11.4\% & 88.6\% & 0.736 & 11.0\% & 89.0\% \\\bottomrule
\end{tabular}
}
\caption{Results of the definition generation task when separated by linguistic type. Accuracy is reported as a proportion of correct answers compared to the total number of neologisms of each linguistic type. The percentages of correct answers that are labeled as 'worse' and 'better' than the human reference sentence by GPT-4 are provided. }\label{tab:definition_linguistic} 
\end{table*}

\begin{table*}[t]
\begin{CJK*}{UTF8}{gbsn}
\small
\centering
\renewcommand{\arraystretch}{0.7}
\begin{tabular}{P{0.11\textwidth}P{0.84\textwidth}}
\toprule
\multicolumn{2}{c}{\fontsize{9}{10.5}\selectfont \textbf{Translation Output Examples}}\vspace{2pt}\\\toprule
\multirow{10.5}{*}{\textbf{Fire Weather}} & {\fontsize{8}{8}\selectfont  {\textbf{Input:} They will not issue official warnings until \textbf{fire weather} is forecast to occur.}}\vspace{2pt}\\
& \hspace{23pt} {\fontsize{7}{7.5}\selectfont \it (Fire weather is the use of meteorological parameters such as relative humidity, wind speed, mixing heights, and soil moisture} \vspace{1pt}\\
& \hspace{23pt} {\fontsize{7}{7.5}\selectfont \it to determine whether conditions are favorable for fire growth and smoke dispersion.)}
\\\cmidrule{2-2}
& {\fontsize{8}{8}\selectfont \textbf{Model Output} (\hspace{0.083333em}\sethlcolor{color1bg}\textbf{\textcolor{black}{\hl{Good}}\hspace{0.083333em}): 在预测发生\underline{火灾天气}之前，他们不会发布官方警告。}} \vspace{3pt}\\
& \hspace{80pt} {\fontsize{7}{7.5}\selectfont  \it (Before predicting the occurrence of \textbf{fire weather}, they will not issue an official warning.)}
\\\cmidrule{2-2}
& {\fontsize{8}{8}\selectfont \textbf{Human Translation:} 他们直到预测到\underline{火灾天气}才会发布官方警告。} \vspace{2pt}\\
& \hspace{70pt} {\fontsize{7}{7.5}\selectfont  \it (They will only issue official warnings when a \textbf{fire weather} is forecasted.) } \vspace{3pt}\\\toprule

\multirow{9.5}{*}{\textbf{Dupe}} & {\fontsize{8}{8}\selectfont  {\textbf{Input:} Discover new affordable \textbf{dupes} for luxury expensive makeup products.}} \vspace{2pt}\\
& \hspace{23pt} {\fontsize{7}{7.5}\selectfont  \it (A dupe is an abbreviation of the word "duplicate".)} \\\cmidrule{2-2}
& {\fontsize{8}{8}\selectfont \textbf{Model Output} (\hspace{0.083333em}\sethlcolor{color2bg}\textbf{\textcolor{black}{\hl{Unnatural}}\hspace{0.083333em}):  为奢华昂贵的彩妆产品发现新的负担得起的\underline{复制品}。}} \vspace{3pt}\\
& \hspace{96pt} {\fontsize{7}{7.5}\selectfont \it (Find new affordable \textbf{replicas} for luxurious and expensive makeup products.)}
\\\cmidrule{2-2}
& {\fontsize{8}{8}\selectfont \textbf{Human Translation:} 探索奢侈昂贵化妆品的平价\underline{替代新品}。} \vspace{2pt}\\
& \hspace{70pt} {\fontsize{7}{7.5}\selectfont \it (Explore new affordable \textbf{alternatives} for luxurious and expensive makeup products.)} \vspace{3pt}\\\toprule

\multirow{9}{*}{\textbf{Snowvember}} & {\fontsize{8}{8}\selectfont  {\textbf{Input:} When sleet started falling during Thanksgiving it was officially \textbf{snowvember}.}}\vspace{2pt}\\
& \hspace{23pt} {\fontsize{7}{7.5}\selectfont  \it (Snowvember refers to a particular November that experiences a lot of snowfall.)}
\\\cmidrule{2-2}
& {\fontsize{8}{8}\selectfont \textbf{Model Output} (\hspace{0.083333em}\sethlcolor{color3bg}\textbf{\textcolor{black}{\hl{Literal}}\hspace{0.083333em}):  当感恩节期间开始下雨夹雪时，正式进入了\underline{十一月雪}。}} \vspace{3pt}\\
& \hspace{85pt} {\fontsize{7}{7.5}\selectfont  \it (When it starts to rain and snow during Thanksgiving, it officially enters the \textbf{November snow}.)}
\\\cmidrule{2-2}
& {\fontsize{8}{8}\selectfont \textbf{Human Translation:} 感恩节开始下的雨夹雪标志着‘雪月’的正式开始。} \vspace{2pt}\\
& \hspace{70pt} {\fontsize{7}{7.5}\selectfont  \it (The start of sleet during Thanksgiving marks the official beginning of \textbf{'Snowvember'.}) } \vspace{3pt}\\\toprule

\multirow{9}{*}{\textbf{Trollbaiting}} & {\fontsize{8}{8}\selectfont  {\textbf{Input:} \textbf{Trollbaiting} has caused my growth on social media this past month to be super high.}} \vspace{2pt}\\
& \hspace{23pt} {\fontsize{7}{7.5}\selectfont  \it (Trollbaiting describes when an internet user knowingly invites the hatred of a highly reactionary group of trolls.)} \\\cmidrule{2-2}
& {\fontsize{8}{8}\selectfont \textbf{Model Output} (\hspace{0.083333em}\sethlcolor{color4bg}\textbf{\textcolor{black}{\hl{Partial}}\hspace{0.083333em}):   在过去的一个月里，我在社交媒体上的增长速度超快。}} \vspace{3pt}\\
& \hspace{84pt} {\fontsize{7}{7.5}\selectfont  \it (In the past month, my growth on social media has been extremely fast.)}
\\\cmidrule{2-2}
& {\fontsize{8}{8}\selectfont \textbf{Human Translation:} 过去一个月里，通过\underline{挑衅网络喷子}，我在社交媒体上的增长极为迅速。} \vspace{2pt}\\
& \hspace{70pt} {\fontsize{7}{7.5}\selectfont  \it (In the past month, by \textbf{provoking internet trolls}, my growth on social media has been extremely rapid.)} \vspace{3pt}\\\toprule

\multirow{13}{0.13\textwidth}{\textbf{Forever Chemicals}} & {\fontsize{8}{8}\selectfont  {\textbf{Input:} The environment cannot break down \textbf{forever chemicals}, and they will remain in our bodies for years if ingested.}} \vspace{2pt}\\
& \hspace{23pt} {\fontsize{7}{7.5}\selectfont  \it (Forever chemicals are used to make products grease-proof, water-proof, stick-proof, and stain-resistant and are toxic to humans} \vspace{1pt}\\
& \hspace{23pt} {\fontsize{7}{7.5}\selectfont \it and nearly indestructible.)} \\\cmidrule{2-2}
& {\fontsize{8}{8}\selectfont \textbf{Model Output} (\hspace{0.083333em}\sethlcolor{color5bg}\textbf{\textcolor{black}{\hl{Mistranslation}}\hspace{0.083333em}):  环境不能\underline{永远分解化学物质}，如果摄入它们，它们会在我们体内停留数年。}} \vspace{3pt}\\
& \hspace{111pt} {\fontsize{7}{7.5}\selectfont \it (The environment cannot \textbf{always break down chemicals}, and if ingested, they can stay in our} \vspace{1pt}\\
& \hspace{111pt} {\fontsize{7}{7.5}\selectfont \it bodies for years.)}
\\\cmidrule{2-2}
& {\fontsize{8}{8}\selectfont \textbf{Human Translation:} 环境无法分解\underline{永久化学物质}，一旦摄入，这些物质将在我们的身体中残留多年。} \vspace{2pt}\\
& \hspace{70pt} {\fontsize{7}{7.5}\  \selectfont  \it (The environment cannot break down \textbf{'forever chemicals',} and once ingested, these substances will remain in}\vspace{1pt}\\
& \hspace{70pt} {\fontsize{7}{7.5} \selectfont \it our bodies for many years.)} \vspace{3pt}\\\toprule

\multirow{10}{*}{\textbf{Blud}} & {\fontsize{8}{8}\selectfont  {\textbf{Input:} What is \textbf{blud} talking about I can't understand.}}\vspace{2pt}\\ 
& \hspace{23pt} {\fontsize{7}{7.5}\selectfont \it (Blud is slang that is used to address men and means bro.) } \\\cmidrule{2-2}
& {\fontsize{8}{8}\selectfont \textbf{Model Output} (\hspace{0.083333em}\sethlcolor{color6bg}\textbf{\textcolor{black}{\hl{Copy}}\hspace{0.083333em}):   \underline{blud} 在说什么我听不懂。}} \vspace{3pt}\\
& \hspace{79pt} {\fontsize{7}{7.5}\selectfont  \it (\textbf{blud}, I don't understand what you are saying.)}\\\cmidrule{2-2}
& {\fontsize{8}{8}\selectfont \textbf{Human Translation:} 我听不懂这\underline{哥们}在说什么。} \vspace{2pt}\\
& \hspace{70pt} {\fontsize{7}{7.5}\selectfont  \it (I can't understand what this \textbf{guy} is saying.)} \vspace{3pt}\\\toprule

\multirow{9.5}{*}{\textbf{Noctor}} & {\fontsize{8}{8}\selectfont  {\textbf{Input:} Is the narcissist in your life a \textbf{noctor} and diagnosing you?}} \vspace{2pt}\\
& \hspace{23pt} {\fontsize{7}{7.5}\selectfont (A noctor is a health professional (usually nurse) who takes on some traditional roles performed by the doctor.)} \\\cmidrule{2-2}
& {\fontsize{8}{8}\selectfont \textbf{Model Output} (\hspace{0.083333em}\sethlcolor{color7bg}\textbf{\textcolor{black}{\hl{Incomprehensible}}\hspace{0.083333em}):   你生活中的自恋者是\underline{夜幕降临}，诊断着你吗？}} \vspace{3pt}\\
& \hspace{121pt} {\fontsize{7}{7.5}\selectfont \it (Is the narcissist in your life like \textbf{nightfall}, diagnosing you?)}
\\\cmidrule{2-2}
& {\fontsize{8}{8}\selectfont \textbf{Human Translation:} 那个在你生活中的自恋者是不是\underline{冒充医生}给你做诊断？} \vspace{2pt}\\
& \hspace{70pt} {\fontsize{7}{7.5}\selectfont  \it (Is the narcissist in your life \textbf{pretending to be a doctor} and diagnosing you?)} \vspace{3pt}\\\toprule
\end{tabular}
\caption{Example model outputs for all possible translation categories. For each neologism example, the English input and Chinese output is reported. A gold reference definition of the neologism is provided. (Neologism definitions and English translations are shown for information only.)}
\label{tab:translation_examples}
\end{CJK*}
\end{table*}

\begin{table*}[t]
\centering
\small
\renewcommand{\arraystretch}{0.88}
\resizebox{0.98\textwidth}{!}{
\begin{tabular}{ll|cccccc||cc}
\toprule
& & \multicolumn{6}{c||}{\textbf{GPT-4 Eval. (750)}} & \multicolumn{2}{c}{\textbf{Human Eval. (100)}} \\

& \textbf{Model} & \cellcolor{red!50}\textbf{Incorrect} & \cellcolor{red!50}\textbf{Omitted} & \cellcolor{darkgreen!50}\textbf{Worse} & \cellcolor{darkgreen!50}\textbf{Better} & \textbf{Avg. Length} & \textbf{Acc. ($\uparrow$)} & \textbf{Acc. ($\uparrow$)} & \textbf{Cohen's $\kappa$ ($\uparrow$)} \\
\midrule
\parbox[t]{2mm}{\multirow{6}{*}{\rotatebox[origin=c]{90}{Pre-trained}}} & GPT-J 6B & \cellcolor{red!50}494 & \cellcolor{red!50}0 & \cellcolor{darkgreen!50}190 & \cellcolor{darkgreen!50}66 & 19.04 & 0.341 & 0.38 & 0.711 \\
& LLaMA-1 7B & \cellcolor{red!50}384 & \cellcolor{red!50}24 & \cellcolor{darkgreen!50}240 & \cellcolor{darkgreen!50}102 & 16.69 & 0.456 & 0.48 & 0.697 \\
& OLMo 7B & \cellcolor{red!50}297 & \cellcolor{red!50}0 & \cellcolor{darkgreen!50}234 & \cellcolor{darkgreen!50}219 & 19.24 & 0.604 & 0.64 & 0.729 \\
& Mistral 7B & \cellcolor{red!50}280 & \cellcolor{red!50}0 & \cellcolor{darkgreen!50}291 & \cellcolor{darkgreen!50}179 & 17.35 & 0.627 & 0.64 & 0.768 \\
& LLaMA-2 7B & \cellcolor{red!50}311 & \cellcolor{red!50}42 & \cellcolor{darkgreen!50}251 & \cellcolor{darkgreen!50}144 & 19.49 & 0.529 & 0.61 & 0.681 \\
& LLaMA-2 13B & \cellcolor{red!50}262 & \cellcolor{red!50}81 & \cellcolor{darkgreen!50}255 & \cellcolor{darkgreen!50}152 & 18.76 & 0.544 & 0.56 & 0.698 \\
& LLaMA-2 70B & \cellcolor{red!50}191 & \cellcolor{red!50}94 & \cellcolor{darkgreen!50}260 & \cellcolor{darkgreen!50}205 & 17.29 & 0.620 & 0.67 & 0.827 \\
& GPT 3.5 & \cellcolor{red!50}95 & \cellcolor{red!50}96 & \cellcolor{darkgreen!50}\textbf{46} & \cellcolor{darkgreen!50}513 & 41.21 & 0.745 & 0.72 & 0.828 \\\midrule
\parbox[t]{2mm}{\multirow{6}{*}{\rotatebox[origin=c]{90}{Instruct}}} & Flan-T5 Large & \cellcolor{red!50}638 &  \cellcolor{red!50}0 & \cellcolor{darkgreen!50}100 & \cellcolor{darkgreen!50}12 & 15.42 & 0.149 & 0.17 & 0.670 \\
& Alpaca 7B & \cellcolor{red!50}374 & \cellcolor{red!50}0 & \cellcolor{darkgreen!50}112 & \cellcolor{darkgreen!50}264 & 34.18 & 0.501 & 0.56 & 0.761 \\
& OLMo Instruct 7B & \cellcolor{red!50}297 & \cellcolor{red!50}0 & \cellcolor{darkgreen!50}144 & \cellcolor{darkgreen!50}309 & 27.37 & 0.604 & 0.68 & 0.737 \\
& Mistral Instruct 7B & \cellcolor{red!50}307 & \cellcolor{red!50}0 & \cellcolor{darkgreen!50}165 & \cellcolor{darkgreen!50}278 & 22.15 & 0.591 & 0.65 & 0.643 \\
& LLaMA-2 Chat 7B & \cellcolor{red!50}320 & \cellcolor{red!50}0 & \cellcolor{darkgreen!50}155 & \cellcolor{darkgreen!50}275 & 24.02 & 0.573 & 0.58 & 0.758 \\
& LLaMA-2 Chat 13B & \cellcolor{red!50}313 & \cellcolor{red!50}0 & \cellcolor{darkgreen!50}178 & \cellcolor{darkgreen!50}259 & 24.14 & 0.583 & 0.66 & 0.731 \\
& LLaMA-2 Chat 70B & \cellcolor{red!50}300 & \cellcolor{red!50}1 & \cellcolor{darkgreen!50}200 & \cellcolor{darkgreen!50}249 & 24.22 & 0.599 & 0.64 & 0.771 \\
& GPT 4 & \cellcolor{red!50}\text  {106} & \cellcolor{red!50}18 & \cellcolor{darkgreen!50}62 & \cellcolor{darkgreen!50}\textbf{564} & 38.09 & \textbf{0.835} & \textbf{0.85} & \textbf{0.891} \\
\end{tabular}
} 
\caption{Full results of the Definition Generation task showing the number of \hlc[darkgreen!50]{\textbf{correct}} and \hlc[red!50]{\textbf{incorrect}} answers per model. A sample of 100 neologisms are manually evaluated, and Cohen's Kappa is calculated to determine annotator agreement between GPT-4 and human evaluation. Model accuracy is reported for both manual and GPT-4 evaluation based on the number of definitions evaluated. Correct outputs are divided between being \underline{worse} or \underline{better} than the human reference sentences. Wrong definitions are divided between \underline{incorrect} or \underline{omitted} output where models do not provide a definition when prompted. Average length of definitions is reported with number of words.}
\label{tab:opendomain_full}
\end{table*}

\begin{table*}[t]
\small
\centering
\setlength{\tabcolsep}{3pt}
\renewcommand{\arraystretch}{0.7}
\begin{tabular}{P{0.13\textwidth}P{0.84\textwidth}}
\toprule
\multicolumn{2}{c}{\fontsize{9}{10.5}\selectfont \sethlcolor{lightblue}\textbf{a) \textcolor{black}{\hl{Better}}} \textbf{Model Output Examples}}\vspace{2pt}\\\toprule
\multirow{6}{0.13\textwidth}{\textbf{Inclusive Education}} & {\fontsize{8}{8}\selectfont  {\textbf{Reference Definition:} \it Inclusive education refers to all students being able to access and gain equal opportunities to education and learning. }} \\\cmidrule{2-2}
& {\fontsize{8}{8}\selectfont  \sethlcolor{lightblue}\textbf{\textcolor{black}{\hl{Flan -- T5 Large}}} \textbf{Output:} \it Inclusive education is a philosophy that emphasizes the inclusion of all students, regardless of their disability, gender, race, religion, or sexuality.} \vspace{2pt}\\\toprule
\multirow{7.2}{*}{\textbf{Greenwashing}} & {\fontsize{8}{8}\selectfont  {\textbf{Reference Definition:} \it Greenwashing is the act or practice of making a product, policy, activity, etc. appear to be more environmentally friendly or less environmentally damaging than it really is.}} \\\cmidrule{2-2}
& {\fontsize{8}{8}\selectfont  \sethlcolor{lightblue}\textbf{\textcolor{black}{\hl{Alpaca 7B}}} \textbf{Output:} \it Greenwashing is the practice of misleading consumers about the environmental practices of a company or product. It involves claiming to be environmentally friendly without having any genuine commitment to sustainability.}  \vspace{2pt}\\\toprule
\multirow{4.5}{*}{\textbf{TCG}} & {\fontsize{8}{8}\selectfont  {\textbf{Reference Definition:} \it TCG is an acronym for "trading card game."}} \\\cmidrule{2-2}
& {\fontsize{8}{8}\selectfont  \sethlcolor{lightblue}\textbf{\textcolor{black}{\hl{LLaMA -- 2 Chat 7B}}} \textbf{Output:} \it TCG stands for Trading Card Game, a type of game where players use cards to represent characters, spells, and other elements to battle each other.} \vspace{2pt}\\\toprule
\multirow{4.5}{*}{\textbf{Mermaidcore}} &  {\fontsize{8}{8}\selectfont  {\textbf{Reference Definition:} \it Mermaidcore is an aesthetic focusing on imagery and fashion related to mermaids.}} \\\cmidrule{2-2}
& {\fontsize{8}{8}\selectfont  \sethlcolor{lightblue}\textbf{\textcolor{black}{\hl{LLaMA -- 2 Chat 13B}}} \textbf{Output:} \it Mermaidcore is a type of aesthetic that features pastel colors, sparkly accessories, and flowing fabrics, inspired by the mythical creatures known as mermaids.}
\vspace{2pt}\\\toprule
\multirow{4.5}{*}{\textbf{Wokeness}} & {\fontsize{8}{8}\selectfont  {\textbf{Reference Definition:} \it Wokeness is the quality of being alert and concerned about social injustice and discrimination.}} \\\cmidrule{2-2}
& {\fontsize{8}{8}\selectfont  \sethlcolor{lightblue}\textbf{\textcolor{black}{\hl{LLaMA -- 2 Chat 70B}}} \textbf{Output:} \it It is a term used to describe the state of being aware of and attuned to social injustices, particularly those related to racism, sexism, and economic inequality.}
\vspace{2pt}\\\toprule
\multirow{5.5}{*}{\textbf{Longcovid}} & {\fontsize{8}{8}\selectfont {\textbf{Reference Definition:} \it Longcovid is the signs and conditions that continue or develop after acute COVID-19 infection. }} \\\cmidrule{2-2}
& {\fontsize{8}{8}\selectfont  \sethlcolor{lightblue}\textbf{\textcolor{black}{\hl{GPT -- 4}}} \textbf{Output:} \it Long COVID, also known as post-acute sequelae of SARS-CoV-2 infection (PASC), refers to a range of symptoms that can continue for weeks or months after the acute phase of a COVID-19 infection has resolved. Symptoms may include fatigue, breathlessness, brain fog, depression and anxiety, pain, and other long-term complications.}
\vspace{2pt}\\\toprule
\multicolumn{2}{c}{\fontsize{9}{10.5}\selectfont \sethlcolor{pink}\textbf{b) \textcolor{black}{\hl{Omitted}} Model Output Examples}}\vspace{2pt}\\\toprule
\multirow{7.5}{*}{\textbf{Falstalgia}} & {\fontsize{8}{8}\selectfont  {\textbf{Reference Definition:} \it Falstalgia is a misguided feeling of nostalgia, a longing for a past that really wasn't that great and you weren't around for. }} \\\cmidrule{2-2}
& {\fontsize{8}{8}\selectfont  \sethlcolor{pink}\textbf{\textcolor{black}{\hl{GPT -- 4}}} \textbf{Output:} \it I'm sorry, "falstalgia" does not appear to be a recognized term in English language or any other languages. It could be a typo. If you meant "nostalgia," it is a sentimental longing or wistful affection for the past, typically for a period or place with happy personal associations.}  \vspace{2pt}\\\toprule
\multirow{4.5}{*}{\textbf{Flurona}} & {\fontsize{8}{8}\selectfont  {\textbf{Reference Definition:} \it Flurona is a coinfection with influenza and COVID-19. }} \\\cmidrule{2-2}
& {\fontsize{8}{8}\selectfont  \sethlcolor{pink}\textbf{\textcolor{black}{\hl{GPT -- 3.5}}} \textbf{Output:} \it I'm sorry, but I couldn't find any information on the term "flurona." It is possible that it is a newly coined term or a term that is not widely recognized.} \vspace{2pt}\\\bottomrule
\end{tabular}

\caption{Example model outputs of the definition generation task. \textbf{a)} Instruction-tuned model outputs evaluated as better than the reference definition by GPT-4 and \textbf{b)} GPT model outputs that omit definitions are provided.}
\label{tab:length_instruct}
\end{table*}

\end{document}